%% file: ieee-iv.tex
\documentclass[journal]{IEEEtran}

\usepackage{graphicx}
\usepackage{cite}
\usepackage{xcolor}
\usepackage{hyperref}
\usepackage[per-mode=symbol]{siunitx}
\usepackage{proof}
\usepackage{bussproofs}
\usepackage{mathpartir}
\usepackage{rotating}
\usepackage{subcaption}

\usepackage{algorithm2e}
\SetKwComment{Comment}{/* }{ */}
\RestyleAlgo{ruled}

\usepackage[all]{xy} \SelectTips{cm}{}  
\newdir{ >}{{}*!/-8pt/@{>}}
\newdir{|>}{%
	!/1.6pt/@{|}*:(1,-.2)@^{>}*:(1,+.2)@_{>}}

\usepackage{todonotes}
\newcommand{\todoil}[1]{\todo[inline,caption={}]{#1}}

\newcommand{\todooptil}[1]{}

\usepackage{mathtools}
\usepackage{hyperref}
\usepackage[capitalize, nameinlink]{cleveref}

\usepackage{aliascnt}
\usepackage[appendix=inline]{apxproof}

\theoremstyle{plain}
\newtheoremrep{theorem}{Theorem}[section]
\Crefname{theorem}{Theorem}{Theorems}
\newaliascnt{propositioncnt}{theorem}
\newtheoremrep{proposition}[propositioncnt]{Proposition}
\Crefname{propositioncnt}{Proposition}{Propositions}
\newaliascnt{lemmacnt}{theorem}
\newtheoremrep{lemma}[lemmacnt]{Lemma}
\Crefname{lemmacnt}{Lemma}{Lemmas}
\newaliascnt{corollarycnt}{theorem}
\newtheoremrep{corollary}[corollarycnt]{Corollary}
\Crefname{corollarycnt}{Cor.}{Cor.}
\newaliascnt{factcnt}{theorem}

\Crefname{factcnt}{Fact}{Fact}
\newaliascnt{assumptioncnt}{theorem}
\newtheorem{assumption}[assumptioncnt]{Assumption}
\Crefname{assumptioncnt}{Asm.}{Asm.}
\theoremstyle{definition}
\newaliascnt{remarkcnt}{theorem}
\newtheorem{remark}[remarkcnt]{Remark}
\Crefname{remarkcnt}{Remark}{Remarks}
\newaliascnt{notationcnt}{theorem}
\newtheorem{notation}[notationcnt]{Notation}
\Crefname{notationcnt}{Notation}{Notation}

\newaliascnt{requirementcnt}{theorem}

\Crefname{requirementcnt}{Requirement}{Requirement}
\newaliascnt{requirementscnt}{theorem}
\newtheorem{requirements}[requirementscnt]{Requirements}
\Crefname{requirementscnt}{Requirements}{Requirements}
\newcounter{researchquestioncnt}
\newtheorem{researchquestion}[researchquestioncnt]{RQ}
\Crefname{researchquestioncnt}{RQ}{RQ}

\theoremstyle{definition}
\newaliascnt{definitioncnt}{theorem}
\newtheoremrep{definition}[definitioncnt]{Definition}
\Crefname{definitioncnt}{Definition}{Definitions}
\newaliascnt{examplecnt}{theorem}
\newtheoremrep{example}[examplecnt]{Example}
\Crefname{examplecnt}{Example}{Examples}

\crefformat{section}{Section~#2#1#3}
\Crefname{table}{Table}{Table}
\Crefname{figure}{Figure}{Figure}
\Crefname{equation}{}{}
\Crefname{line}{Line}{Line}

\usepackage{stmaryrd,amssymb,amsmath}

\usepackage{mathrsfs}

\usepackage{wrapfig}
\usepackage{comment}
\specialcomment{auxproof}
{\mbox{}\newline\textbf{BEGIN: AUX-PROOF}\dotfill\newline}
{\mbox{}\newline\textbf{END: AUX-PROOF}\dotfill\newline}
\excludecomment{auxproof}

\newcommand{\dHL}{\ensuremath{\mathrm{dFHL}}}
\newcommand{\dL}{\ensuremath{\mathrm{dL}}}    

\newcommand{\xtgt}{y_{\mathrm{tgt}}}
\newcommand{\ytgt}{y_{\mathrm{tgt}}}
\newcommand{\ytgtn}[1]{y_{\mathrm{tgt}#1}}
\newcommand{\bmax}{b_{\mathrm{max}}}
\newcommand{\bmin}{b_{\mathrm{min}}}
\newcommand{\amax}{a_{\mathrm{max}}}
\newcommand{\vmin}{v_{\mathrm{min}}}
\newcommand{\vmax}{v_{\mathrm{max}}}
\newcommand{\tlc}{t_{\mathrm{LC}}}

\newcommand{\bb}{\mathtt{b}}
\newcommand{\cc}{\mathtt{c}}

\newcommand{\DM}{\text{DM}}
\newcommand{\AC}{\text{AC}}
\newcommand{\ACCA}{\ensuremath{\text{AC+}\text{\textit{RSS}}^{\text{CA}}}}
\newcommand{\ACGA}{\ensuremath{\text{AC+}\text{\textit{RSS}}^{\text{GA}}}}

\newcommand{\POV}[1]{\ensuremath{\mathsf{POV}{#1}}}
\newcommand{\SV}{\ensuremath{\mathsf{SV}}}

\newcommand{\aheadSL}{\ensuremath{\mathsf{aheadSL}}}
\newcommand{\behindSL}{\ensuremath{\mathsf{behindSL}}}
\newcommand{\true}{\ensuremath{\mathsf{true}}}
\newcommand{\false}{\ensuremath{\mathsf{false}}}

\newcommand{\BC}{BC}

\newcommand{\hoare}[3]{\left\{#1\right\}~#2~\left\{#3\right\}}
\newcommand{\hquad}[4]{\left\{#2\right\}~#3~\left\{#4\right\}\colon{}#1}

\newcommand{\dRSS}{\ensuremath{\mathsf{dRSS}}}
\newcommand{\dRSSpm}{\ensuremath{\dRSS_\pm}}

\newcommand{\videopointer}[1]{\footnote{\url{https://bit.ly/#1}}}

\def\videoAC{\videopointer{3r3IvRW}}
\def\videoACtwo{\videopointer{33qJy6w}}
\def\videoACRSSCA{\videopointer{3FqRYIp}}
\def\videoACRSSCAtwo{\videopointer{3zKCuNR}}
\def\videoACRSSGA{\videopointer{3qdCcMl}}
\def\videoACRSSGAtwo{\videopointer{31Ib7Ye}}

\newcommand{\bigquadnnl}[4]{
  \begin{array}{r}
    \left\{#2\right\} {#3} \left\{#4\right\}
    \colon{}#1
  \end{array}
}

\newcommand{\Safe}{\mathsf{Safe}}
\newcommand{\Env}{\mathsf{Env}}
\newcommand{\Goal}{\mathsf{Goal}}

\newcommand{\Variables}{\ensuremath{V}}
\newcommand{\expa}{\ensuremath{e}}
\newcommand{\fun}{\ensuremath{f}}

\newcommand{\val}{\ensuremath{v}}
\newcommand{\var}{\ensuremath{x}}
\newcommand{\term}{\ensuremath{e}}
\newcommand{\vars}{\ensuremath{\mathbf{x}}}
\newcommand{\funs}{\ensuremath{\mathbf{f}}}
\newcommand{\funsa}{\ensuremath{\mathbf{f}}}
\newcommand{\funsb}{\ensuremath{\mathbf{g}}}
\newcommand{\invariant}{\ensuremath{e_\mathsf{inv}}}
\newcommand{\invarianti}[1]{\ensuremath{e_\mathsf{inv,#1}}}
\newcommand{\variant}{\ensuremath{e_\mathsf{var}}}
\newcommand{\varianti}[1]{\ensuremath{e_\mathsf{var,#1}}}
\newcommand{\terminator}{\ensuremath{e_\mathsf{ter}}}
\newcommand{\terminatori}[1]{\ensuremath{e_\mathsf{ter,#1}}}

\newcommand{\asserta}{\ensuremath{A}}
\newcommand{\assertb}{\ensuremath{B}}
\newcommand{\assertc}{\ensuremath{C}}
\newcommand{\assertd}{\ensuremath{D}}
\newcommand{\asserte}{\ensuremath{E}}
\newcommand{\assertf}{\ensuremath{F}}
\newcommand{\safetya}{\ensuremath{S}}

\newcommand{\subst}[3]{{#1}[{#2} / {#3}]}

\newcommand{\coma}{\alpha}
\newcommand{\comb}{\beta}
\newcommand{\dynamics}[2]{\delta_{#1}^{#2}}

\newcommand{\skipClause}{\mathsf{skip}}
\newcommand{\seqClause}[2]{{#1};{#2}}
\newcommand{\assignClause}[2]{#1 \mathop{{:}{=}} #2}

\newcommand{\dwhileKeyword}{\mathsf{dwhile}}
\newcommand{\dwhileHeader}[1]{\dwhileKeyword\,(#1)}
\newcommand{\dwhileClause}[2]{\dwhileHeader{#1}\left\{\,#2\,\right\}}
\newcommand{\dwhileClauseNb}[2]{\dwhileHeader{#1}\,#2}
\newcommand{\odeClause}[2]{\dot{#1} = #2}

\newcommand{\whileKeyword}{\mathsf{while}}
\newcommand{\whileHeader}[1]{\whileKeyword\,(#1)}
\newcommand{\whileClause}[2]{\whileHeader{#1}\,#2}

\newcommand{\ifKeyword}{\mathsf{if}}
\newcommand{\ifHeader}[1]{\mathsf{if}\,(#1)}
\newcommand{\elseKeyword}{\mathsf{else}}
\newcommand{\ifThenElse}[3]{\ifHeader{#1}\,#2\,\elseKeyword{}\,#3}

\newcommand{\caseKeyword}{\mathsf{case}}
\newcommand{\case}[4]{\caseKeyword\,({#1})\,{#2}\ldots({#3})\,{#4}}

\newcommand{\limply}{\Rightarrow}
\newcommand{\bigland}{\bigwedge}
\newcommand{\biglor}{\bigvee}

\newcommand{\skipcrule}{\textsc{Skip}}
\newcommand{\seqrule}{\textsc{Seq}}
\newcommand{\assignrule}{\textsc{Assign}}
\newcommand{\ifrule}{\textsc{If}}
\newcommand{\whilerule}{\textsc{Wh}}
\newcommand{\dwhilerule}{\textsc{DWh}}
\newcommand{\dwhilesolrule}{\textsc{DWh-Sol}}
\newcommand{\limprule}{\textsc{LImp}}
\newcommand{\conjrule}{\textsc{Conj}}
\newcommand{\caserule}{\textsc{Case}}

\newcommand{\store}{\ensuremath{\rho}}
\newcommand{\update}[3]{\ensuremath{{#1}[{#2} \to {#3}]}}

\newcommand{\sem}[2]{\ensuremath{\left\llbracket {#1} \right\rrbracket_{#2}}}
\newcommand{\statea}{\ensuremath{s}}
\newcommand{\state}[2]{\ensuremath{\langle {#1}, {#2} \rangle}}
\newcommand{\red}[1]{\to}
\newcommand{\redext}[1]{\to_{#1}}
\newcommand{\convergeSymbol}{\Downarrow}
\newcommand{\converge}[2]{\ensuremath{{#1} \mathrel{\convergeSymbol} {#2}}}
\newcommand{\lieder}[3]{\mathcal{L}_{\odeClause{#1}{#2}}\,#3}
\newcommand{\liederdyn}[2]{\mathcal{L}_{#1}\,#2}
\newcommand{\partialder}[2]{\frac{\partial #1}{\partial #2}}

\newcommand{\set}[1]{\left\{ {#1} \right\}}
\newcommand{\setcomp}[2]{\left\{ {#1} \,  \middle| \, {#2} \right\}}

\newcommand{\N}{\ensuremath{\mathbb{N}}}
\newcommand{\R}{\ensuremath{\mathbb{R}}}
\newcommand{\Var}{\ensuremath{\mathbf{Var}}}

\newcommand{\NA}{N/A}

\newcommand{\safetystring}{How does the GA-RSS-supervised controller perform in terms of
safety?}
\newcommand{\goalstring}{How does the GA-RSS-supervised controller perform in terms of
accomplishing its goal?}
\newcommand{\practicalitystr}{Can the GA-RSS-supervised controller be useful in
practice}
\newcommand{\practicalitystring}{\practicalitystr{}?}
\newcommand{\practicalitystringlong}{\practicalitystr{} (e.g.~in terms
of computation speed and weakness of the RSS condition)?}
\newcommand{\metricsstring}{How does the GA-RSS-supervised controller perform in terms
of other significant metrics (progress, comfort, etc.)?}
\newcommand{\intrusionstring}{How often is \AC{} in control during
execution?}

\newcommand{\comai}{\coma_i}
\newcommand{\comaone}{\coma_1}
\newcommand{\comatwo}{\coma_2}
\newcommand{\comathree}{\coma_3}
\newcommand{\comafive}{\coma_5}
\newcommand{\comasix}{\coma_6}
\newcommand{\comaseven}{\coma_7}
\newcommand{\comaij}{\coma_{i,j}}
\newcommand{\comathreeseven}{\coma_{3,7}}
\newcommand{\comafourseven}{\coma_{4,7}}
\newcommand{\comasixseven}{\coma_{6,7}}
\newcommand{\inv}{\mathsf{inv}}
\newcommand{\safetyinv}{\safetya_\mathsf{inv}}

\newcommand{\KeYmaeraX}{\textsc{KeYmaera~X}}  

\newif\iffull\fulltrue 

\begin{document}
%
\title{Goal-Aware RSS for Complex Scenarios\\ via Program Logic
}
%
%
%

\author{
  Ichiro Hasuo$^{1,7,*}$,
  Clovis Eberhart$^{1,8,*}$,
  James Haydon$^{1,*}$,
  J\'{e}r\'{e}my Dubut$^{1,8}$,
  Rose Bohrer$^{2,\dagger}$,
  Tsutomu Kobayashi$^{1}$,
  Sasinee Pruekprasert$^{1}$,
  Xiao-Yi Zhang$^{1}$,
  Erik Andr\'{e} Pallas$^{3,\ddagger}$,
  Akihisa Yamada$^{4,1}$,
  Kohei Suenaga$^{5,1}$,
  Fuyuki Ishikawa$^{1}$,
  Kenji Kamijo$^{6}$,
  Yoshiyuki Shinya$^{6}$, and
  Takamasa Suetomi$^{6}$
\thanks{
\copyright~2022 IEEE.  Personal use of this material is permitted.  Permission from IEEE must be obtained for all other uses, in any current or future media, including reprinting/republishing this material for advertising or promotional purposes, creating new collective works, for resale or redistribution to servers or lists, or reuse of any copyrighted component of this work in other works.

The work is partially supported by  ERATO HASUO Metamathematics for Systems Design Project (No.\ JPMJER1603) and ACT-I (No.\ JPMJPR17UA), JST; and Grants-in-aid No.\ 19K20215 \& 19K20249, JSPS.
}
\thanks{$^{1}$%
National Institute of Informatics (NII),
Tokyo 101-8430, Japan.
        { \{hasuo, jhaydon, eberhart, dubut, t-kobayashi, sasinee, xiaoyi, f-ishikawa\}@nii.ac.jp}}%
\thanks{$^{2}$
Dept.\ Computer Science, Worcester Polytechnic Institute, 100 Institute Road,
Worcester, MA 01609-2280, USA. {rose.bohrer.cs@gmail.com}
}
\thanks{$^{3}$
Inst.\ Software \& Systems Engineering, University of Augsburg, Universit\"{a}tstra\ss{}e 6a, D-86135 Augsburg, Germany. {erik.pallas@t-online.de}
}
\thanks{$^{4}$
 Cyber Physical Security Research Center, AIST, Aomi 2-4-7, Tokyo 135-0064, Japan.  { akihisa.yamada@aist.go.jp}%
}
\thanks{$^{5}$
Graduate School of Informatics, Kyoto University,
Kyoto 606-8501, Japan.  { ksuenaga@fos.kuis.kyoto-u.ac.jp}%
}
\thanks{$^{6}$ 
 Mazda Motor Corporation,
Fuchu 730-8670, Japan.  { \{kamijyo.k, shinya.y, suetomi.t\}@mazda.co.jp}%
}
\thanks{$^{7}$ 
SOKENDAI (The Graduate University for Advanced Studies),  Japan. 
}
\thanks{$^{8}$ 
Japanese-French Laboratory for Informatics (IRL 3527), Tokyo, Japan
}
\thanks{$^{*}$ 
Equal contribution.
}
\thanks{$^{\dagger}$ 
The work was done during R.B.'s employment at NII, Tokyo.
}
\thanks{$^{\ddagger}$ 
The work was done during E.P.'s internship at NII, Tokyo.
}
}

\markboth{
Accepted for publication in \emph{IEEE Transactions on Intelligent Vehicles}
}%
{Hasuo \MakeLowercase{\textit{et al.}}: Goal-Aware RSS for Complex Scenarios via Program Logic}
%

\IEEEpubid{
\copyright~2022 IEEE
}


\maketitle

\begin{abstract}
We introduce a \emph{goal-aware} extension of 
responsibility-sensitive safety (RSS), a recent methodology for rule-based safety guarantee for automated driving systems (ADS). Making RSS rules guarantee goal achievement---in addition to collision avoidance as in the original RSS---requires complex planning
over long sequences of manoeuvres.
To deal with the complexity, we introduce a compositional reasoning framework based on program logic, in which one can systematically develop RSS rules for smaller subscenarios and combine them to obtain RSS rules for bigger scenarios. As the basis of the framework, we introduce a  program logic $\dHL$  that accommodates continuous dynamics and safety conditions. Our framework presents a $\dHL$-based  workflow for deriving goal-aware RSS rules; we discuss its software support, too.
We conducted experimental evaluation using RSS rules in a safety architecture. Its results show that
 goal-aware RSS is indeed effective in realising both collision avoidance and goal achievement.

\end{abstract}

\begin{IEEEkeywords}
automated driving, safety, rule-based safety, responsibility-sensitive safety (RSS), program logic,  Floyd--Hoare logic, differential dynamics, simplex architecture
\end{IEEEkeywords}

%
\IEEEpeerreviewmaketitle



\section{Introduction}\label{sec:intro}
%
%
%
%

\IEEEPARstart{S}{afety}  of automated driving systems (ADS) is a problem of growing industrial and social interest. New technologies in sensing and planning (such as lidars and deep neural networks) are making ADS technologically possible. However, towards the social acceptance of ADS, their safety should be guaranteed, explained, and agreed upon.

This paper is about \emph{responsibility-sensitive safety} (RSS)~\cite{ShalevShwartzSS17RSS}---a recent rule-based approach to ADS safety.  Our contribution is to 
make the RSS framework \emph{goal-aware},
so that logical ``safety rules'' in RSS 
\begin{itemize}
 \item not only guarantee collision avoidance (as in the original RSS~\cite{ShalevShwartzSS17RSS}),
 \item but also guarantee \emph{goal achievement}, such as changing lanes and stopping at a designated position on the highway shoulder (\cref{ex:pullover}).
\end{itemize}
Goal-aware RSS rules typically involve multiple manoeuvres (accelerating, braking, changing lanes, etc.); 
deriving goal-aware RSS rules and proving their correctness is therefore much more complex compared to the original RSS.
 As technical contribution, we introduce logical, methodological and software infrastructures that realise goal-aware RSS. They are namely
 1) a program logic suited for our purpose (called $\dHL$, \cref{sec:dHL}), 2) a logical workflow for compositional derivation of goal-aware RSS rules (\cref{sec:workflow}), and 3) software support for the workflow (\cref{sec:toolSupport}). We demonstrate the value of our goal-aware RSS by experiments in a safety architecture (\cref{sec:exp}).

\subsection{(Collision-Avoiding) Responsibility-Sensitive Safety}\label{subsec:introCARSS}
(The original) responsibility-sensitive safety (RSS)~\cite{ShalevShwartzSS17RSS} is an approach to ADS safety that has been attracting growing attention. 
RSS aims to  provide \emph{safety rules} that are rigorously formulated in mathematical terms. Unlike most algorithms and techniques studied for ADS, RSS is not so much about \emph{how to drive safely}; it is rather about breaking down the ultimate goal (namely safety \emph{in the future}) into concrete  conditions that only depend on the \emph{current system state}. The core idea of RSS is that those safety rules  should guarantee  ADS safety in the rigorous form of \emph{mathematical proofs}. 

\IEEEpubidadjcol 

Here is an outline of  original RSS~\cite{ShalevShwartzSS17RSS}. We will often call the original RSS~\cite{ShalevShwartzSS17RSS}  \emph{collision-avoiding RSS (CA-RSS)}, in contrast to our extension that we call  \emph{goal-aware RSS (GA-RSS)}. When we simply say RSS, the argument should apply to both CA- and GA-RSS.\footnote{Our introduction of CA-RSS here adapts some terminologies for our purpose of extending it later; the terminologies can therefore differ from those used in~\cite{ShalevShwartzSS17RSS}.
Another logic-oriented introduction to CA-RSS is found in~\cite{Hasuo22RSSarXiv}. }

CA-RSS introduces \emph{RSS rules} in a manner specific to different driving scenarios (driving in a single lane, changing lanes, other vehicles in front or behind, etc.). An RSS rule is a pair $(A,\alpha)$ of 
\begin{itemize}
 \item a logical assertion  $A$ called an \emph{RSS condition}, and
 \item a control strategy $\alpha$ called a \emph{proper response}.
\end{itemize}
An RSS rule is subject to the following requirements. 

\begin{requirements}[requirements on RSS rules, in CA-RSS]\label{req:reqOnRSSRules}
Let $(A,\alpha)$ be an RSS rule.
 Consider an arbitrary execution $E$ of the proper response $\alpha$; assume that the RSS condition $A$ is satisfied at the beginning of $E$. Then 
 \begin{itemize}
 \item (the \emph{collision avoidance} requirement) the execution $E$ in question must exhibit no collision; and
 \item (the \emph{responsibility} requirement) the execution $E$ must satisfy the RSS responsibility principles.
 \end{itemize}
\end{requirements}
The last \emph{RSS responsibility principles}, taken literally from~\cite{ShalevShwartzSS17RSS}, are listed below (cf.\ \cref{rem:RSSRespPrinciples}).
\begin{enumerate}
 	 \item Don't hit the car in front of you.
	 \item 
Don't cut in recklessly.
	 \item 
Right of way is given, not taken.
	 \item 
Be cautious in areas with limited visibility.
	 \item 
If you can avoid a crash without causing another one, you must.
\end{enumerate}
One significance of the RSS framework is that the safety \emph{in the future} (that is, safety during the whole execution $E$ of $\alpha$) is reduced to the RSS condition $A$ \emph{at present} (that is, one that can be checked at the beginning of $E$). In other words,  the  truth of $A$ at present guarantees the safety in the future. 
  This means that, in particular, $A$ and $\alpha$ must take into
  account all possible future evolutions of the driving situation,
  such as sudden braking or acceleration of other vehicles, etc.

Another significance of RSS is the \emph{assume-guarantee reasoning} via responsibilities. Establishing the collision avoidance requirement (\cref{req:reqOnRSSRules}) for the subject vehicle (\SV{}) is usually impossible without suitable assumptions on other vehicles' behaviours---imagine a malicious vehicle that actively chases others and hits them. In RSS, one can impose the RSS responsibility principles on other vehicles and limit their behaviours; reciprocally, \SV{}  must obey the same principles, too.

\begin{remark}\label{rem:RSSRespPrinciples}
The five RSS responsibility principles (as we call them) are often called ``safety rules'' and ``common sense rules'' in the RSS literature such as~\cite{ShalevShwartzSS17RSS}. These principles are often presented as the main concept of RSS---especially in the presentation to the general public, such as Mobileye/Intel's webpage.\footnote{\url{https://www.mobileye.com/responsibility-sensitive-safety}} However, we believe that the logical framework of RSS (including reduction of the future to the present and assume-guarantee reasoning, as discussed above) is at least as important. The focus of the current paper is formalising and extending this logical framework of RSS.
\end{remark}

\begin{figure}[tbp]
 \centering
 \includegraphics[bb=66 217 305 271,clip,width=13em]{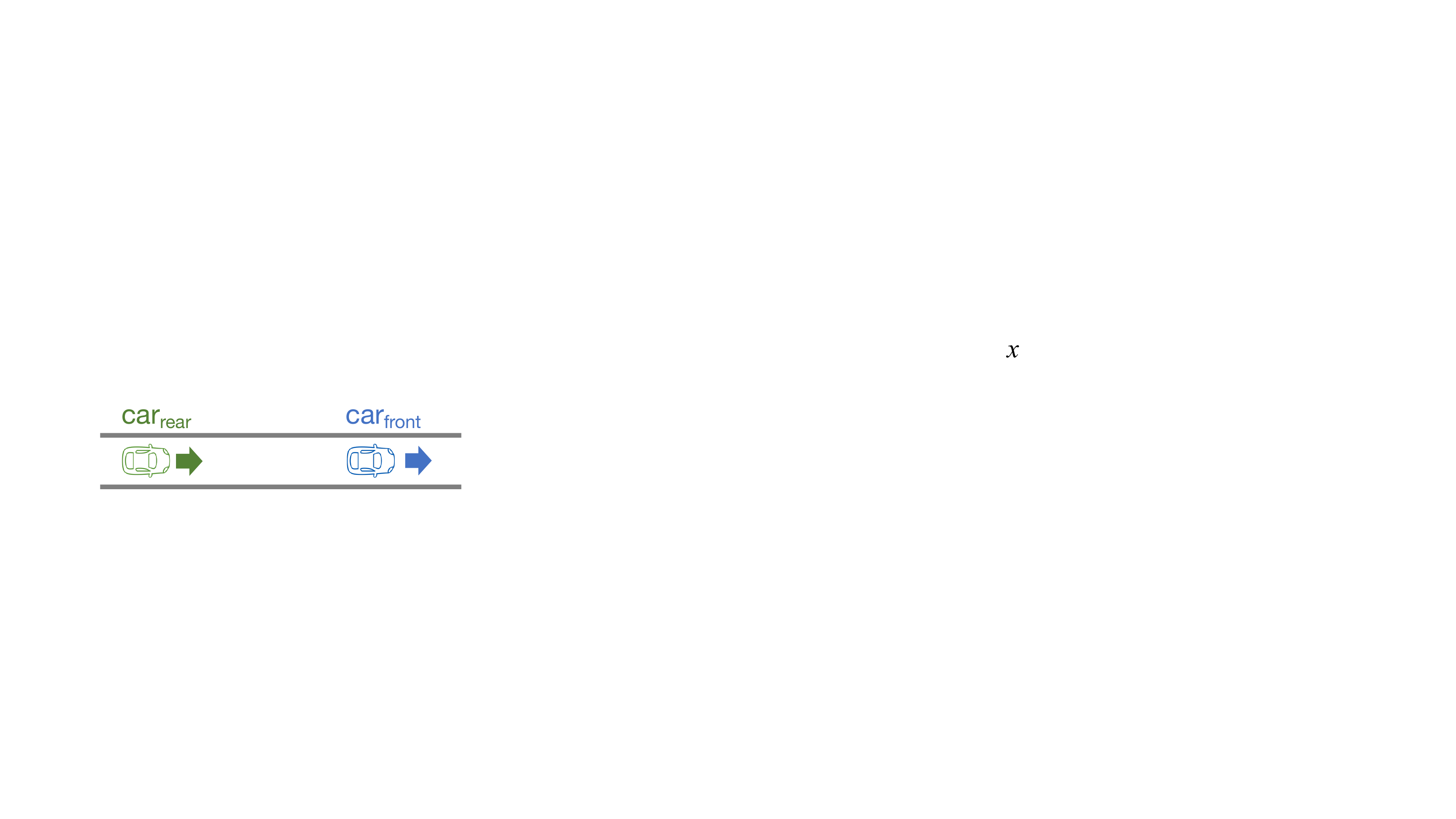}
 \caption{The one-way traffic scenario}
 \label{fig:onewayTraffic}
\end{figure}

\begin{example}[a CA-RSS rule for one-way traffic]\label{ex:onewayTraffic}
Consider the one-way traffic scenario shown
  in~\cref{fig:onewayTraffic}, where the subject vehicle
  (\SV{}, $\mathrm{car}_\mathrm{rear}$) drives behind another car
  ($\mathrm{car}_\mathrm{front}$). The (collision-avoiding) RSS rule for this simple scenario,
  presented in~\cite{ShalevShwartzSS17RSS}, is $(A,\alpha)$ defined as follows.

\underline{\bfseries The RSS condition $A$}
  The RSS condition $A$ is
  \begin{equation}\label{eq:RSSOnewayTraffic}
    A \;=\;\bigl(y_{f} - y_{r} > \dRSS(v_{f}, v_{r})\bigr),
  \end{equation}
  where $\dRSS(v_{f}, v_{r})$ is the \emph{RSS safety distance} defined by
  \begin{equation}\label{eq:RSSMinDist}
    \begin{aligned}
    &   \dRSS(v_{f}, v_{r}) :=\\
    &  \max\left(\,0,\,
          v_{r}\rho + \frac{1}{2}\amax  \rho^2 + \frac{(v_{r} + \amax  \rho)^2}{2\bmin} -\frac{v_{f}^2}{2\bmax}\,\right).
    \end{aligned}
  \end{equation}
  Here $y_{f}, y_{r}$ are the positions of the two cars, and
  $v_{f}, v_{r}$ are their velocities (their dynamics are modelled in the
  1-dimensional lane coordinate). The other parameters are as follows: $\rho$
  is the maximum \emph{response time} that $\mathrm{car}_\mathrm{rear}$
  might take to initiate the required braking; $a_{\max}$ is the maximum
  (forward) acceleration rate of $\mathrm{car}_\mathrm{rear}$;
  $\bmin$ is the maximum comfortable braking rate for
  $\mathrm{car}_\mathrm{rear}$; and $\bmax$ is the maximum emergency
  braking rate for $\mathrm{car}_\mathrm{front}$.

 \underline{\bfseries The proper response $\alpha$}
 The proper response $\alpha$  dictates  \SV{}
        ($\mathrm{car}_\mathrm{rear}$) to engage the maximum comfortable braking (at rate $\bmin$)
        when condition~\cref{eq:RSSOnewayTraffic} is about to be violated.

That the RSS rule $(A,\alpha)$ satisfies the collision avoidance requirement (\cref{req:reqOnRSSRules}) is proved in the original RSS paper~\cite{ShalevShwartzSS17RSS}. We also give a formal proof later in \cref{ex:oneway-proof}, using the logic $\dHL$ we introduce for our purpose of formalising reasoning in RSS. 
\end{example}


\subsection{Usages of RSS}\label{subsec:introRSSUsages}
Before introducing our goal-aware extension of RSS, we discuss some usages of (CA- and GA-)RSS, hoping that the discussion further illustrates the goals and features of RSS.

A distinguishing feature of RSS is that it gives \emph{a priori} rules for rigorous safety guarantee.
This is in contrast with 
\begin{itemize}
 \item 
 many optimisation- and learning-based planning algorithms for safe driving, such as~\cite{mcnaughton2011motion} (they do not offer rigorous safety guarantee),
 \item 
 testing-based approaches for ADS safety, such as~\cite{LuoZAJZIWX21ASEtoAppear} (they do not offer rigorous safety guarantee, either), and
 \item 
 runtime verification approaches for ADS safety by reachability analysis, such as~\cite{LiuPA20IV,PekA18IROS}.
\end{itemize}
(See \cref{subsec:relatedAndFutureWork} for further discussion.)
This feature has enabled multiple unique usages of RSS, as we discuss below. These usages have been already pursued in the literature for CA-RSS; we expect similar usages for our GA-RSS as well.

One usage of RSS is for \emph{attribution of liability}~\cite{ShashuaSS18NHTSA}, that is, to identify culpable parties in accidents. RSS
rules are designed so that there is no collision as long as all parties comply with them (\cref{req:reqOnRSSRules}); therefore, in an accident, at least one party was
\emph{not} compliant and is therefore culpable. 

Another usage is as a \emph{safety metric} (discussed and/or used
in~\cite{SilberlingWAKL20,WangLS20RealisticSingleShotLongTerm,AltekarEWCWCRJ20,ZhaoZMSLLZ_IV20,WengRDSB20IV}). Here,
the risk of a given situation can be measured by either 1) the degree with which the RSS condition of a relevant RSS rule is violated, or 2) whether the proper response of the RSS rule is not engaged while it should.

 Another obvious usage of RSS is for \emph{formal reasoning} about ADS safety:
by proving that  \SV{} complies with RSS rules, one can prove \emph{a priori} that  \SV{} is
never responsible for accidents. Often one does not go so far as
formally proving \SV{}'s compliance with RSS rules. Even in that case, collecting
empirical evidences for RSS compliance, e.g.\ by testing, allows one to
establish logical \emph{safety cases}. The importance of such safety cases are
emphasised in standards such as UL 4600~\cite{UL4600_202004}; the use of RSS is
advocated in the current efforts towards the IEEE 2846 standard.

Yet another usage of RSS is as part of a \emph{safety architecture}, whose detailed introduction is deferred to \cref{subsec:introRSSSafetyArchitecture}.  This is a variation of the last usage  (formal safety reasoning), but is more widely and easily deployable, and is therefore attracting a lot of attention (see e.g.~\cite[Figure~1]{OborilS20IV}).  Our experimental evaluation (\cref{sec:exp}) follows this usage.

After all,  RSS rules are not only for \emph{making ADS safer} but also for \emph{limiting liabilities}. In the dawn of automated driving today, ADS vendors are under a lot of pressure to ensure the safety of their products, fearing the possibilities of unexpected or excessive liabilities. RSS rules cut clear mathematical bounds of the vendors' liabilities, easing their safety assurance efforts. 

\subsection{Goal-Aware RSS}\label{subsec:goalAwareRSS}
We seek a \emph{goal-aware} extension of the original (collision-avoiding) RSS~\cite{ShalevShwartzSS17RSS}, so that the RSS rules  are not only concerned with collision avoidance but also with
achieving a goal.

\begin{requirements}[requirements on RSS rules, in GA-RSS]\label{req:reqOnRSSRulesGA}
In \emph{goal-aware RSS (GA-RSS)}, an RSS rule $(A,\alpha)$ must satisfy the following: for any execution $E$ of the proper response $\alpha$ that starts at a state where the RSS condition $A$ is true,
\begin{itemize}
 \item (\emph{collision avoidance}, the same as in \cref{req:reqOnRSSRules}),
 \item (\emph{responsibility}, the same as in \cref{req:reqOnRSSRules}), and
 \item (the \emph{goal achievement} requirement) the specified goal is achieved at the end of $E$. 
\end{itemize}
\end{requirements}
Deriving such goal-aware RSS rules and establishing their correctness (in the sense of \cref{req:reqOnRSSRulesGA}) pose multiple technical challenges. They include
\begin{itemize}
 \item the formalisation of goals to be achieved,
 \item the identification of proper responses $\alpha$, which would involve multiple manoeuvres (accelerating, braking, changing lanes, etc.), 
 \item the identification of RSS conditions $A$ that guarantee both collision avoidance and goal achievement along/after complex controls described by $\alpha$, 
\end{itemize}
and so on. The technical contribution of the current paper is a program logic framework that addresses these challenges. It enables compositional derivation of goal-aware RSS rules, as we will describe in~\cref{subsec:introCompDeriv}.



The following is our leading example for goal-aware RSS. 


\begin{example}[the pull over scenario]\label{ex:pullover}
\begin{auxproof}
   \todoil{Change the SOS phone into a bus stop? Let's also drive on the left (to match with the Mazda pathplanner)}
\end{auxproof}
  Consider the scenario shown in \cref{fig:pullover}.\footnote{We assume that cars drive on the left, as in Japan, UK and other countries.} Here  \SV{} is initially in Lane~1;
  its goal is to pull over to Lane~3 (the shoulder) at the specified position $\ytgt$.
  There are three principal other vehicles (\POV{}s); two  \POV{}s are in Lane~2
  and the other is in Lane~1. This scenario is relevant to automated emergency stop, an important example of level-4 ADS conditions.



Our aim here is to design an RSS condition $A$ and a proper response $\alpha$ that satisfy \cref{req:reqOnRSSRulesGA}. 
We find that the design of such $(A,\alpha)$ is  harder  than in the collision-avoiding case in \cref{ex:onewayTraffic}. Major challenges include the following.
\begin{itemize}
 \item (Complexity of a scenario) Achieving the ultimate goal (stopping in Lane~3 at $\ytgt$) is achieved by a series of subgoals, such as changing lanes.
 \item (High-level manoeuvre planning) There can be multiple high-level manoeuvre sequences that are feasible. In the current scenario, they are specifically 1) to merge between POV2 and POV1, and 2) to merge after POV1. These will have different corresponding RSS conditions, and we have to systematically compute them.
 \item (Multiple constraints at odds) To merge between POV2 and POV1, \SV{} may need to accelerate in Lane~1, in order to make enough space behind. However, doing so  incurs the risk of
  driving too fast to stop at $\ytgt$ in Lane~3.
 \item (Safety vs.\ goal-achievement) Proper responses should achieve both goal achievement and collision avoidance, which may be at odds as well.   For example, the acceleration discussed above should also take into account the distance from POV3. 
\end{itemize}
  
It is obvious that collision-avoiding RSS rules do not suffice to ensure goal achievement. For example, avoiding collision without an eye to the ultimate goal can trap \SV{} in Lane~1, making it reach the position $\ytgt$ without changing lanes. We experimentally show that this can indeed happen (\cref{sec:exp}). 


\begin{auxproof}
   \todoil{The point about increased journey-time is only relevant if the target
    location is a turn-off. I don't think the example should be an ``emergency''
    anyways, since in this case we can imagine that the whole system is
    operating under a different ``RSS-stage'', where safety (non-collision)
    should be the only concern.}
\end{auxproof}

\end{example}

\subsection{Compositional Derivation of GA-RSS Rules by Program Logic}\label{subsec:introCompDeriv}
Reasoning under the level of complexity in~\cref{ex:pullover} is hardly seen in the existing RSS literature. To address the challenge, in this paper, we propose a structured and compositional approach by the application of \emph{program logic}. 

\begin{figure}[tbp]\centering
 \begin{minipage}{.23\textwidth}
  \centering
\includegraphics[bb=0 0 337 458,clip,scale=.35]{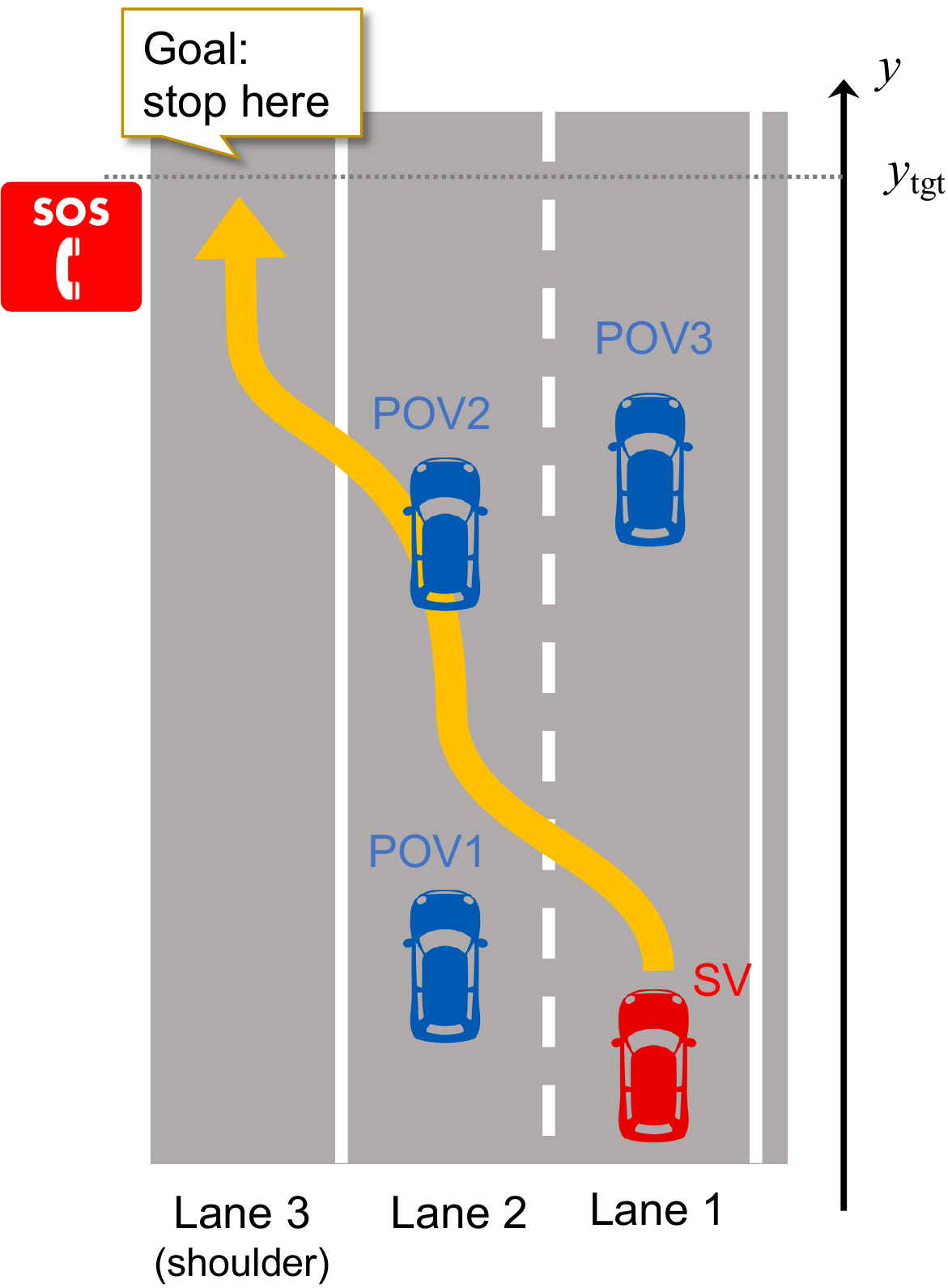}
 \caption{The pull over scenario}
 \label{fig:pullover}
 \end{minipage}
 \hfill
 \begin{minipage}{.23\textwidth}
  \centering
 \includegraphics[bb=75 28 435 465,clip,scale=.35]{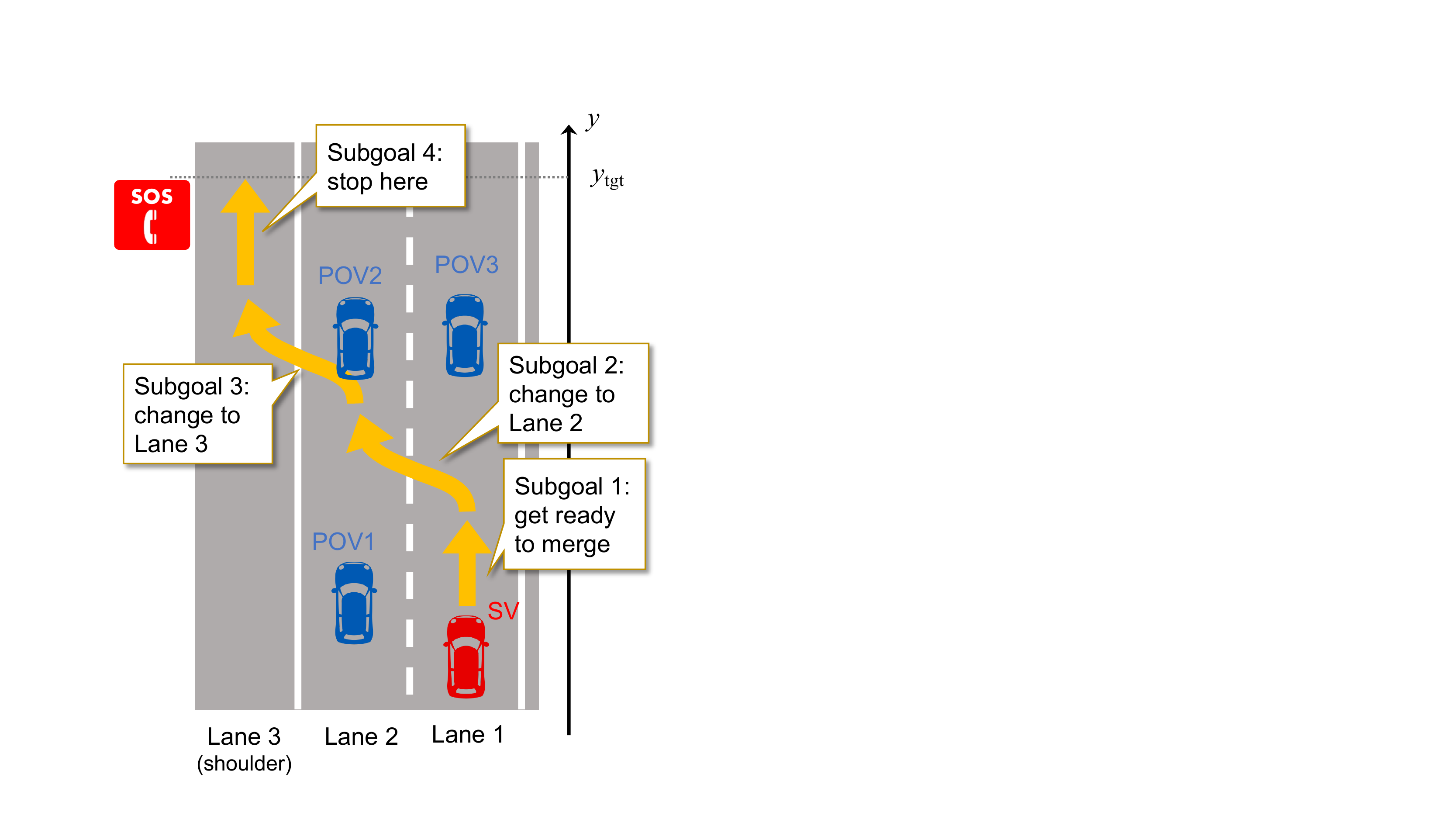}
 \caption{Subgoals in the pull over scenario}
 \label{fig:pulloverSubgoals}
 \end{minipage}
\end{figure}

More specifically, our approach in this work is
\begin{itemize}
 \item firstly to decompose a  scenario into
        \emph{subscenarios}, along  subgoals such  as those  shown in \cref{fig:pulloverSubgoals}, 
 \item to identify proper responses for each subscenario (which is easier since subscenarios are simpler, see~\cref{fig:pulloverSubgoals}),  and
 \item  to identify \emph{preconditions} of those subscenario proper responses so that each precondition guarantee both 1) the goal of the subscenario and 2) the precondition of the subsequent proper response. Here we reason backwards along a sequence of subscenarios (from Subgoal 4 to Subgoal 1 in \cref{fig:pulloverSubgoals}), much like backward predicate transformers in program logic~\cite{Dijkstra75}.
\end{itemize}
A proper response for the whole scenario is then obtained by combining the proper responses for subscenarios; so is the corresponding RSS condition for the whole scenario.

In~\cref{sec:workflow}, we formulate the above workflow in terms of the program logic that we introduce in~\cref{sec:dHL}---the latter is called \emph{differential Floyd--Hoare logic} $\dHL$. The logic $\dHL$ extends classic \emph{Floyd--Hoare logic}~\cite{Hoare69} in 1) accommodation of continuous-time dynamics specified by ODEs and 2) what we call \emph{safety conditions} that must hold all the time during execution. In~\cref{sec:dHL}, we introduce derivation rules for  $\dHL$ that addresses these extensions; we prove their soundness too.

The second extension discussed above (safety conditions) makes the logic $\dHL$ use \emph{Hoare quadruples} $\hquad{S}{A}{\alpha}{B}$, instead of triples $\hoare{A}{\alpha}{B}$ in the original Floyd--Hoare logic. This extension follows the idea formally presented in~\cite{BoerHR97}; see \cref{subsec:relatedAndFutureWork} for further discussions. Explicating a safety condition $S$ allows us to reason simultaneously about  goal achievement (modelled by the postcondition $B$) and collision avoidance (modelled by $S$). 

Our logic $\dHL$ can be seen as a variant of Platzer's \emph{differential dynamic logic $\dL$}~\cite{Platzer18}---we believe that embedding of $\dHL$ in $\dL$ is possible. Among a number of differences, a major one is our choice of the Hoare-style syntax ($\hoare{A}{\alpha}{B}$ with the safety extension $:S$ discussed above) rather than the dynamic logic-style one ($A\Rightarrow [\alpha]B$, as in $\dL$). This syntactic choice fits the purpose of formalising our workflow (\cref{sec:workflow}), where the emphasis is on compositional reasoning along sequential compositions. 


We also discuss software support for the workflow in \cref{sec:toolSupport}. Our current implementation is partially formalised in the sense that 1) rule applications in $\dHL$ are not formalised, but 2) symbolic reasoning about real numbers (such as solving quadratic equations) is formalised in Mathematica. Moreover, the interactive features of Mathematica notebooks are exploited so that even the informal part of reasoning is well-documented and thus trackable. We also discuss prospects of full formalisation.




\subsection{RSS-Supervised Controller: RSS in a Safety Architecture}\label{subsec:introRSSSafetyArchitecture}
We continue \cref{subsec:introRSSUsages} and discuss the usage of (CA- and GA-)RSS that is the most relevant to us, namely in a safety architecture. We use GA-RSS in this way to experimentally demonstrate its significance (\cref{sec:exp}). 



\begin{figure}[tbp]\centering
 \includegraphics[bb=256 199 717 421,clip,width=18em]{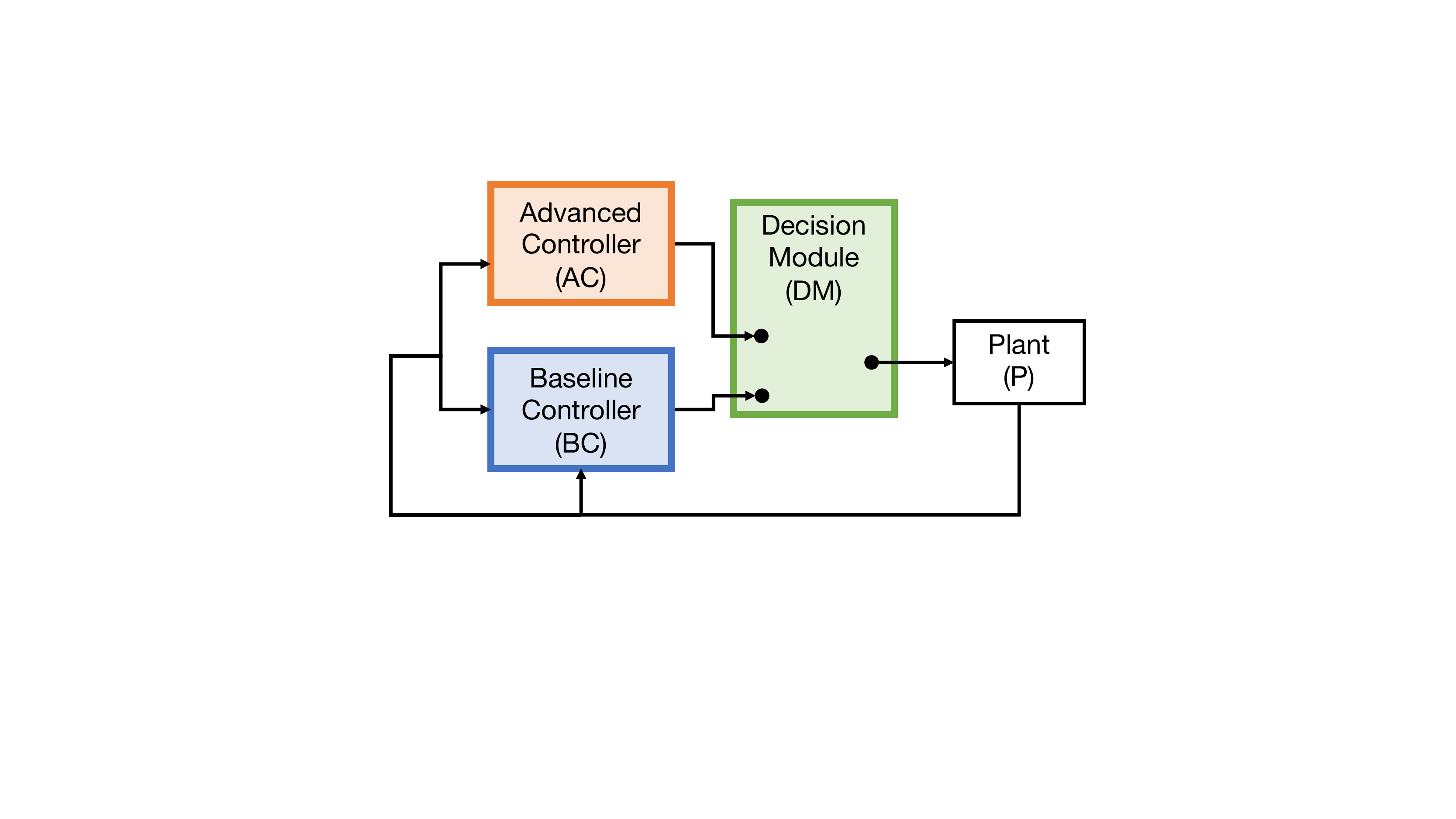}
 \caption{The simplex architecture}
 \label{fig:simplex}
\end{figure}
A prototypical safety architecture is the \emph{simplex architecture} shown in
\cref{fig:simplex}~\cite{CrenshowGRSK07,SetoKSC98}. Here, the \emph{advanced
  controller} (AC) is a complex controller that pursues not only safety but
also other performance measures (such as comfort, progress, and fuel efficiency); the
\emph{baseline controller} (BC) is a simpler controller with a strong emphasis on
safety; and the \emph{decision module} (\DM{}) switches between the two
controllers.  \DM{} tries to use  AC as often as possible for its superior
performance. However, when  \DM{} finds that the current situation is safety
critical, it switches to  BC, whose behaviours are more predictable and easier
to analyse. 

The point of the simplex architecture is that the system's safety can be formally verified even if  AC is a black box. Logically,  \DM{} enforces \emph{contracts} that  AC should respect. The safety of the whole system can then be established by formally reasoning about \DM{}, the plant (P), and BC (that takes over the control in case AC cannot comply with the contracts). 
Specifically, to formally prove that a system guided by a simplex architecture is
  safe, it is enough to show that \BC{} is, and that \DM{} gives
  control to \BC{} soon enough.
  Safety of \AC{} is irrelevant to the proof here. 
This point is especially appealing for ADS, whose AC typically involves a number of learning and optimisation components and thus is very hard to formally analyse.

The components of an RSS rule $(A,\alpha)$   map naturally to the simplex architecture:
\begin{itemize}
  \item \DM{} can be made so that it implements the RSS condition $A$. It uses AC as long as $A$ is robustly satisfied; however, when $A$ is about to be violated, it switches the control  to BC.

  \item  BC can implement the  proper response $\alpha$. Safety of its execution is then  guaranteed by \cref{req:reqOnRSSRules}.
  \item If it happens that  the robust satisfaction of the RSS condition $A$ is restored during BC's execution, \DM{} can switch back from BC to AC.
\end{itemize}
In this way,  a possibly unsafe AC can be made safe, via suitable intervention of 
RSS-based \DM{} and BC. The whole system built this way will be
called an \emph{RSS-supervised controller}.


In~\cref{sec:exp}, we present our implementation of 
\begin{itemize}
 \item the GA-RSS rule for~\cref{ex:pullover} (pull over) that we derive in~\cref{sec:workflow}, and
 \item  the CA-RSS rule in~\cref{ex:onewayTraffic} (one-way traffic), naturally adapted to the multi-lane setting of~\cref{ex:pullover},
\end{itemize}
both in the simplex architecture (AC is a trajectory planner based on sampling and optimisation). Our experiments for the scenario in~\cref{ex:pullover}  demonstrate that the GA-RSS rule indeed achieves the goal safely, while the CA-RSS rule fails to do so.

\subsection{Contributions}
 This paper introduces the idea of \emph{goal-aware RSS} (GA-RSS). We claim that goal-aware RSS rules may be developed for
        complex scenarios, and that they are suited for use in the simplex architecture. These claims are backed up by 
 the following technical contributions; they collectively establish a program logic framework for GA-RSS.
\begin{itemize}
  \item We introduce a program logic $\dHL$ (\emph{differential Floyd--Hoare logic}) as a logical foundation for GA-RSS (\cref{sec:dHL}). It extends classic Floyd--Hoare logic by 1) differential dynamics and 2) safety conditions ($S$ in our \emph{Hoare quadruples} $\hquad{S}{A}{\alpha}{B}$). We introduce derivation rules for $\dHL$ and prove their soundness.

The
main novelty here is dealing with the combination of the two extensions, specifically in the $(\dwhilerule)$ rule in \cref{fig:dFHL-rules}.

  \item We develop a compositional workflow for deriving GA-RSS rules. It is formulated in terms of $\dHL$, exploiting the organising power of the logic. We also describe software support for the workflow by Mathematica.

  \item We run the workflow for the pull over scenario (\cref{ex:pullover}). The resulting GA-RSS rule is implemented in the simplex architecture. Its experimental comparison with 1) no simplex and 2) a CA-RSS rule demonstrates the value of GA-RSS.

\end{itemize}

\subsection{Related  and Future Work}\label{subsec:relatedAndFutureWork}
Much of the related work on RSS has been already discussed. Recent extensions of RSS include a risk-aware one~\cite{OborilS20IV}  and one that allows swerves as evasive manoeuvres~\cite{deIacoSC20IV}. These extensions shall be pursued in our current goal-aware framework. In particular, allowing swerves should be possible, and it will significantly improve the progress of a RSS-supervised controller.

Inclusion of safety conditions in the Floyd--Hoare logic---in the form of Hoare quadruples---is also pursued in~\cite{BoerHR97}, in the context of verification of concurrent systems. Our logic $\dHL$ combines the idea with the machinery of $\dL$~\cite{Platzer18} for handling continuous dynamics. In particular, our main technical novelty---namely  an inference rule for continuous dynamics and safety (\cref{subsec:dHLRules})---does not appear in~\cite{Platzer18,BoerHR97}.

Some RSS rules have been implemented and are offered as a library~\cite{GassmannOBLYEAA19IV}. Integration of the goal-aware RSS rules we derive in this paper, in the library, is future work. One advantage of doing so is that the GA-RSS rules will then accommodate varying road shapes.


This paper studies logical derivation of GA-RSS rules, with a prospect of \emph{fully formal} derivation
 (see~\cref{subsec:towardsFullFormalization}). The problem of formally verifying correctness of RSS rules is formulated and investigated in~\cite{RoohiKWSL18arxiv}. Their formulation is based on a rigorous notion of signal; they argue that none of the existing \emph{automated} verification tools is suited for the verification problem. This concurs with our experience so far---in particular, formal treatment of other participants' responsibilities (in the RSS sense) seems to require human intervention. At the same time, in our preliminary manual verification experience in \KeYmaeraX, we see a lot of automation opportunities. Developing proof tactics dedicated to those will ease manual verification efforts.

An idea similar to that of RSS-supervised controllers (\cref{subsec:introRSSSafetyArchitecture}) is found in~\cite{BaheriNKGTF20IV}.  BC in~\cite{BaheriNKGTF20IV} is a learning-based controller that is realised by an RNN and is trained to follow given safety rules. This is unlike our RSS-based BC that executes explicitly RSS proper responses. There is no statistical learning, hence no uncertainties from black-box learning, in our BC.

In~\cite{PekA18IROS}, a rigorous guarantee of ADS safety is pursued via the notion of \emph{invariably safe set}. The latter is defined in terms of backward reachability analysis, and in that sense, the work is similar to RSS and the current work. The biggest difference is that the approach in~\cite{PekA18IROS} is about \emph{runtime} and \emph{numeric} verification while RSS is about \emph{static}, \emph{a priori} and \emph{symbolic} rules. Consequently, many usages of RSS discussed in \cref{subsec:introRSSUsages} do not apply to~\cite{PekA18IROS}.
\begin{auxproof}
 Other major difference between~\cite{PekA18IROS} and RSS include the following.
 \begin{itemize}
 \item Their method
 determines safety of \emph{given} control strategies, while  RSS  \emph{provides} concrete control strategies in the form of proper responses.
 \item   Other traffic participants' predicted motion is input to their method that must be computed elsewhere. In contrast, in RSS, bounds of the others' future motion is part of the RSS framework---they are derived from the RSS responsibility principles (\cref{subsec:introCARSS}). 
 \end{itemize}
\end{auxproof}
Moreover, the symbolic nature of RSS is what allows compositional derivation of rules, the key contribution of this work. At the same time, the numeric and online nature of~\cite{PekA18IROS} will probably yield less conservative control actions. Overall, it seems that the two works target at different classes of driving situations:~\cite{PekA18IROS} for urban scenarios (less structure, shorter-term control); this work is for highway scenarios (more structure, longer-term control).

Formal (logical, deductive) verification of ADS safety is also pursued in~\cite{RizaldiISA18} using the interactive theorem prover Isabelle/HOL~\cite{NipkowPW02}. The work uses a white-box model of a controller, and a controller must be very simple. This is unlike RSS and the current work, which allows black-box \AC{}s and thus accommodates various real-world controllers such as sampling-based path planners (\cref{subsec:introRSSSafetyArchitecture}).

In the presence of perceptual uncertainties (such as  errors in position measurement and object recognition), it becomes harder for  \BC{}s and \DM{}s to ensure safety. Making  \BC{}s tolerant of perceptual uncertainties is pursued in~\cite{SalayCEASW20PURSS,DBLP:conf/nfm/KobayashiSHCIK21}. One way to adapt \DM{}s is to enrich their input so that they can better detect  potential hazards. Feeding DNNs' confidence scores is proposed in~\cite{AngusCS19arxiv};  in~\cite{ChowRWGJALKC20}, it is proposed for \DM{}s to look at inconsistencies between perceptual data of different modes.



\subsection{Organisation of the Paper}
In~\cref{sec:dHL}, we introduce our program logic $\dHL$, introducing its syntax, semantics, and derivation rules. We prove the soundness of the derivation rules, too (\cref{thm:dHLsoundness}). In~\cref{sec:problemFormulation}, we formulate our problem of deriving GA-RSS rules, based on the mathematical notion of driving scenario that we also introduce there. Our workflow for compositional derivation of GA-RSS rules is presented in~\cref{sec:workflow}, where our main theorem is the correctness of the workflow (\cref{thm:correctness}) assuming the correctness in each subscenario (the condition~\cref{eq:backPropAssignmentCond}). We use the pull over scenario (\cref{ex:pullover}) as a leading example, and derive a GA-RSS rule for it. Software support for the workflow is discussed in~\cref{sec:toolSupport} (the current partially formal one and the prospects of full formalisation).  Our experimental evaluation is discussed in~\cref{sec:exp}, where our implementation of the GA-RSS rules for \cref{ex:pullover} in the simplex architecture is compared with those without BC and with CA-RSS. In~\cref{sec:concl} we conclude.


\section{Differential Floyd-Hoare Logic $\dHL$}\label{sec:dHL}

\subsection{The Syntax of $\dHL$: Assertions, Hybrid Programs, and Hoare Quadruples}

\begin{notation}\label{notation:integerInterval}
In this paper, we let $[1,N]$ denote the set $\{1,2,\dotsc, N\}$ of integers, where $N$ is a positive integer.
\end{notation}

\subsubsection{Overview of the Syntax}

In this section, we describe three ingredients to formalise our rules:
assertions, hybrid programs, and Hoare quadruples.
Let us give some intuition before we delve into formal definitions.

\emph{Assertions} are logical
objects describing qualitative properties of states.
For example, in the one-way traffic scenario of
Example~\ref{ex:onewayTraffic}, if we denote by $y_f$ and $v_f$ the
position and velocity of the front car and $y_r$ and $v_r$ those of
the rear car, we will write an assertion $\neg(v_f = 0 \wedge v_r = 0) \wedge y_r <
y_f$ to describe the configurations where at least one car is not
stopped and for which there is no collision.

\emph{Hybrid programs} are a combination of usual programs of
imperative languages (such as IMP~\cite{Winskel93}) and differential equations that express continuous dynamics.
Our syntax therefore contains
assignments, if-branchings, etc., but also constructs allowing the state to
change following the solutions of differential equations. The terminology
``hybrid program'' comes from differential dynamic logic~\cite{Platzer18},
which uses a slightly different syntax, but to which our syntax can be
translated.

Finally, \emph{Hoare quadruples} 
\begin{math}
 \hquad{\safetya}{\asserta}{\coma}{\assertb}
\end{math}
relate both assertions and hybrid
programs to formally specify and prove correctness of the latter. 
 The traditional
 Floyd--Hoare logic~\cite{Hoare69} uses \emph{Hoare triples}  $\hoare{\asserta}{\coma}{\assertb}$ that roughly means the truth of a \emph{precondition} $A$ guarantees the truth of a \emph{postcondition} $B$ after the execution of a program $\coma$. 
In $\dHL$, following~\cite{BoerHR97}, we extend the above classic syntax and write
\begin{equation*}
	\hquad{\safetya}{\asserta}{\coma}{\assertb}
\end{equation*}
with the intention that
\begin{itemize}
 \item  every execution of the hybrid program $\alpha$,
 if it  starts from
a state satisfying the assertion $A$ (the precondition),
 \item terminates in a
state satisfying the assertion $B$ (the postcondition),  and
 \item moreover, respects the
assertion $S$ (the \emph{safety condition}) at all times during the execution.
\end{itemize}
The addition of a safety condition $S$ allows us to reason about collision avoidance in RSS, while the goal of a scenario is naturally modelled as a postcondition. Later in~\cref{sec:problemFormulation},  driving scenarios and GA-RSS rules (cf.\ \cref{subsec:goalAwareRSS}) are modelled as components of Hoare quadruples.


Assertions, hybrid programs, and Hoare quadruples form the syntax of
\emph{differential Floyd-Hoare logic} ($\dHL$ for short).

\subsubsection{Formal Definition}
We formally define $\dHL$. An example is in \cref{ex:onewaydHL}.

\begin{definition}[($\dHL$) assertions]
  A \emph{term} is a rational polynomial on a fixed infinite set  $\Variables$ of
  variables.
  \emph{$\dHL$ assertions} are generated by the grammar
  \[A,B\; ::=\; \true \mid e \sim f \mid A \land B \mid A \lor B \mid \lnot A \mid A \limply B\]
  where $e$, $f$ are terms and
  $\sim \ \in \set{=, \leq, <, \neq}$.
\end{definition}
  A $\dHL$ assertion can be \emph{open} or \emph{closed} (or both, or none).
  \emph{Openness} and \emph{closedness} are defined
  recursively: $\true$ is both open and closed, $e < f$ and $e \neq f$
  are open, $e \leq f$ and $e = f$ are closed, $A \land B$ and $A \lor
  B$ are open (resp.\ closed) if both components are, $\neg A$ is open
  (resp.\ closed) if $A$ is closed (resp.\ open), and $A \limply B$ is
  open (resp.\ closed) if $A$ is closed and $B$ open (resp.\ $A$ open
  and $B$ closed).
Note that open $\dHL$ assertions describe open subsets of $\R^\Variables$.
\begin{definition}\label{def:hybridPrograms}
  \emph{Hybrid programs} (or \emph{$\dHL$ programs}) are given by the
  syntax:
  \begin{align*}
  	\alpha,\beta \quad ::=\quad
  		& \skipClause{} \mid
        \alpha;\beta \mid
        \assignClause{\var}{\term} \mid
        \ifThenElse{\asserta}{\alpha}{\beta} \mid \\
      & \whileClause{\asserta}{\alpha} \mid
        \dwhileClause{\asserta}{\odeClause{\vars}{\funs}}.
  \end{align*}
  We sometimes drop the braces in
  $\dwhileClause{\asserta}{\odeClause{\vars}{\funs}}$ for readability.
  In $\dwhileClause{\asserta}{\odeClause{\vars}{\funs}}$, $\vars$ and
  $\funs$ are lists of the same length, respectively of (distinct)
  variables and terms, and $\asserta$ is open.
\end{definition}
All constructs are usual ones from imperative programming, except for the
\emph{differential while} construct $\dwhileKeyword$.
It encodes the differential dynamics: $\odeClause{\vars}{\funs}$ denotes a
system of differential equations, and
$\dwhileClause{\asserta}{\odeClause{\vars}{\funs}}$ denotes a dynamical system
following the differential equations until the condition $\asserta$ is
falsified.
Openness of $\asserta$ ensures that, if $\asserta$ is falsified at some point, then
there is the smallest time $t_{0}$ when it is falsified, and the system
follows the dynamics for time $t_{0}$.

\begin{remark}
  \label{rem:term_extension}
  It is possible to extend the language of terms, by allowing more
  functions than just polynomials.
  In that case, the syntax $\dot{\vars} = \funs$ is only allowed when
  $\funs$ is locally Lipschitz continuous to ensure existence and
  uniqueness of solutions, by the Picard-Lindel\"{o}f theorem.
  One should also make sure that any term of the syntax possesses
  partial derivatives with respect to all variables in order to use
  the rules of Section~\ref{subsec:dHLRules}.
\end{remark}

Our programming language syntax (\cref{def:hybridPrograms}) is inspired by that in $\dL$~\cite{Platzer18}, but comes with
significant changes.
It is imperative and deterministic (see
Lemma~\ref{lem:confluence}), which makes it easier to use for
practitioners, while expressive enough to encode interesting models.
This also makes it more suited to Hoare logic and total correctness,
which is crucial for applications to automated driving.

%
%
%
%

\begin{figure*}
  \small
  \begin{mathpar}
    \bottomAlignProof
    \AxiomC{ }
    \UnaryInfC{$\state{\seqClause{\skipClause}{\comb}}{\store} \red{0}
      \state{\comb}{\store}$}
    \DisplayProof

    \bottomAlignProof
    \AxiomC{$\state{\coma}{\store} \red{t} \state{\coma'}{\store'}$}
    \UnaryInfC{$\state{\seqClause{\coma}{\comb}}{\store} \red{t}
      \state{\seqClause{\coma'}{\comb}}{\store'}$}
    \DisplayProof

    \bottomAlignProof
    \AxiomC{ }
    \UnaryInfC{$\state{\assignClause{\var}{\expa}}{\store} \red{0}
      \state{\skipClause}{\store[\var \to \sem{\expa}{\store}]}$}
    \DisplayProof

    \bottomAlignProof
    \AxiomC{$\store \vDash \asserta$}
    \UnaryInfC{$\state{\ifThenElse{\asserta}{\coma}{\comb}}{\store} \red{0}
      \state{\coma}{\store}$}
    \DisplayProof

    \bottomAlignProof
    \AxiomC{$\store \nvDash \asserta$}
    \UnaryInfC{$\state{\ifThenElse{\asserta}{\coma}{\comb}}{\store} \red{0}
      \state{\comb}{\store}$}
    \DisplayProof

    \bottomAlignProof
    \AxiomC{$\store \nvDash \asserta$}
    \UnaryInfC{$\state{\whileClause{\asserta}{\coma}}{\store} \red{0}
      \state{\skipClause}{\store}$}
    \DisplayProof

    \bottomAlignProof
    \AxiomC{$\store \vDash \asserta$}
    \UnaryInfC{$\state{\whileClause{\asserta}{\coma}}{\store} \red{0}
      \state{\seqClause{\coma}{\whileClause{\asserta}{\coma}}}{\store}$}
    \DisplayProof

    \bottomAlignProof
    \AxiomC{$
      t \geq 0 \quad
      \hat{\var}(0) = \store \quad
      \frac{d\hat{\var}}{dt}(t) = \sem{\funs}{\hat{\var}(t)} \quad
      \store' = \hat{\var}(t) \quad
      \forall t' \leq t.\, \hat{\var}(t') \vDash \asserta
    $}
    \RightLabel{$(*)$}
    \UnaryInfC{$\state{\dwhileClause{\asserta}{\odeClause{\vars}{\funs}}}{\store} \red{t}
      \state{\dwhileClause{\asserta}{\odeClause{\vars}{\funs}}}{\store'}$}
    \DisplayProof

    \bottomAlignProof
    \AxiomC{$
      t \geq 0 \quad
      \hat{x}(0) = \store \quad
      \frac{d\hat{\var}}{dt}(t) = \sem{\funs}{\hat{\var}(t)} \quad
      \store' = \hat{\var}(t) \quad
      \forall t' < t.\, \hat{\var}(t') \vDash \asserta \quad
      \hat{\var}(t) \nvDash \asserta
    $}
    \RightLabel{$(*)$}
    \UnaryInfC{$\state{\dwhileClause{\asserta}{\odeClause{\vars}{\funs}}}{\store} \red{t}
      \state{\skipClause}{\store'}$}
    \DisplayProof
  \end{mathpar}
  \caption{Reduction relation $\rightarrow$ on states; rules annotated with $(*)$
    are discussed in \cref{rem:reduction}}
  \label{fig:reduction}
\end{figure*}

We define an operational semantics for our syntax:
\begin{definition}[semantics]
  A \emph{store} is a function from variables to reals.
  \emph{Store update} is denoted $\update{\store}{\var}{\val}$;
  it maps $\var$ to $\val$ and any other variable $\var'$ to
  $\store(\var')$.
  The \emph{value} $\sem{\term}{\store}$ of a term $\term$ in a store
  $\store$ is a real defined as usual by induction on $e$ (see for
  example~\cite[Section~2.2]{Winskel93}).
  The \emph{satisfaction} relation between stores $\rho$ and
  $\dHL$ assertions $A$, denoted $\store \vDash A$, is also defined as usual
  (see~\cite[Section~2.3]{Winskel93}).

  A \emph{state} is a pair $\state{\coma}{\store}$ of a hybrid program
  and a store.
  The \emph{reduction} relation on states is defined in
  Figure~\ref{fig:reduction}.
  A state $s$ \emph{reduces} to $\statea'$ if $\statea \red{}^*
  \statea'$, where $\red{}^*$ is the reflexive transitive closure of
  $\red{}$.
  A state $\statea$ \emph{converges} to $\store$, denoted
  $\converge{\statea}{\store}$, if there exists a reduction sequence
  $\statea \red{}^* \state{\skipClause}{\store}$.
\end{definition}
Let us explain how to read Figure~\ref{fig:reduction}: hypotheses are
listed above the horizontal line, and the conclusion below it.
For example, $\state{\seqClause{\skipClause}{\comb}}{\store}$ can
always reduce to $\state{\comb}{\store}$ (there are no hypotheses),
and if $\state{\coma}{\store}$ reduces to $\state{\coma'}{\store'}$,
then $\state{\seqClause{\coma}{\comb}}{\store}$ reduces to
$\state{\seqClause{\coma'}{\comb}}{\store'}$.
\begin{remark}
  \label{rem:reduction}
  In the reduction rules for $\dwhileKeyword$, $\hat{x}$ is the global
  solution to the differential equation $\dot{\vars} = \funs$  with
  initial condition $\hat{x}(0) = \store$.
  As a side condition (left untold for readability in the rule), we
  assume that all variables not mentioned in $\vars$ are left
  untouched during the transition, so if $y$ is not in $\vars$, then
  $\store'(y) = \store(y)$.
\end{remark}

Convergence $\convergeSymbol$ corresponds to complete executions of
programs (until termination), while reduction $\red{}^*$ corresponds
to potentially partial executions, which can reach any intermediate state of the
computation.
\begin{example}
  \label{ex:reduction}
  The state $\state{\coma}{\store}$, where
  \[
    \coma\; =\; \left(\,\dwhileClause{\var > 0}{\odeClause{\var}{-1}} \;; \quad
    \assignClause{\var}{\var-1}\right)\rlap{,}
  \]
  and $\store(\var) = 2$, can reduce
  \begin{itemize}
    \item to $\state{\coma}{\update{\store}{\var}{v}}$ for any $v \in
      (0,2]$,
    \item to
      $\state{\assignClause{\var}{\var-1}}{\update{\store}{\var}{0}}$,
    \item and to $\state{\skipClause}{\update{\store}{\var}{-1}}$, 
  \end{itemize}
  but only the last one corresponds to convergence (namely $\converge{\state{\coma}{\rho}}{\update{\store}{\var}{-1}}$).
\end{example}

\begin{lemma}[confluence]
  Our language of hybrid programs is confluent.
  That is, if $s \red{}^* s_1$ and $s \red{}^* s_2$, then there exists
  $s'$ such that $s_1 \red{}^* s'$ and $s_2 \red{}^* s'$.
  In particular, if $\converge{\statea}{\store}$, $s$ cannot converge
  to any other store $\store' \neq \store$.
  \label{lem:confluence}
\end{lemma}
Confluence basically means that the language is deterministic in the
sense that, no matter the reduction sequence, a program always
converges to the same value.
This holds because reduction in our language is mainly deterministic,
except for the $\dwhileKeyword$ rules, in which case the reduction
that has run for the smaller amount of time can be reduced again to
catch up with the other reduction.

\begin{auxproof}
\todoil{Let's move these proofs to the appendix. The T-IV reviewers will be horrified by them.}
\begin{proof}
  The reduction relation $\red{}^*$ can be equipped with pairs $(t,n)
  \in \R \times \N$, where $t$ represents time spent in the continuous
  $\dwhileKeyword$ steps, and $n$ the number of discrete steps
  executed after the last $\dwhileKeyword$ step.
  If $s \redext{t_i,n_i}^* s_i$ for $i \in \set{1,2}$ with $(t_1,n_1)
  < (t_2,n_2)$ for the lexicographic order, then $s_1 \red{}^* s_2$.
  The only non-obvious point is when two sequences of $\dwhileKeyword$
  rules are applied for different total amounts of time $t_1 < t_2$,
  in which case, necessarily $n_1 = 0$, and the $\dwhileKeyword$ rule
  can be applied on $s_1$ to catch up with $s_2$'s reduction (the
  systems follow the same path by the Picard-Lindel\"{o}f theorem),
  and then the rest of $s_2$'s reduction sequence can be appended to
  $s_1$'s.
\end{proof}
\end{auxproof}

Finally, we define validity of Hoare quadruples:
\begin{definition}[Hoare quadruples]\label{def:hoareQuad}
  A \emph{Hoare quadruple} is a quadruple
  $\hquad{\safetya}{\asserta}{\coma}{\assertb}$ of three $\dHL$ assertions
  $\asserta$, $\assertb$, and $\safetya$, and a hybrid program
  $\coma$.
  It is \emph{valid} if, for all stores $\store$ such that $\store
  \vDash \asserta$,
  \begin{itemize}
    \item there exists $\store'$ such that
      $\converge{\state{\coma}{\store}}{\store'}$ and $\store' \vDash
      \assertb$, and
    \item for all reduction sequences $\state{\coma}{\store} \red{}^*
      \state{\comb}{\store'}$, $\store' \vDash \safetya$.
  \end{itemize}
\end{definition}
Hoare quadruples have safety conditions $S$ in addition to the usual components of Hoare triples. They are required to specify safety properties, which must hold at all
times.
In traditional programming, one is usually only interested in
input-output behaviours: as long as a program returns a valid value,
it does not matter which intermediate states it went through.
In contrast, the intermediate states matter in our case, since there
may be a collision or safety violation halfway through an execution
that reaches the desired target.

The safety of all intermediate states is ensured by the definition of
$\statea \to \statea'$.
The interesting case is that of the differential dynamics, where the
dynamics can be stopped at any point in time, and thus $\statea'$ can
be the state reached at any point of the dynamics.

Also note that this semantics is \emph{total correctness}, rather than
\emph{partial correctness}.
A Hoare triple $\hoare{\asserta}{\coma}{\assertb}$ is valid for
partial correctness if, roughly, any terminating execution of $\coma$ under the precondition $\asserta$ satisfies the postcondition $\assertb$. In particular, if $\coma$ is not terminating, then the Hoare triple is trivially true, regardless of the truth of $B$. This is not desired since we want to ensure goal achievement (modelled by the postcondition $B$).  In contrast,  total correctness additionally requires the existence of a terminating execution, which suits our purpose.
 See~\cite{Winskel93} for more details on partial and total correctness. 

\begin{figure}\centering
\begin{math}\footnotesize
   \begin{array}{ll}
    \textnormal{1}\quad & \assignClause{t}{0}\ ; \\
     \textnormal{2}\quad & \dwhileClause{v_f > 0 \land t < \rho}{\boxed{\;\dynamics{f}{}\;}\;,\;
                          \boxed{\;\dynamics{r}{1}\;}}\ ; \\
    \textnormal{3}\quad & \ifHeader{v_f = 0}
                          \left[ \dwhileClauseNb{t < \rho}{\boxed{\;\dynamics{r}{1}\;}}\ ;
                          \dwhileClauseNb{v_r > 0}{\boxed{\;\dynamics{r}{2}\;}} \,\right] \\
    \textnormal{4}\quad & \elseKeyword \\
    \textnormal{5}\quad & \quad \dwhileClause{v_f > 0 \land v_r >
                          0}{\boxed{\;\dynamics{f}{}\;}\;,\; \boxed{\;\dynamics{r}{2}\;}}\ ; \\
    \textnormal{6}\quad & \quad \ifHeader{v_f = 0}\, \dwhileClauseNb{v_r >
                          0}{\boxed{\;\dynamics{r}{2}\;}} \\
    \textnormal{7}\quad & \quad \elseKeyword\ \dwhileClauseNb{v_f >
                          0}{\boxed{\;\dynamics{f}{}\;}}
  \end{array}
\end{math}  
\caption{Hybrid program $\coma$ for the one-way traffic scenario (\cref{ex:onewaydHL})}
  \label{fig:system-one-way-new}
\end{figure}

\begin{example}[the one-way traffic scenario]\label{ex:onewaydHL}
  The scenario for \cref{ex:onewayTraffic} can be modelled in $\dHL$. 
  We start by modelling the dynamics of the different agents involved
  in the scenario (the front and rear cars) as a hybrid program
  $\coma$ (see \cref{fig:system-one-way-new}).
  We then model the property that we want to show (namely, that the
  cars can stop without colliding if they are far enough apart) into
  a Hoare quadruple (see~\cref{eq:oneway-quad}). 
  In what follows we explain the modelling ($\coma$ in \cref{fig:system-one-way-new} and the  Hoare quadruple in~\cref{eq:oneway-quad}). 
  We defer the proof of validity of~\cref{eq:oneway-quad} to
  \cref{ex:oneway-proof}.

In this scenario, we aim to show 
  that whatever the front car is doing, the rear car can properly respond without colliding with the front car,
  as long as it respects the RSS safety distance ($\dRSS(v_{f}, v_{r})$ in~\cref{eq:RSSMinDist}).

The worst case 
 is when the front car breaks at the
  maximal braking rate $\bmax$, while the rear car is accelerating with the maximal acceleration rate
  $\amax$ during the reaction time $\rho$ before engaging the proper response $\alpha$ (namely decelerating with the maximal comfortable
  braking rate $\bmin$).
  If we denote by $y_f$ and $v_f$ the position and velocity of the
  front car and by $y_r$ and $v_r$ those of the rear car, then these
  behaviours of the cars can be written as the hybrid programs
  \begin{align*}
    \coma_f \ &= \left( \dwhileClauseNb{v_f > 0}{\boxed{\;\dynamics{f}{}\;}} \, \right) \rlap{,} \\
    \coma_r \ &= \left( \assignClause{t}{0}; \dwhileClauseNb{t <
      \rho}{\boxed{\;\dynamics{r}{1}\;}}\ ; \dwhileClauseNb{v_r >
      0}{\boxed{\;\dynamics{r}{2}\;}} \, \right) \rlap{,}
  \end{align*}
  where the $\dynamics{}{}$'s represent the dynamics of each car:
  \begin{equation}\label{eq:defDeltas}
\begin{aligned}
     \dynamics{f}{} \ &= \ \left( \dot{y_f} = v_f, \dot{v_f} = -\bmax \right) \rlap{,} \\
    \dynamics{r}{1} \ &= \ \left( \dot{y_r} = v_r, \dot{v_r} = \amax, \dot{t} = 1 \right) \rlap{,}
      \\
    \dynamics{r}{2} \ &= \ \left( \dot{y_r} = v_r, \dot{v_r} = -\bmin \right) \rlap{.}
\end{aligned}  
\end{equation}
  By manually combining the two hybrid programs $\coma_f, \coma_r$---letting them run in parallel---we obtain the
  hybrid program $\coma$ in Figure~\ref{fig:system-one-way-new}.
  
  This hybrid program $\coma$ models the situation in which both the front and
  rear cars are following their worst case behaviours as long as they
  are not both stopped.
  The logical structure of $\coma$ enumerates all the different states
  the scenario can be in: whether the front car
  has stopped braking or not, and whether the rear car is still
  accelerating, has engaged the proper response, or finished braking.
  For example, the $\dwhileKeyword$ on
  Line~2 of $\coma$ in \cref{fig:system-one-way-new}
  corresponds to a state where the
  front car is braking and the rear car is still accelerating, and the
  $\ifKeyword$ on Line~3 corresponds to the case where the front car
  has stopped braking before the rear car starts engaging the proper
  response.

 We can then model the whole scenario as the following Hoare quadruple. 
  \begin{equation}
    \left\{
      \begin{array}{l}
        v_r \geq 0 \land v_f \geq 0 \land {} \\
        y_f - y_r > \dRSS(v_f,v_r)
      \end{array}
    \right\}
    \begin{array}[t]{lr}
      \coma & \hspace{-4pt} \{ v_r = 0 \land v_f = 0 \} \\[+.3em]
      & {} : y_r < y_f \rlap{.}
    \end{array}
    \label{eq:oneway-quad}
  \end{equation}


  The postcondition states that both cars have stopped, while the
  precondition models the situations in which we want to prove that
  there is no collision (namely, when the cars are farther than an RSS
  safety distance apart).
  The safety condition $y_r < y_f$ models the fact that there is no
  collision along the dynamics.
  We will prove that this $\dHL$ quadruple is valid, using
  derivation rules for $\dHL$, in \cref{ex:oneway-proof}.
\end{example}

\subsection{Derivation Rules in $\dHL$}\label{subsec:dHLRules}

\begin{figure*}
  \small
  \begin{mathpar}
    \bottomAlignProof
    \AxiomC{ }
    \RightLabel{(\skipcrule)}
    \UnaryInfC{$\hquad{\asserta}{\asserta}{\skipClause}{\asserta}$}
    \DisplayProof

    \bottomAlignProof
    \AxiomC{$\hquad{\safetya}{\asserta}{\coma}{\assertb}$}
    \AxiomC{$\hquad{\safetya}{\assertb}{\comb}{\assertc}$}
    \RightLabel{(\seqrule)}
    \BinaryInfC{$\hquad{\safetya}{\asserta}{\coma ; \comb}{\assertc}$}
    \DisplayProof

    \bottomAlignProof
    \AxiomC{ }
    \RightLabel{(\assignrule)}
    \UnaryInfC{$\hquad{\asserta \lor \subst{\asserta}{\expa}{\var}}{\subst{\asserta}{\expa}{\var}}{\assignClause{\var}{\expa}}{\asserta}$}
    \DisplayProof

    \bottomAlignProof
    \AxiomC{$\begin{array}{c}
      \hquad{\safetya}{\asserta \land \assertb}{\coma}{\assertc} \\
      \hquad{\safetya}{\neg \asserta \land \assertb}{\comb}{\assertc}
      \end{array}$}
    \RightLabel{(\ifrule)}
    \UnaryInfC{$\hquad{\safetya}{\assertb}{\ifThenElse{\asserta}{\coma}{\comb}}{\assertc}$}
    \DisplayProof

    \bottomAlignProof
    \AxiomC{$\hquad{\safetya}{\asserta \land \assertb \land \variant \gtrsim 0 \land \variant = \var}{\coma}{\assertb \land \variant \gtrsim 0 \land \variant \leq \var - 1}$}
    \RightLabel{(\whilerule)$^\dagger$}
    \UnaryInfC{$\hquad{\safetya}{\assertb \land \variant \gtrsim 0}{\whileClause{\asserta}{\coma}}{\neg \asserta \land \assertb \land \variant \gtrsim 0}$}
    \DisplayProof

    \bottomAlignProof
    \AxiomC{$\begin{array}{rlll}
        \mathsf{inv\colon} & \asserta \Rightarrow \invariant \sim 0 & \variant \geq 0 \land \invariant \sim 0 \Rightarrow \lieder{\vars}{\funs}{\invariant} \simeq 0\\
        \mathsf{var\colon} & \asserta \Rightarrow \variant \geq 0 & \variant \geq 0 \land \invariant \sim 0 \Rightarrow \lieder{\vars}{\funs}{\variant} \leq \terminator\\
        \mathsf{ter\colon} & \asserta \Rightarrow \terminator < 0 & \variant \geq 0 \land \invariant \sim 0 \Rightarrow \lieder{\vars}{\funs}{\terminator} \leq 0\\
      \end{array}$}
    \RightLabel{(\dwhilerule)$^\dagger$}
    \UnaryInfC{$\hquad{\invariant \sim 0 \land \variant \geq 0}{\asserta}{\dwhileClauseNb{\variant > 0}{\odeClause{\vars}{\funs}}}{\variant = 0 \land \invariant \sim 0}$}
    \DisplayProof

    \bottomAlignProof
    \AxiomC{$\begin{array}{@{}cc@{}}
        & \asserta \limply \asserta' \\
        \hquad{\safetya'}{\asserta'}{\coma}{\assertb'} & \safetya' \land \assertb' \limply \assertb \\
        & \safetya' \limply \safetya
      \end{array}$}
    \RightLabel{(\limprule)}
    \UnaryInfC{$\hquad{\safetya}{\asserta}{\coma}{\assertb}$}
    \DisplayProof

    \bottomAlignProof
    \AxiomC{$\hquad{\safetya}{\asserta}{\coma}{\assertb}$}
    \AxiomC{$\hquad{\safetya'}{\asserta}{\coma}{\assertb'}$}
    \RightLabel{(\conjrule)}
    \BinaryInfC{$\hquad{\safetya \land \safetya'}{\asserta}{\coma}{\assertb \land \assertb'}$}
    \DisplayProof

    \bottomAlignProof
    \AxiomC{$\asserta_0 \limply (\exists t \geq 0.~\assertc_{{<}t}
      \land \neg \assertc_t \land \assertb_t \land \safetya_{\leq t})$}
    \RightLabel{(\dwhilesolrule)}
    \UnaryInfC{$\hquad{\safetya}{\asserta}{\dwhileClauseNb{\assertc}{\odeClause{\vars}{\funs}}}{\assertb}$}
    \DisplayProof

  \end{mathpar}
  \caption{\dHL{} rules for total correctness; rules with $^\dagger$
  have side conditions discussed in Assumption~\ref{assum:dHL-rules}.
  Rule $(\dwhilesolrule)$ is discussed separately in \cref{subsubsec:dwhilesol}.}
  \label{fig:dFHL-rules}
\end{figure*}

We present a set of rules to derive valid Hoare quadruples, listed in
Figure~\ref{fig:dFHL-rules}.
Like the rules in Figure~\ref{fig:reduction}, hypotheses are listed
above the horizontal line, and the conclusion below it.
For example, the $(\seqrule)$ rule can be read as: if
$\hquad{\safetya}{\asserta}{\coma}{\assertb}$ and
$\hquad{\safetya}{\assertb}{\comb}{\assertc}$ are provable in $\dHL{}$,
then so is $\hquad{\safetya}{\asserta}{\coma ; \comb}{\assertc}$.

\begin{assumption}
  \label{assum:dHL-rules}
  In the $(\whilerule)$ rule, $\gtrsim \ \in \set{>,\geq}$ (meaning that all the occurrences of $\gtrsim$ should be replaced with the same $>$ or $\geq$) and $\var$
  is a fresh variable.
  In the $(\dwhilerule)$ rule, $(\sim, \simeq) \in \set{(=,=),
  (>,\geq), (\geq, \geq)}$, and the dynamics $\odeClause{\vars}{\funs}$
  is assumed to have a global solution.
\end{assumption}
  \todooptil{(optional) Clovis: next sentence could be rephrased if we have a good
  idea.}
  Note that the existence of global solutions is not a constraint in
  practice, since their non-existence would imply that some physical
  quantity diverges to infinity, which is impossible in a physical
  system.

Some hypotheses of the $(\dwhilerule)$ and $(\limprule)$ rules are
$\dHL$ assertions, by which we mean that these assertions must be
\emph{valid}, that is, satisfied by all stores.
For example, the precondition $\asserta \limply \variant \geq 0$ means
that, for any $\store$, if $\store \vDash \asserta$, then
$\sem{\variant}{\store} \geq 0$.


Most of the rules in Figure~\ref{fig:dFHL-rules} are standard Hoare
logic rules when stripped of their safety conditions, so we only
discuss the exceptions: $(\whilerule)$, $(\dwhilerule)$ and $(\dwhilesolrule)$.

\subsubsection{The $(\whilerule)$ Rule}
The first exception is the $(\whilerule)$ rule, which is an
alternative used for \emph{total} correctness (similar for example to the one
found in~\cite{DBLP:conf/fase/HuismanJ00}), while Hoare logic is more
often used for \emph{partial} correctness.
The parts that prove total correctness are those that involve the
\emph{variant} $\variant$, which decreases with each iteration by at
least $1$, and must be positive (or non-negative), so the loop must
stop at some point. This notion of variant is similar to those of \emph{ranking function}~\cite{floyd1993assigning} and \emph{Lyapunov function}~\cite{Khalil96}. 

\subsubsection{The $(\dwhilerule)$ Rule}
The second exception is the $(\dwhilerule)$ rule, which uses the
notion of \emph{Lie derivative}.
The term $\lieder{\vars}{\funs}{\expa}$ is called the \emph{Lie derivative
of $\expa$ with respect to the dynamics $\odeClause{\vars}{\funs}$}.
If $\vars$ is the list of 
variables $\var_1, \ldots, \var_n$,
and $\funs$ is the list of terms $\fun_1, \ldots, \fun_n$, its formal definition is
\[
	\lieder{\vars}{\funs}{\expa} \; =\; \sum\limits_{i=1}^n \partialder{\expa}{\var_i}
  \fun_i,
\]
where the terms $\partialder{\expa}{\var_i}$ are the
\emph{partial derivatives of $\expa$}
whose definitions are by induction on the structure of the term $\expa$ as
usual.
The fundamental lemma of the Lie derivative (see~\cite{trautman08}),
crucial for proving the soundness of the $(\dwhilerule)$ rule is the
following:
\begin{lemma}
  Assume given any solution $\hat x: \R_{\geq 0} \to \R^n$ of the
  differential equations $\odeClause{\vars}{\funs}$.
  Then the derivative of the function $t \mapsto \sem{\expa}{\hat x(t)}$
  is given by $t \mapsto \sem{\lieder{\vars}{\funs}{\expa}}{\hat x(t)}$.
  \label{lem:lieder}
\end{lemma}
\begin{proof}
  This is just an application of the chain rule.
\end{proof}

%

The $(\dwhilerule)$ rule is similar to the $(\whilerule)$ rule, in
that it contains an \emph{invariant} $\invariant$ ($\assertb$ in
$(\whilerule)$), a \emph{variant} $\variant$, and a \emph{terminator}
$\terminator$ (decreasing by $1$ in $(\whilerule)$).
The $\mathsf{inv}$ condition states that the invariant holds at the
start and is preserved by the dynamics, so it must hold at all times
along the dynamics.

The other conditions are only present to ensure that the loop
eventually terminates.
The $\mathsf{var}$ condition essentially means that the variant
$\variant$ must decrease along the dynamics (if the terminator
$\terminator$ is always negative).
But this is not enough, as the variant could get asymptotically
closer to $0$, without ever reaching it.
The condition $\mathsf{ter}$ ensures that this never happens, by
showing that the terminator $\terminator$ is not only negative, but
below a fixed negative value.





Note that, even though the $(\dwhilerule)$ rule may look like it is
about a single variable, $\variant$ is typically a term that contains
several variables, which makes it expressive enough to prove
interesting properties of driving systems.
For example, in~\eqref{eq:sub4pf1} in \cref{subsec:backProp}, $\variant$ depends on both $y$ and
$v$.

\begin{remark}\label{rem:extensionOfWandDW}
 We note that $(\whilerule)$ and $(\dwhilerule)$ can be made more general.
 For example, instead of the usual order on a single term $\variant$, $(\whilerule)$ could use a lexicographic order on several
 terms, or any well-founded order.
 This is also true for $(\dwhilerule)$, where we could use more general
 forms than $\variant > 0$ as the variant.

Indeed, in \cref{sec:workflow}, we will use a
 ``multiple-invariant multiple-variant'' generalization of  $(\dwhilerule)$; it is presented in \cref{fig:dwhile} in \cref{app:dRSS}. 
 In this section, we use the current simpler forms of the rules that are easy to describe and manipulate, and yet share their essence with the generalized forms.

\end{remark}

\subsubsection{The $(\dwhilesolrule)$ Rule}\label{subsubsec:dwhilesol}
Finally, let us discuss the $(\dwhilesolrule)$ rule in detail.
It uses explicit solutions, which makes it further from the spirit of
Hoare logic, but it is still valid.
Let us assume that $\odeClause{\vars}{\funs}$ has a closed form solution, that is, a
function $\hat \var: \mathbb{R}^n\times\mathbb{R} \to \mathbb{R}^n$ such that $\hat \var(\var_0,0) = \var_0$ and
$\frac{d \hat \var}{dt}(\var_0,t) = \sem{\funs}{\hat \var (\var_0,t)}$.
The only premise of $(\dwhilesolrule)$ is the $\dHL$ assertion shown in
\cref{fig:dFHL-rules}, where
$\asserta_t$ is a shorthand for the assertion
$\subst{\asserta}{\hat{\var}(\var_0,t)}{\var}$ (and similarly for
$\assertb_t$ and $\assertc_t$), while $\assertc_{{<}t}$ is a shorthand
for $\forall s \in [0,t).~ \subst{\assertc}{\hat \var(\var_{0},
s)}{\var}$ (and similarly for $\safetya_{\leq t}$).
Intuitively,
this rule means that for all states $x$ where the assertion $A$ holds, there is
some time $t$ when the condition $C$ just becomes false, and it is enough to
prove the assertion $B$ holds at time $t$, and that $S$ holds for all times from
$0$ to $t$.

\subsubsection{The $\caseKeyword$ Construct}
If we denote by $\case{\asserta_1}{\coma_1}{\asserta_n}{\coma_n}$ the
obvious nesting of $\ifKeyword$ constructs, then the following rule
can be derived from repeated uses of $(\ifrule)$ and $(\limprule)$:
\begin{equation}
  \def\defaultHypSeparation{\hskip.1in}
  \AxiomC{$\hquad{\safetya}{\asserta_1}{\hspace{-3pt}\coma_1\hspace{-3pt}}{\assertb}$}
  \AxiomC{\ldots}
  \AxiomC{$\hquad{\safetya}{\asserta_n}{\hspace{-3pt}\coma_n\hspace{-3pt}}{\assertb}$}
  \RightLabel{(\caserule)}
  \TrinaryInfC{$\hquad{\safetya}{\biglor_{i=1}^n \asserta_i}{\case{\asserta_1}{\coma_1}{\asserta_n}{\coma_n}}{\assertb}$}
  \DisplayProof
  \def\defaultHypSeparation{\hskip.2in}
  \label{eq:caserule}
\end{equation}
It is useful in our framework for automated driving: given
hybrid programs $\coma_1, \ldots, \coma_n$ that satisfy the same postcondition
$\assertb$ and safety condition $\safetya$, but with different
preconditions $\asserta_i$,
$\case{\asserta_1}{\coma_1}{\asserta_n}{\coma_n}$ also satisfies
$\assertb$ and $\safetya$, but on the more general precondition
$\asserta_1 \lor \ldots \lor \asserta_n$, as demonstrated in
\cref{subsec:collectAndComputeGlobal}.

\subsubsection{Soundness}
Soundness of \dHL{} can be proved:
\begin{theorem}\label{thm:dHLsoundness}
  Only valid Hoare quadruples can be proved in \dHL{}.
\end{theorem}

\begin{proof}
The proof is done by induction on the size of the proof tree and case
analysis of the first rule used. All cases are rather standard except for the
additional requirement of the safety condition,
so let us develop only the case when the last rule is
$(\dwhilerule)$ in details.

Let us assume the premises of $(\dwhilerule)$ are valid,
and assume given a store $\store$ such that $\store \vDash \asserta$.
The goal is to prove that
    $\state{\dwhileClause{\variant >
    0}{\odeClause{\vars}{\funs}}}{\store}$ converges to a store
    $\store'$ with $\store' \vDash \variant = 0 \land \invariant \sim
    0$, and
    for all reduction sequences $\state{\dwhileClause{\variant >
    0}{\odeClause{\vars}{\funs}}}{\store} \red{}^*
    \state{\coma}{\store''}$, $\store'' \vDash \invariant \sim 0 \land
    \variant \geq 0$.
  Let $\hat x: \R_{\geq 0} \to \R^n$ be the solution of
  $\odeClause{\vars}{\funs}$ with $\hat x(0) = \store$, $K = \setcomp{t
  \geq 0}{ \forall t' \leq t.\, \sem{\variant}{\hat{x}(t)} > 0}$, and
  $K' = \setcomp{t \geq 0}{ \forall t' \leq t.\,
  \sem{\variant}{\hat{x}(t)} > 0 \land \sem{\invariant}{\hat{x}(t)}
  \sim 0}$.
  Let $T$ be the supremum of $K'$.

\textbf{Step 1: $K = K'$ and $\sem{\invariant}{\hat{x}(T)} \sim 0$.}
  For the first point, since $K' \subseteq K$, it is sufficient to
  show that $\invariant \sim 0$ holds in $K$.
  Let us assume that $\sim$ is $=$ (other cases are similar).
  By $\mathsf{inv}$ and Lemma~\ref{lem:lieder}, the function $t \in K
  \mapsto \sem{\invariant}{\hat x(t)}$ is $0$ at $t = 0$ and of
  constant derivative $0$.
  This means its value at $t \in K$ is also $0$, that is,
  $\sem{\invariant}{\hat x(t)} = 0$.
  For the second point, because the function above is continuous and
  constantly equal to $0$ on $K$, we also have
  $\sem{\invariant}{\hat{x}(T)} \sim 0$ (other cases are similar).

\textbf{Step 2: $K$ is bounded.}
  By Step~1, $\sem{\invariant}{\hat x(t)} \sim 0$ for all $t \in K$.
  By $\mathsf{ter}$ and Lemma~\ref{lem:lieder}, the function $t
  \mapsto \sem{\terminator}{\hat x(t)}$ is negative at $t = 0$ and of
  non-positive derivative, i.e., non-increasing, on $K$.
  This means that $\sem{\terminator}{\hat x(t)} \leq
  \sem{\terminator}{\store}$.
  Similarly, by $\mathsf{var}$ and Lemma~\ref{lem:lieder}, the
  derivative of the function $v: t \mapsto \sem{\variant}{\hat x(t)}$
  is bounded by $\sem{\terminator}{\hat x(t)}$.
  By monotonicity of integrals:
  \[
    v(t) - v(0) = \int_{[0,t]} \dot{v}(t)dt
  	\leq \int_{[0,t]} \sem{\terminator}{\store}dt = t\cdot\sem{\terminator}{\store},
  \]
  so for $t > - \frac{\sem{\variant}{\store}}{\sem{\terminator}{\store}} \geq 0$,
  $\sem{\variant}{\hat x(t)} < 0$, so $t \notin K$, hence $K$ is
  bounded by $-
  \frac{\sem{\variant}{\store}}{\sem{\terminator}{\store}}$.

\textbf{Step 3: analysis of the exit time.}
By definition of $K'$, the loop ends at time $T$ (since $T$ is finite
by Step~2).
By Step~1, $T$ is also the supremum of $K$.
Furthermore, by openness of the condition $\variant > 0$ and the continuity of the
solution $\hat x$, if $T$ belonged to $K$, then there would exist $\epsilon > 0$, such that for all $t' \leq T + \epsilon$, $t' \in K$,
which would contradict the supremality of $T$. This means that
$\sem{\variant}{\hat x(T)} \leq 0$. Again by continuity of the solution,
$\sem{\variant}{\hat x(T)}$ is in the closure of $K$, which is included in $\R_{\geq 0}$.
Consequently, $\sem{\variant}{\hat x(T)} = 0$.

\textbf{Step 4: convergence.} The previous analysis implies that
$\state{\dwhileClause{\variant > 0}{\odeClause{\var}{\fun}}}{\store}$ converges to
$\store' = \hat x(T)$, for which $\store' \vDash \variant = 0$ (by Step~3) and
$\store' \vDash \invariant \sim 0$ (by Step~1).

\textbf{Step 5: safety.} By uniqueness of the solutions of $\odeClause{\vars}{\funs}$, we
can prove by induction on the number of reduction steps that
if $\state{\dwhileClause{\variant > 0}{\odeClause{\vars}{\funs}}}{\store} \red{}^*
		\state{\coma}{\store''}$, then $\store'' = \hat x(t)$ for some
$0 \leq t \leq T$.
By Step~3, $\store'' \vDash \variant \geq 0$, and by Step~1,
$\store'' \vDash \invariant \sim 0$.
\qedhere
\end{proof}

\subsubsection{Example}

We exemplify formal reasoning in $\dHL$ using the one-way traffic
scenario (\cref{ex:onewayTraffic,ex:onewaydHL}). 
We only show a typical part of the proof here, and refer the
interested reader to \cref{app:dRSS} for the complete formal proof.

\begin{example}[proving safety of the one-way traffic scenario]
  \label{ex:oneway-proof}
 We
  show how to prove the validity of the Hoare quadruple~\cref{eq:oneway-quad}---which we shall write as $\hquad{y_r < y_f}{A}{\alpha}{B}$---for the one-way traffic scenario in
  Example~\ref{ex:onewaydHL}.


 Here we use a slightly extended version of the
  $(\dwhilerule)$ rule, namely one that combines several variants and
  invariants. See \cref{rem:extensionOfWandDW}.
  The exact form of the rule is given in \cref{fig:dwhile} in
  \cref{app:dRSS}.

  The proof then relies on finding an invariant that implies $y_r <
  y_f$ and that is preserved by the dynamics $\alpha$.
  As always with proofs in program logics, finding a suitable invariant is
  difficult.
  In our case, a suitable invariant turns out to be
\begin{equation}\label{eq:oneWayInv}
     y_f - y_r - \dRSS(v_f, v_r, \rho - t) > 0\rlap{,}
\end{equation}
  where $t$ is the current time (note that we make the $\rho$
  parameter explicit throughout the proof, because it is important
  there).
  Explicit use of the assertion~\cref{eq:oneWayInv} as an invariant is not common in the literature---it is not used in~\cite{ShalevShwartzSS17RSS} for example---showing the subtlety of finding invariants.



Once a suitable invariant is found, constructing a $\dHL$ proof is  relatively simple: for each program
  construct, we apply the corresponding rule of $\dHL$.
    We present only part of the validity proof here, focusing on Line~2 of the program $\coma$ (\cref{fig:system-one-way-new}). The rest of the proof is similar.

We let $\coma'$ denote Line~2, that is,
%
  \[
   \coma' \;=\;
   \left(
    \dwhileClause{v_f > 0 \land v_r > 0 \land t <
      \rho}{\boxed{\;\dynamics{f}{}\;}\;, \;\boxed{\;\dynamics{r}{1}\;}}
  \right).
  \]
Here we used snippets $\dynamics{f}{},\dynamics{r}{1}$ from~\cref{eq:defDeltas}.

Then we want to prove that the following Hoare quadruple is valid:
\begin{equation}\label{eq:hoareQuadExampleLine2}
     \hquad{
      y_f - y_r - \dRSS(v_f, v_r, \rho - t) > 0
    }{
      \asserta'
    }{
      \coma'
    }{
      \assertb'
    } \rlap{,}
\end{equation}  
where
  \begin{align*}
    \asserta' &=
      \left(
      \begin{array}{l}
        v_r \geq 0 \land v_f \geq 0 \land t = 0 \land {} \\
        y_f - y_r > \dRSS(v_f, v_r, \rho)
      \end{array}
      \right) \rlap{,} \\
    \assertb' &=
      \left(
      \begin{array}{l}
        ((v_f \geq 0 \land t = \rho) \lor (v_f = 0 \land t \leq \rho)) \land {} \\
        v_r \geq 0 \land y_f - y_r > \dRSS(v_f, v_r, \rho - t)
      \end{array}
      \right) \rlap{.}
  \end{align*}
  This quadruple can be directly proved by applying the $(\dwhilerule)$ rule
  (and the $(\limprule)$ rule) with the following variants and invariants:
  \begin{itemize}
    \item $\invarianti{1}  \; = \; (v_r \geq 0)$,
    \item $\invarianti{2}  \; = \; (y_f - y_r - \dRSS(v_f, v_r, \rho -
      t) > 0)$,
    \item $\varianti{1}    \; = \; v_f$, \quad
          $\terminatori{1} \; = \; - \bmax$,
    \item $\varianti{2}    \; = \; \rho - t$, \quad
          $\terminatori{2} \; = \; -1$.
  \end{itemize}
  The only non-obvious point is that $\invarianti{2}$ is preserved by
  the dynamics. We first observe
  \[
    \liederdyn{\dynamics{f}{},\dynamics{r}{1}}{\invarianti{2}} =
    \left\{
      \begin{array}{ll}
        0 & \text{if $\dRSSpm(v_f,v_r,\rho-t) \geq 0$} \\
        v_f - v_r & \text{otherwise,}
      \end{array}
    \right.
  \]
  where $\dRSSpm(v_f,v_r,\rho)$ is given by
  \[
    \dRSSpm(v_f,v_r,\rho) \;=\;
    v_r \rho + \frac{\amax \rho^2}{2} + \frac{(v_r + \amax
      \rho)^2}{2\bmin} - \frac{v_f^2}{2 \bmax} \rlap{.}
  \]
  Therefore, we can infer as follows.
  \begin{align*}
    & \dRSSpm(v_f,v_r,\rho-t) < 0 \\
    & \Longleftrightarrow\; v_r (\rho-t) + \frac{\amax (\rho-t)^2}{2} + {}
      \\
    & \phantom{\,\Longleftrightarrow\,}\quad \frac{(v_r + \amax (\rho-t))^2}{2
      \bmin} - \frac{v_f^2}{2 \bmax} < 0 \\
    & \Longrightarrow\; \frac{(v_r + \amax (\rho-t))^2}{2 \bmin} -
      \frac{v_f^2}{2 \bmax} < 0
      & \text{$(i)$} \\
    & \Longrightarrow\; (v_r + \amax (\rho-t))^2 < v_f^2
      & \text{$(ii)$} \\
    & \Longrightarrow\; v_r^2 < v_f^2
      & \text{$(iii)$} \\
    & \Longrightarrow\; v_r < v_f
      & \text{$(iv)$}
  \end{align*}
  Here $(i)$ and $(iii)$ are because $v_r \geq 0$, $t \leq \rho$ (by
  $\invarianti{1}$ and $\varianti{2}$), and $\amax \geq 0$; $(ii)$
  because $0 \leq \bmin \leq \bmax$; and $(iv)$ because $v_r \geq 0$
  and $v_f \geq 0$ (the latter by $\varianti{1}$).

  The argument above concludes that $\invarianti{2}$ is indeed an invariant, which establishes the validity of the Hoare quadruple~\cref{eq:hoareQuadExampleLine2} on Line 2 of $\alpha$. Combining similar arguments, we prove the validity of  the Hoare quadruple  $\hquad{y_r < y_f}{A}{\alpha}{B}$ (from~\cref{eq:oneway-quad}) for the one-way traffic scenario.
  The rest of the proof can be found in \cref{app:dRSS}.
\end{example}

In the last example, in order to define $\dRSS$~\cref{eq:RSSMinDist} in $\dHL$, we needed to add 
  the $\max$ operator to the syntax for terms. This is straightforward.
\begin{auxproof}
  a term
  for $\max$, which we add to the syntax, and whose partial
  derivatives are defined as:
  \[
    \frac{\partial}{\partial \var}\max(\term,\term') =
    \left\{
      \begin{array}{ll}
        \frac{\partial \term}{\partial \var} & \text{if $\term \geq
          \term'$} \\
        \frac{\partial \term'}{\partial \var} & \text{otherwise,}
      \end{array}
    \right.
  \]
  to be understood as a term of the syntax.
  We can then define $\dRSS(v_f,v_r,\rho)$ as in~\eqref{eq:RSSMinDist}
  (note the explicit dependence on $\rho$).
\end{auxproof}

\section{Problem Formulation}\label{sec:problemFormulation}
\subsection{Modelling of Physical Components: Roads, Lanes, Occupancy, and Vehicle Dynamics}
\label{subsec:modelingPhysicalComponents}

We use the double integrator model as done in the original RSS work~\cite{ShalevShwartzSS17RSS}.
 Occupancy is lane-based. In changing lanes, a vehicle occupies two lanes---this modelling is reasonable in less-congested highway situations. This way we do not need to consider lateral positions of vehicles within a lane; this modelling is even simpler than the lane-based one in~\cite{PekA18IROS}. 

Concretely, we use integers to express lanes ($l=1,2,3, \dotsc$). A vehicle changing lanes from Lane~1 to~2 is expressed by $l=1.5$; it means that 1) the vehicle occupies both Lanes~1 \&~2, as discussed above, and 2) the vehicle is hence subject to the RSS distance responsibilities (\cref{ex:onewayTraffic}) with respect to preceding vehicles both in Lanes~1 \&~2.

 Our scope here is driving situations that are highly structured and thus allow  abstract  modelling in terms of lane occupancy. This is the case typically with highway traffic situations. Many other works, such as~\cite{SchmidtOB06ITSC}, study less structured driving situations; their scope is therefore different from ours.

\subsection{Scenario Modelling}\label{subsec:probFormlScenarioModeling}
What constitutes a mathematical notion of ``driving scenario'' is a difficult question---its answer can change depending on the intended model granularity and the goal of modelling. For our purpose of compositional derivation of goal-aware RSS rules in  $\dHL$, we propose the following definition.

\begin{definition}[(driving) scenario]\label{def:drivingScenario}
 A \emph{(driving) scenario} is a quadruple $\mathcal{S}=(\Var, \Safe, \Env, \Goal)$, where
\begin{itemize}
 \item $\Var$ is a finite set of variables;
 \item $\Safe$ is a $\dHL$ assertion called a \emph{safety condition};
 \item $\Env$ is a $\dHL$ assertion called a \emph{environmental condition}; and
 \item $\Goal$ is a $\dHL$ assertion called a \emph{goal}.
\end{itemize}
It is required that all the variables occurring in $\Safe$, $\Env$, and $\Goal$ belong to $\Var$. 
\end{definition}

The set $\Var$ should cover all the variables that are used for rule derivation; it is a modelling of the physical components involved in the driving scenario in question. We follow \cref{subsec:modelingPhysicalComponents} in deciding $\Var$. 

The three assertions $\Safe, \Env, \Goal$ describe different aspects of a driving scenario, and are thus used differently in our rule derivation workflow (\cref{sec:workflow}). 
\begin{itemize}
 \item $\Safe$ describes  safety conditions for collision avoidance.  \SV{} should satisfy them \emph{all the time} while it drives. 

Typically $\Safe$  requires the RSS safety distance (\cref{ex:onewayTraffic}) between \SV{} and some of \POV{}s. To be precise, the latter \POV{}s are those which are ahead of \SV{} in the same lane. (According to the RSS principles, the distance between \SV{} and a \POV{} \emph{behind} it is \SV{}'s concern only if \SV{} is cutting in---see $\Goal^{(1)}$ in \cref{eq:subgoalsPullOver}, \cref{subsec:subscenarioId}.)
 \item $\Env$ describes additional environmental conditions in driving---these conditions must be satisfied \emph{all the time} during driving, too, but ensuring them is not \SV{}'s responsibility but the environment's.

Environmental conditions typically include 
 1) assumptions on \POV{}s' dynamics (e.g.\ they maintain their speed), and 2) other assumptions imposed in the scenario, such as ``\POV{1} is behind \SV{}'' ($\mathcal{T}_{111}$ in \cref{ex:subscenarioRefinePullOver}). See \cref{ex:scenarioModelingPullover}. 
 \item $\Goal$ describes the goal condition of a driving scenario. It must be true \emph{at the end of} driving. 

Devising proper responses that achieve  $\Goal$---and proving that they do so safely---is a major feature of our framework that  the original (goal-unaware) RSS~\cite{ShalevShwartzSS17RSS} lacks.
\end{itemize}

\begin{example}\label{ex:scenarioModelingPullover}
For the pull over scenario (\cref{ex:pullover}), a scenario $\mathcal{S}=(\Var, \Safe, \Env, \Goal)$ is defined as follows. 

 The set $\Var$ of variables for the pull over scenario, following \cref{subsec:modelingPhysicalComponents}, are
\begin{itemize}
 \item $l,l_{1}, l_{2}, l_{3}$ for the lanes of \SV{} and the three \POV{}s;
 \item $y,y_{1}, y_{2}, y_{3}$ for their (longitudinal) positions; 
 \item $v,v_{1}, v_{2}, v_{3}$ for their (longitudinal) velocities; and
 \item $a,a_{1}, a_{2}, a_{3}$ for their (longitudinal) acceleration rates.
\end{itemize} 

The safety condition $\Safe$ is
\begin{equation}\label{eq:pullOverSafetyCond}
\begin{aligned}
 \Safe \;=\;
 &
 \textstyle\bigwedge_{i=1,2,3}\bigl(\,\aheadSL_{i} \Longrightarrow
 y_{i}-y > \dRSS(v_{i}, v)\,\bigr),
 \\
 &
 \land
   0\le v\le \vmax
  \land
  -\bmin\le a \le \amax\,.
\end{aligned}
\end{equation}
\begin{itemize}
 \item The first conjunct requires that \SV{} maintain the RSS safety distance $\dRSS$ (\cref{ex:onewayTraffic}) from the preceding vehicle. We used the following abbreviation (``ahead in the same lane'').
\begin{align}\label{eq:aheadAndSameLane}
 \aheadSL_{i} 
 \;&=\;
 y_{i}> y \land |l_{i}- l| \le 0.5.
\end{align}
Recall, e.g., that $l=1.5$ means \SV{}'s occupancy of both Lanes~1 and~2 (\cref{subsec:modelingPhysicalComponents}). 
 \item The second conjunct imposes the legal maximum velocity on \SV{}. In this paper, for simplicity, we do not impose the legal minimum speed on \SV{} ($\vmin\le v$) because of its emergency. In practice, this can be justified by turning on \SV{}'s hazard lights.  
  \todooptil{Clovis: we do maintain $\ldots \vmin \leq v \leq \vmax$
    (except in the last lane) in the derivation, though?
\\$\Rightarrow$ Ichiro: I remember proving that was hard at some point. We say as above to be on the safe side.
  }

 \item 
The third conjunct bounds \SV{}'s acceleration, where we require that it brakes comfortably (within $\bmin$) and does not engage emergency braking (not within $\bmax$), much like in \cref{ex:onewayTraffic}. 

\end{itemize}

The environmental condition $\Env$ is as follows.
\begin{align*}
  \label{eq:pullOverEnvAssmp}
 \Env\;=\; 
 &
 \textstyle\bigwedge_{i=1,2,3}
  \bigl(\,\vmin\le v_{i}\le \vmax
  \land a_{i} = 0
  \,\bigr)
 \\
 &\land l_{1}=2 \land l_{2}=2 \land l_{3}=1 \land y_{2} > y_{1}.
\end{align*}
We assume that \POV{}s do not change their speed ($a_{i}=0$)---an assumption we adopt  in this paper to simplify arguments. Violation of this assumption can affect goal achievement (i.e.\ reaching $\ytgt$ in Lane~3), but it does not endanger collision avoidance. See~\cref{subsubsec:discussionOnEnv}.

The goal $\Goal$ is to stop at the intended position, that is,
\begin{equation*}
\Goal \quad=\quad l=3\land y = \ytgt \land v = 0.
\end{equation*}
\end{example}

\begin{remark}[distinguishing $\Safe$ and $\Env$]
 It turns out that the mathematical positions of $\Safe$ and $\Env$ are the same in our workflow in \cref{sec:workflow}. Therefore there is no theoretical need of separating them. 

We nevertheless distinguish them for their conceptual difference:  $\Safe$ is an invariant that \SV{} must maintain, while   $\Env$ is an invariant that  \SV{} can assume. 
  Separating $\Safe$ and $\Env$ also helps modelling the scenario,
  because treating them separately restricts the modeller's focus to
  specific agents.
\begin{auxproof}
(Ichiro: I omitted this since it suggests our current quadruple-based formalization is faulty. We'd need quintuples, with environmental assumptions treated separately)
 Violation of $\Env$ means that the subsequent events are no longer within the scope of the driving scenario $\mathcal{S}$ in question, and therefore \SV{} (or, to be precise, the controller of \SV{} that is in charge of $\mathcal{S}$) is not held responsible.
\end{auxproof}
\end{remark}

\subsection{Our Problem: Goal-Aware RSS Rules as $\dHL$ Quadruples}\label{subsec:problemRulesAsTriples}
Using~\cref{def:drivingScenario}, we can formalise what we are after:
\begin{definition}[goal-aware RSS rule]\label{def:GARSSRule}
  Let $\mathcal{S}=(\Var, \Safe, \Env, \Goal)$ be a driving scenario. A \emph{goal-aware RSS rule} (or \emph{GA-RSS rule}) is a pair $(A,\alpha)$ of 
\begin{itemize}
 \item 
 a $\dHL$ assertion $A$ (called an \emph{RSS condition}), and
 \item 
 a $\dHL$ program $\alpha$ (called a \emph{proper response}), 
\end{itemize}
such that the quadruple
\begin{equation}\label{eq:ruleDerivationAsHoareQuadruple}
 \hquad{\Safe\land\Env}{A}{\alpha}{\Goal}
\end{equation}
is valid.
\end{definition}
Note that an RSS condition $A$ is in the position of a precondition in the quadruple~\cref{eq:ruleDerivationAsHoareQuadruple}. 

\begin{remark}\label{rem:POVDyn}
As a convention, in \cref{def:GARSSRule}, the $\dHL$ program $\alpha$ controls only \SV{}. The actual dynamics of the whole driving situation includes parts that model  \POV{}s' dynamics
too---they are described by $\odeClause{y_{i}}{v_{i}}, \odeClause{v_{i}}{a_{i}}$, where $a_{i}$ is typically constrained in $\Env$. 

We use this convention throughout the paper, describing only the control of \SV{} and leaving \POV{}s' dynamics implicit in $\dHL$ programs. We do so e.g.\ in \cref{ex:properRespIdPullOver}.


\end{remark}






\section{Compositional Derivation of Goal-Aware RSS Rules: a General Workflow}\label{sec:workflow}

In this section, we present a general workflow that compositionally derives a goal-aware RSS rule $(A,\alpha)$. In the workflow, the original scenario $\mathcal{S}$ is split up into a tree $\mathcal{T}$ of subscenarios---such as one shown in \cref{fig:ISSs}. Each subscenario is simplified and has a more specific scope, which allows one to come up with proper responses and their preconditions more easily. These \emph{subscenario} proper responses and preconditions  get bundled up, using $\dHL$ rules such as $(\seqrule)$ and $(\caserule)$, to finally yield a goal-aware RSS rule for the original scenario.

The outline of our rule derivation workflow is Procedure~\ref{alg:workflow}. Some steps of the workflow are illustrated in Figures~\ref{fig:ISSs}--\ref{fig:propagation}.

 \begin{figure*}[tbp]
 \begin{minipage}[t]{\textwidth}
 \centering
 \includegraphics[width=\textwidth]{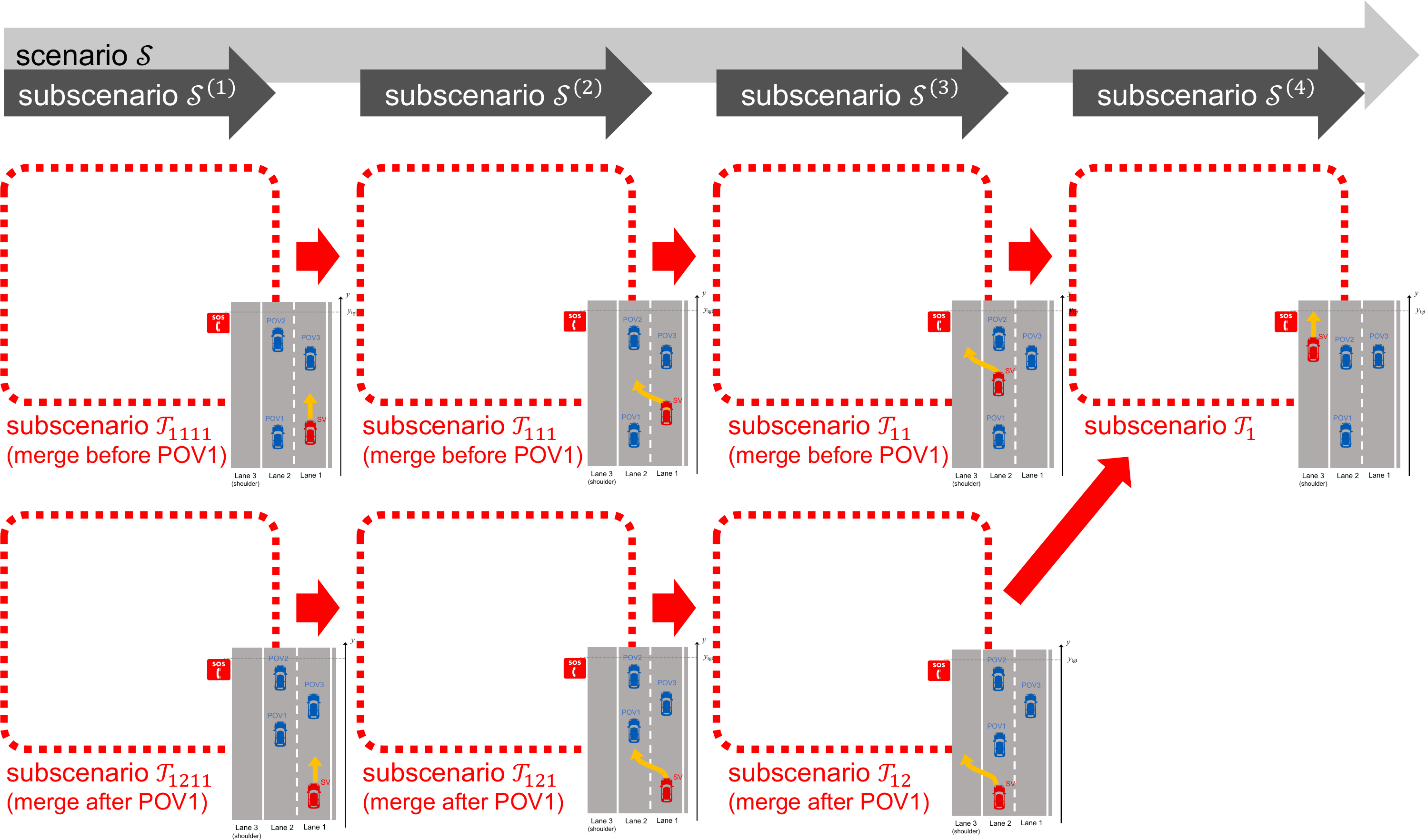} 
 \caption{The subscenario tree $\mathcal{T}$ in \cref{ex:subscenarioRefinePullOver}, obtained on \cref{line:subscenarioRefine} of Procedure~\ref{alg:workflow}, for the pull over scenario $\mathcal{S}$ (\cref{ex:scenarioModelingPullover}). The subscenarios $\mathcal{T}_{1}, \mathcal{T}_{11}, \dotsc$ are defined in \cref{fig:subscenarioDefPullOver}}
 \label{fig:ISSs}
 \end{minipage}
 \hfill
 \begin{minipage}[t]{\textwidth}
 \centering
 \includegraphics[width=\textwidth]{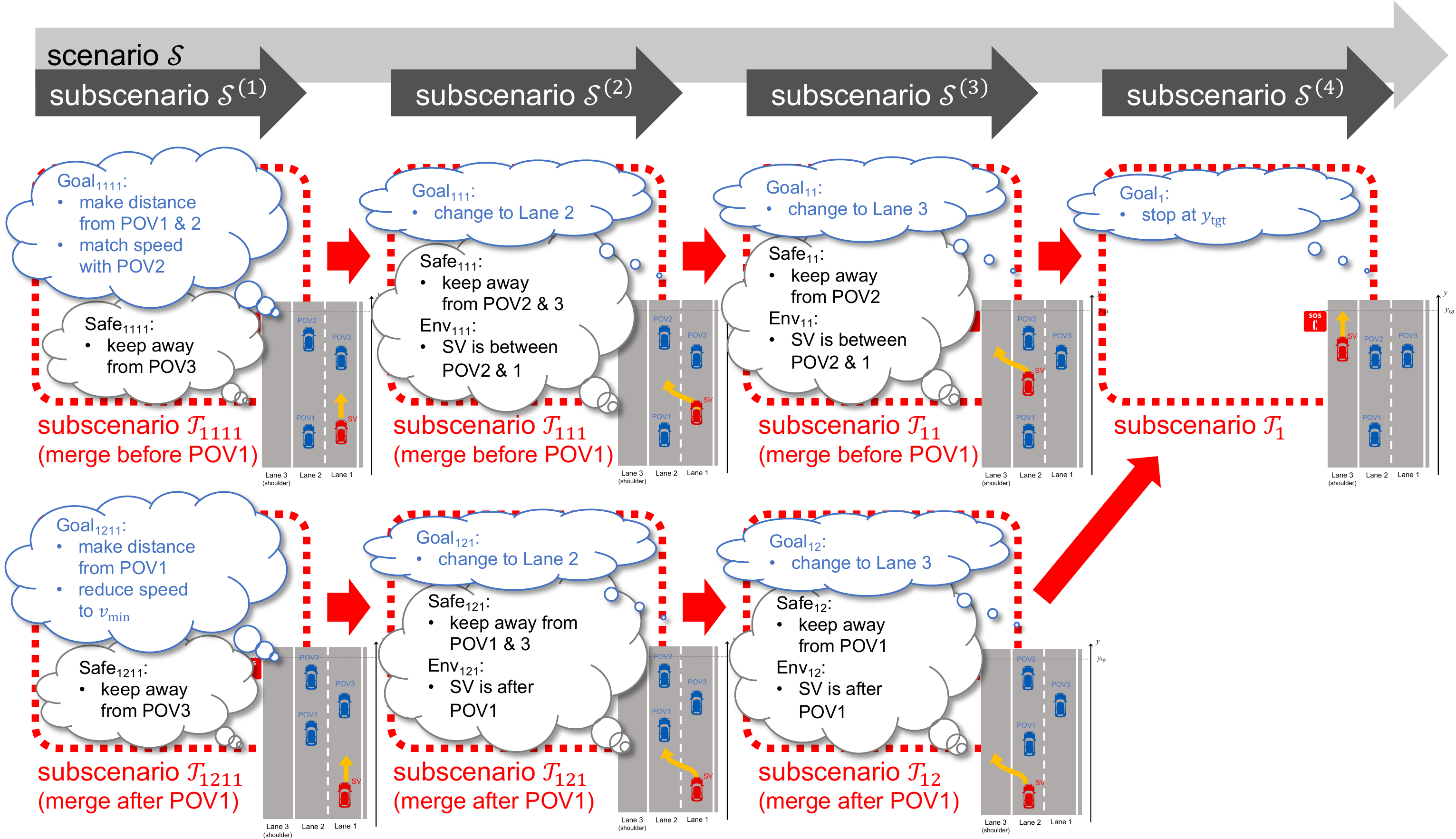} 
 \caption{The subscenario tree $\mathcal{T}$ in \cref{ex:subscenarioRefinePullOver}, highlighting (informally) the goal $\Goal_{w}$, the safety condition $\Safe_{w}$, and the environmental condition $\Env_{w}$ of each subscenario $\mathcal{T}_{w}$. The full formal definition is in \cref{fig:subscenarioDefPullOver}}
 \label{fig:subgoals}
 \end{minipage}
\end{figure*}
\begin{figure*}[tbp]
 \begin{minipage}[t]{\textwidth}
 \centering
 \includegraphics[width=\textwidth]{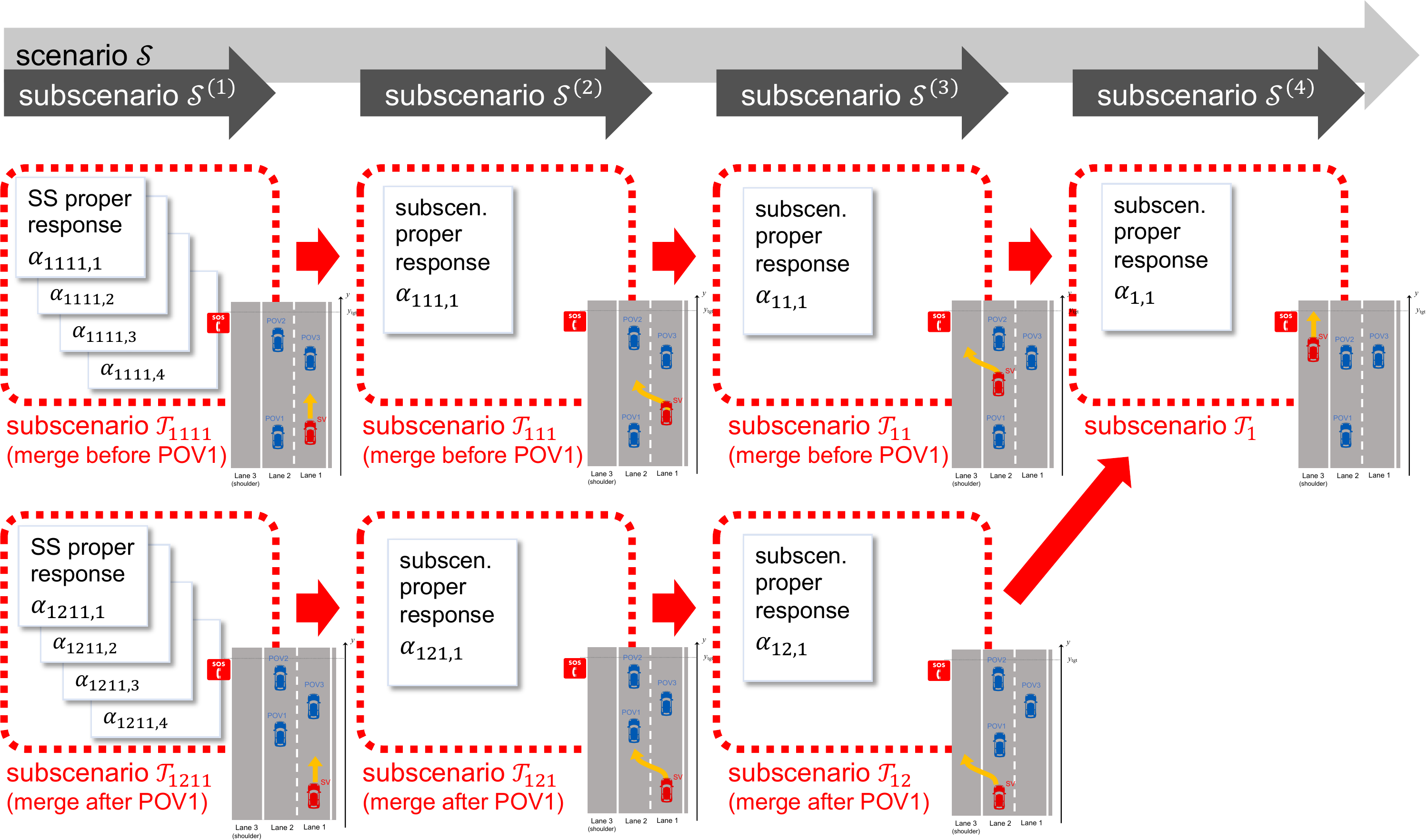} 
 \caption{The subscenario proper responses $\alpha_{1,1}, \alpha_{11,1}, \dotsc$ in \cref{ex:properRespIdPullOver}, obtained on \cref{line:subscenarioPropRespId} of Procedure~\ref{alg:workflow},  for the pull over scenario $\mathcal{S}$ (\cref{ex:scenarioModelingPullover}).}
 \label{fig:maneuverSeqs}
 \end{minipage}
 \hfill
 \begin{minipage}[t]{\textwidth}
 \centering
 \includegraphics[width=\textwidth]{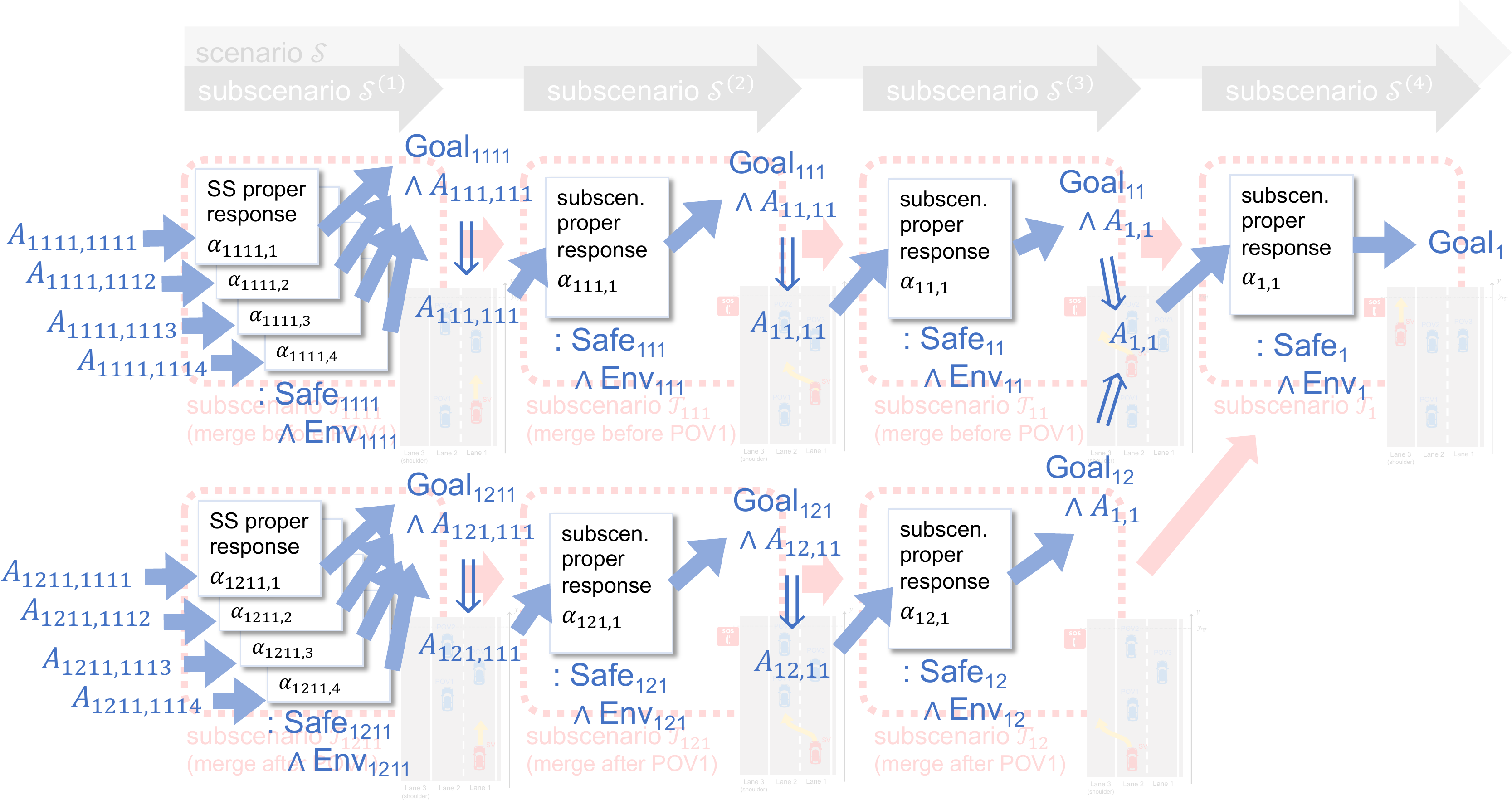} 
 \caption{The subscenario preconditions $A_{1,1}, A_{11,11}, \dotsc$ in \cref{ex:backPropPullOver},  obtained on \cref{line:backProp} of Procedure~\ref{alg:workflow}, for the pull over scenario $\mathcal{S}$ (\cref{ex:scenarioModelingPullover}). Here the solid arrows represent Hoare quadruples (with a program in the middle and a safety condition below); note that they realise the condition~\cref{eq:backPropAssignmentCond}. The double arrows $\Longrightarrow$ represent logical implication.}
 \label{fig:propagation}
 \end{minipage}

 \end{figure*}

\begin{algorithm*}[tbp]
\SetAlgorithmName{Procedure}{procedure}{List of Procedures}
\LinesNumbered
\caption{our workflow for compositional derivation of goal-aware RSS rules }
\label{alg:workflow}
\KwIn{a driving scenario}
\KwOut{a goal-aware RSS rule $(A,\alpha)$ (\cref{def:GARSSRule})}

\emph{Scenario modelling}:
mathematically model the driving scenario, obtaining $\mathcal{S}=(\Var, \Safe, \Env, \Goal)$ (cf.~\cref{subsec:probFormlScenarioModeling})\; \label{line:scenarioId}

\emph{Goal decomposition}:
identify subgoals  $\Goal^{(1)},\dotsc, \Goal^{(N)}$  and decompose  the scenario
 $\mathcal{S}$ into subscenarios $\mathcal{S}^{(1)},\dotsc, \mathcal{S}^{(N)}$ where $\mathcal{S}^{(i)}=(\Var, \Safe, \Env, \Goal^{(i)})$\; \label{line:goalDecomposition}

\emph{Subscenario refinement}:
refine subscenarios, distinguishing cases and strengthening safety and environmental conditions. Then express their causal relationships and obtain a subscenario tree $\mathcal{T}$\; \label{line:subscenarioRefine}

%
\ForEach{subscenario $\mathcal{T}_{w}=(\Var,\Safe_{w}, \Env_{w}, \Goal_{w})$ in $\mathcal{T}$}{
  Identify  \emph{subscenario proper responses}: find programs  $\alpha_{w,1}, \dotsc, \alpha_{w,K_{w}}$  that achieve $\Goal_{w}$ and maintain $\Safe_{w}\land\Env_{w}$ during their execution. (The programs $\alpha_{w,i}$ may include syntactic parameters $F, G, \dotsc$; they are instantiated by concrete expressions on \cref{line:backProp})\; \label{line:subscenarioPropRespId}
}

 Identify \emph{subscenario preconditions}: find $\dHL$ assertions $(A_{w,u})_{w,u}$ for each subscenario proper response $\alpha_{w,i}$, reasoning  backwards from shorter $w$ to longer, so that they guarantee subgoal achievement as well as the next preconditions  (formalised in~\cref{eq:backPropAssignmentCond})\;
\label{line:backProp}

Compute \emph{global proper response and precondition}:
 combine the  subscenario proper responses $\alpha_{w,i}$ and the subscenario preconditions $A_{w,u}$ obtained so far, to obtain  $A$ and $\alpha$ such that $\hquad{\Safe\land\Env}{A}{\alpha}{\Goal}$ is valid\; \label{line:collectAndComputeGlobal}

\Return{$(A,\alpha)$}
\end{algorithm*}

Each step of the workflow is described in detail below. We use the pull over scenario  (\cref{ex:pullover}) as a leading example in its course. Another example scenario, which is more complex, is discussed later in \cref{subsec:limitedVisibility}. 

\subsection{Scenario Modelling (\cref{line:scenarioId})}\label{subsec:scenarioId}
We assume that the input driving scenario is given only in informal terms. In this step, we identify its mathematical modelling  $(\Var, \Safe, \Env, \Goal)$ in the sense of \cref{subsec:probFormlScenarioModeling}. See \cref{ex:scenarioModelingPullover} for a concrete example for the pull over scenario.

\subsection{Subscenario Identification (Lines~\ref{line:goalDecomposition}--\ref{line:subscenarioRefine})}\label{subsec:subscenarioId}
In the two steps on Lines~\ref{line:goalDecomposition}--\ref{line:subscenarioRefine}, we decompose the original problem (namely, to find $A$ and $\alpha$ such that $\hquad{\Safe\land\Env}{A}{\alpha}{\Goal}$) into problems about smaller subscenarios. We aim to identify subscenarios such that  1) they make local objectives and case distinctions explicit, 2)  each subscenario is simpler and more homogeneous, and 3) the safety and environmental conditions for each subscenario are more concrete and specific. These features will make it easier to devise proper responses and preconditions for those subscenarios. 


The following formal definition will be justified in the course of the explanation below, notably in the proof of \cref{thm:correctness}. 
 \begin{definition}[subscenario, subgoal]\label{def:subscenario}
  Let $\mathcal{S}=(\Var, \Safe, \Env, \Goal)$ and $\mathcal{S}'=(\Var, \Safe', \Env', \Goal')$ be scenarios with the same variable set. We say that $\mathcal{S}'$ is a \emph{subscenario} of $\mathcal{S}$ if both of the  logical implications  $\Safe'\land\Env'\Rightarrow\Safe$ and $\Safe'\land\Env'\Rightarrow\Env$ are valid. In this case, $\Goal'$ is called a \emph{subgoal}. 
 \end{definition}


We separate the task of subscenario identification into \emph{goal decomposition} (\cref{line:goalDecomposition}) and \emph{subscenario refinement} (\cref{line:subscenarioRefine}). The separation is not a necessity from the theoretical point of view.
We nevertheless explicate the separation for conceptual and practical reasons: in our experience, the two-step workflow (Lines~\ref{line:goalDecomposition}--\ref{line:subscenarioRefine}) is the way we came up with useful subscenarios. 

\subsubsection{Goal Decomposition (\cref{line:goalDecomposition})}\label{subsubsec:goalDecomp}
On \cref{line:goalDecomposition}, we aim at a series $\Goal^{(1)},\dotsc,\Goal^{(N)}$ of subgoals that naturally paves the way to the original goal $\Goal$. More specifically, we expect the subgoals 
to be such that their achievement in the given order leads to the achievement of $\Goal$. 
On \cref{line:goalDecomposition}, note that the resulting subscenarios $\mathcal{S}^{(i)}=(\Var, \Safe, \Env, \Goal^{(i)})$  all have the same safety and environmental conditions $\Safe, \Env$ as the original scenario $\mathcal{S}$. Strengthening those conditions is left to the next step (\cref{line:subscenarioRefine}).

Note that, in fact, any sequence  $\Goal^{(1)},\dotsc,\Goal^{(N)}$  of $\dHL$ assertions qualifies as the outcome of \cref{line:goalDecomposition}---\cref{def:subscenario}  does not constrain $\Goal'$. 
However, a good choice of subgoals   $\Goal^{(1)},\dotsc,\Goal^{(N)}$   eases the rest of the workflow by making local objectives explicit. It is usually easy to come up with a natural series of subgoals, too, as we demonstrate now. 

\begin{example}\label{ex:goalDecompPullover}
 For the pull over scenario (\cref{ex:pullover,ex:scenarioModelingPullover}), we use the goal decomposition that we informally described in \cref{sec:intro} (Subscenario 1--4). These subscenarios arise from 1) coming to a halt (Subscenario~4), 2) changing lanes (Subscenarios~2--3), and 3) preparing for lane changes, in case there are vehicles in the destination lane (Subscenario~1). 

The corresponding subgoals are formalised as follows. 
\begin{align}\label{eq:subgoalsPullOver}\small
  \notag
  \Goal^{(1)}\;&=\;
  \left(\!\!
  \footnotesize\begin{array}{l}
   y_{2}-y\ge \dRSS(v_{2},v)
   \\ {}\land y-y_{1}\ge\dRSS(v,v_{1})
   \\{}\land
   v_{2} = v
	\end{array}
  \!\!
  \right)
  \lor
  \left(\!\!
  \footnotesize\begin{array}{l}
  y_{1}-y\ge \dRSS(v_{1},v)
  \\{}\land
   \vmin = v
	\end{array}
  \!\!\right)
  \\
  \Goal^{(2)}\;&=\;
  (l = 2)
  \\
  \notag
  \Goal^{(3)}\;&=\;
  (l = 3)
  \\
  \notag
  \Goal^{(4)}\;&=\;
  (l = 3 \land y = \ytgt \land v = 0)
\end{align}
  \todooptil{Clovis: reverse order? Also $v = v_2$ and $v = \vmin$? \\ 

Ichiro: let me take the current order that matches the preceding text. On $v = v_2$, I'm too lazy to fix all the references to the equation... Sorry :(}
We define $\mathcal{S}^{(i)}=(\Var, \Safe, \Env, \Goal^{(i)})$ (for $i=1,\dotsc, 4$), where $\Var, \Safe, \Env$ are the ones in \cref{ex:scenarioModelingPullover}.
These subscenarios $\mathcal{S}^{(i)}$ appear at the top of \cref{fig:ISSs}.

The two disjuncts in $\Goal^{(1)}$ represent 1) the case of \SV{} merging between \POV{2} and  \POV{1}, and 2) that of merging behind \POV{1}, respectively. (For simplicity, we ignore the case of merging in front of \POV{2}.) In the former case,  keeping enough distance from \POV{1} is deemed to be the responsibility of \SV{}---although \POV{1} is behind \SV{}, it is \SV{}'s lane-changing manoeuvre that creates the duty of distance keeping. One can also see this responsibility as an instance of the RSS responsibility principle 2) ``Don't cut in recklessly''---see \cref{subsec:introCARSS}.

In the first disjunct of  $\Goal^{(1)}$, we additionally require that \SV{}'s velocity matches that of the preceding vehicle. We do so because 1) it is a natural driving practice, and 2) it eases the safety analysis of the later subscenarios (see the case for $\mathcal{T}_{11}$ in \cref{ex:backPropPullOver},  for example). For the second disjunct, for similar reasons, we require that \SV{}'s velocity is the legal minimum.
  \todooptil{Clovis: not so much about ``safety analysis of later
  subscenarios'' as about solvability of the current subscenario? \\

Ichiro: I don't know the precise meaning of ``solvability of the current subscenario'' so let's go with the current writing, unless it causes a big trouble.}
\end{example}

\subsubsection{Subscenario Refinement (\cref{line:subscenarioRefine})}\label{subsubsec:subscenarioRef}
The case distinction in $\Goal^{(1)}$ of \cref{ex:goalDecompPullover} (to merge before or after \POV{1}) is typical in our workflow: there are different possible inter-vehicle relationships; distinguishing cases with respect to them makes each case simpler and more focused. 

On \cref{line:subscenarioRefine}, we make such case distinction explicit as different subscenarios. Relating the resulting subscenarios by their causal relationship, we obtain a \emph{tree} of subscenarios. See~\cref{fig:ISSs} for an example.

\begin{notation}
 We  use words $w\in (\mathbb{Z}_{>0})^{*}$ to designate nodes of a tree $\mathcal{T}$, as is common in the literature. Specifically,  1) the root of $\mathcal{T}$ is denoted by 
$\varepsilon$
(where $\varepsilon$ stands for the empty word), and  2) the $k$-th child of a node
$w$
is denoted by
$wk$.
\end{notation}

\begin{definition}[subscenario tree]\label{def:subscenarioTree}
  Let $\mathcal{S}=(\Var, \Safe, \Env, \Goal)$ be a scenario. A \emph{subscenario tree} $\mathcal{T}$ for $\mathcal{S}$ is a finite tree 
\begin{itemize}
 \item whose root is not labelled (we write $\bullet$ for the root label), 
 \item whose non-root node $w$ is labelled by a subscenario $\mathcal{T}_{w}$ of $\mathcal{S}$ (cf.\ \cref{def:subscenario}), and
 \item additionally, for every node of depth 1 (i.e.\ $\mathcal{T}_{w}$ with $|w|=1$), the corresponding subscenario $\mathcal{T}_{w}=(\Var,\Safe_{w}, \Env_{w}, \Goal_{w})$  satisfies $\Safe_{w}\land\Env_{w}\land\Goal_{w}\Rightarrow \Goal$, where $\Goal$ is the goal of $\mathcal{S}$.
\end{itemize}
Hence $\mathcal{T}_{\varepsilon}=\bullet$ for the root, and $\mathcal{T}_{w}$ is a subscenario for  $w\neq \varepsilon$.
\end{definition}
In the third item above, a subscenario $\mathcal{T}_{w}$ with $|w|=1$ is one of those which are executed at the end (see $\mathcal{T}_{1}$ in \cref{fig:ISSs} for an example). The item is a natural requirement that its goal $\Goal_{w}$ implies the goal $\Goal$ of the whole scenario $\mathcal{S}$, potentially with the help of $\Safe_{w}$ and $\Goal_{w}$.



A subscenario tree $\mathcal{T}$ arises naturally from the outcome of \cref{line:goalDecomposition} (namely $\mathcal{S}^{(1)},\dotsc, \mathcal{S}^{(N)}$) by distinguishing cases, as demonstrated below. Note that case distinction also helps concretising safety conditions. 

\begin{example}\label{ex:subscenarioRefinePullOver}
Continuing \cref{ex:goalDecompPullover}, we obtain the subscenario tree $\mathcal{T}$ shown in \cref{fig:ISSs} as a possible outcome of \cref{line:subscenarioRefine}.  We do so  by distinguishing cases of \SV{} merging  before or after \POV{1}. 
%
The subscenarios $\mathcal{T}_{w}$ in $\mathcal{T}$ are defined in \cref{fig:subscenarioDefPullOver}, where $\mathcal{T}_{w}=(\Var,\Safe_{w}, \Env_{w}, \Goal_{w})$. We use the following abbreviation; it is much like $\aheadSL_{i}$ in~\cref{eq:aheadAndSameLane}. 
\begin{align*}
 \behindSL_{i} 
 \;&=\;
 y_{i}< y \land |l_{i}- l| \le 0.5.
\end{align*}

The design of the subscenarios $\mathcal{T}_{w}$ is described below. Some key conditions therein are highlighted in \cref{fig:subgoals}. 

\underline{\bfseries The subscenario $\mathcal{T}_{1}$}
This comes from $\mathcal{S}^{(4)}$ in \cref{ex:goalDecompPullover}. The condition $l=3$ in the original goal $\Goal^{(4)}$ is moved to the safety condition $\Safe_{1}$ since it has to be maintained throughout rather than  achieved at the end. Requiring $l=3$ in $\mathcal{S}^{(4)}$ automatically discharges the RSS safety distance requirement in the overall safety condition $\Safe$ (see~\cref{eq:pullOverSafetyCond}) since $\aheadSL_{i}$ is false. As a result, the subscenario safety condition $\Safe_{1}$ is much simplified.

\underline{\bfseries The subscenario $\mathcal{T}_{11}$}
This comes from $\mathcal{S}^{(3)}$ in \cref{ex:goalDecompPullover}, and assumes that \SV{} has merged between \POV{1} and \POV{2}. The last assumption is found in the environmental condition $\Env_{11}$. Consequently, the RSS  distance requirement is simplified: in $\Safe_{11}$, only the one for \POV{2} is required. 

Note that we also assume $0\le v\le v_{2}$ as part of the safety condition. This assumption may not be necessary but simplifies the subsequent reasoning a lot, especially when it comes to proving maintenance of the RSS safety distance. This assumption can be enforced, too, by requiring velocity matching in our subgoals ($v_{2}=v$ and $\vmin = v$ in $\Goal^{(1)}$ in~\cref{eq:subgoalsPullOver}, and thus in $\Goal_{1111}, \Goal_{1211}$ in \cref{fig:subscenarioDefPullOver}).

\underline{\bfseries The subscenarios $\mathcal{T}_{12}, \mathcal{T}_{111}, \mathcal{T}_{121}$} Similarly to $\mathcal{T}_{11}$, we 1) explicate case distinction in the  environmental conditions $\Env_{w}$, and 2) simplify the safety conditions $\Safe_{w}$, adding some extra assumptions (such as $v\le v_{2}$) if we find them useful.

\underline{\bfseries The subscenarios $\mathcal{T}_{1111}, \mathcal{T}_{1211}$} 
These come from the two disjuncts of $\Goal^{(1)}$ (see~\cref{eq:subgoalsPullOver}): their goals are precisely those disjuncts; and the safety conditions $\Safe_{1111}, \Safe_{1211}$ are the original safety condition $\Safe$ simplified using $l=1$.

\underline{\bfseries Each $\mathcal{T}_{w}$ is indeed a subscenario.}
It is not hard to show that each  $\mathcal{T}_{w}$ is indeed a subscenario of $\mathcal{S}$ from \cref{ex:scenarioModelingPullover}, in the sense of \cref{def:subscenario}, as required in \cref{def:subscenarioTree}. 
\begin{itemize}
 \item 
  For $\mathcal{T}_{1}$, we  have to show that $\Safe_{1}\land\Env_{1}\Rightarrow \Safe$ holds. Since $l=3$ is in $\Safe_{1}$ and $l_{1}=2, l_{2}=2, l_{3}=1$ are in $\Env_{1}$, we see that $\aheadSL_{i}$ is false for each $i=1,2,3$; this makes $\Safe$ in~\cref{eq:pullOverSafetyCond} trivially true.
 \item For $\mathcal{T}_{11}$,  $\Safe_{11}\land\Env_{11}\Rightarrow \Safe$ can be shown as follows. Note first that $0\le v\le \vmax$ is inferred from $0\le v\le v_{2}$ (in $\Safe_{11}$) and $v_{2}\le \vmax$ (in $\Env_{11}$).

If $l=3$ then $\Safe$ is trivially true, much like in the above. Otherwise $l=2.5$ holds, which forces $\behindSL_{1}$ to hold (by $\Env_{11}$). Therefore $\aheadSL_{1}$ is false (it contradicts with $\behindSL_{1}$), and $\Safe$ is equivalent to $y_{2}-y>\dRSS(v_{2},v)$. The last is required in $\Safe_{11}$. 
 
\item Proofs for $\mathcal{T}_{12}, \mathcal{T}_{111}, \mathcal{T}_{121}$ are similar to the one for $\mathcal{T}_{11}$. 
\end{itemize}

\cref{def:subscenarioTree} additionally requires $\Safe_{1}\land\Env_{1}\land\Goal_{1}\Rightarrow \Goal$, whose validity is obvious.
\end{example}

\begin{figure}[tbp]
\scalebox{.8}{\begin{minipage}{.48\textwidth}
  \begin{align*}
 \mathcal{T}_{1}:
 &\begin{array}[t]{rl}
  \Safe_{1} \;=\;
  &
  l = 3  
  \land
   0\le v\le \vmax
  \land
  -\bmin\le a \le \amax
  \\ 
  \Env_{1} \;=\;
  &
  \Env
  \\
  \Goal_{1} \;=\;
  &
  y = \ytgt \land v = 0
 \end{array}
 \\[-.6em]
\begin{minipage}{3em}
  \dotfill
 \end{minipage}
 &\begin{minipage}{\textwidth}
  \dotfill
 \end{minipage}
 \\[-.6em]
 \mathcal{T}_{11}:
 &\begin{array}[t]{rl}
  \Safe_{11} \;=\;
  &
  (l=2.5\lor l= 3) 
 \land 0\le v \le v_{2}\land
  y_{2}-y\ge \dRSS(v_{2},v) 
  \\
  &
  \land
  -\bmin\le a \le \amax
  \\ 
  \Env_{11} \;=\;
  &
  \Env 
  \land (l=2.5 \Rightarrow \behindSL_{1} \land \aheadSL_{2})  
  \\
  \Goal_{11} \;=\;
  &
  l = 3 
 \end{array}
  \\[-.6em]
\begin{minipage}{3em}
  \dotfill
 \end{minipage}
 &\begin{minipage}{\textwidth}
  \dotfill
 \end{minipage}
 \\[-.6em]
 \mathcal{T}_{12}:
 &\begin{array}[t]{rl}
  \Safe_{12} \;=\;
  &
  (l=2.5\lor l= 3)
 \land 0\le v \le v_{1}
 \land
  y_{1}-y\ge \dRSS(v_{1},v)
  \\
  &{}
  \land
  -\bmin\le a \le \amax
  \\ 
  \Env_{12} \;=\;
  &
  \Env
  \land (l=2.5 \Rightarrow \aheadSL_{1})
  \\
  \Goal_{12} \;=\;
  &
  l = 3 
 \end{array}
  \\[-.6em]
\begin{minipage}{3em}
  \dotfill
 \end{minipage}
 &\begin{minipage}{\textwidth}
  \dotfill
 \end{minipage}
 \\[-.6em]
 \mathcal{T}_{111}:
 &\begin{array}[t]{rl}
  \Safe_{111} \;=\;
  &
  (l=1.5\lor l= 2) \land   0\le v \le v_{2}
 \\&{}\land
 y_{2}-y\ge \dRSS(v_{2},v)\land y_{3}-y\ge \dRSS(v_{3},v)
 \\
  &{}
  \land
  -\bmin\le a \le \amax
  \\ 
  \Env_{111} \;=\;
  &
  \Env \land \behindSL_{1} \land \aheadSL_{2}
  \\
  \Goal_{111} \;=\;
  &
  l = 2
 \end{array}
  \\[-.6em]
\begin{minipage}{3em}
  \dotfill
 \end{minipage}
 &\begin{minipage}{\textwidth}
  \dotfill
 \end{minipage}
 \\[-.6em]
 \mathcal{T}_{121}:
 &\begin{array}[t]{rl}
  \Safe_{121} \;=\;
  &
  (l=1.5\lor l= 2) \land  0\le v \le v_{1} 
  \\
  & {}\land y_{1}-y\ge \dRSS(v_{1},v) \land y_{3}-y\ge \dRSS(v_{3},v)
 \\
  &{}
  \land
  -\bmin\le a \le \amax
  \\ 
  \Env_{121} \;=\;
  &
  \Env \land\aheadSL_{1}
  \\
  \Goal_{121} \;=\;
  &
  l = 2
 \end{array}
  \\[-.6em]
\begin{minipage}{3em}
  \dotfill
 \end{minipage}
 &\begin{minipage}{\textwidth}
  \dotfill
 \end{minipage}
 \\[-.6em]
 \mathcal{T}_{1111}:
 &\begin{array}[t]{rl}
  \Safe_{1111} \;=\;
  &
  l=1 \land
  y_{3}-y\ge \dRSS(v_{3},v)
 \\
  &{}
  \land
   0\le v\le \vmax
  \land
  -\bmin\le a \le \amax
  \\ 
  \Env_{1111} \;=\;
  &
  \Env
  \\
  \Goal_{1111} \;=\;
  &
  y_{2}-y\ge \dRSS(v_{2},v)\land y-y_{1}\ge\dRSS(v,v_{1})
  \\
  & {}\land v_{2} = v
 \end{array}
  \\[-.6em]
\begin{minipage}{3em}
  \dotfill
 \end{minipage}
 &\begin{minipage}{\textwidth}
  \dotfill
 \end{minipage}
 \\[-.6em]
 \mathcal{T}_{1211}:
 &\begin{array}[t]{rl}
  \Safe_{1211} \;=\;
  &
  l=1 \land
  y_{3}-y\ge \dRSS(v_{3},v)
 \\
  &{}
  \land
   0\le v\le \vmax
  \land
  -\bmin\le a \le \amax
  \\ 
  \Env_{1211} \;=\;
  &
  \Env
  \\
  \Goal_{1211} \;=\;
  &
  y_{1}-y\ge \dRSS(v_{1},v)
  \land \vmin = v
 \end{array}
\end{align*} \end{minipage}
}
\caption{The subscenarios $\mathcal{T}_{w}
$ in Fig.~\ref{fig:subgoals} for the pull over scenario. See \cref{ex:subscenarioRefinePullOver}}
\label{fig:subscenarioDefPullOver}
\end{figure}

\subsection{Identifying Subscenario Proper Responses (\cref{line:subscenarioPropRespId})}\label{subsec:properRespIdSubscenario}
On \cref{line:subscenarioPropRespId}, for each subscenario $\mathcal{T}_{w}=(\Var,\Safe_{w}, \Env_{w}, \Goal_{w})$ in the subscenario tree $\mathcal{T}$,
we find $\dHL$ programs $\alpha_{w,1},\dotsc, \alpha_{w,K_{w}}$ such that each $\alpha_{w,i}$ achieves the goal $\Goal_{w}$ maintaining $\Safe_{w}\land\Env_{w}$ under a certain precondition. These programs  $\alpha_{w,1},\dotsc, \alpha_{w,K_{w}}$  are \emph{proper responses} for the subscenario $\mathcal{T}_{w}$. 

There can be a number of such proper responses: collecting more of them and thus being more comprehensive is  desired here,  but it is not mandatory. As we will shortly see in \cref{subsec:collectAndComputeGlobal}, missing some proper responses may lead to a stronger precondition (i.e.\ a stronger RSS condition, \cref{def:GARSSRule}) than necessary, but the resulting precondition may still be weak enough to be useful.

 The above requirement on proper responses $\alpha_{w,i}$---that they ``achieve $\Goal_{w}$ maintaining $\Safe_{w}\land\Env_{w}$''---is made precise as follows.
\begin{equation}\label{eq:desiredPropResp}
\qquad
\begin{minipage}{.4\textwidth}
  Under some precondition $A_{w,i}$, the Hoare quadruple $\hquad{\Safe_{w}\land\Env_{w}}{A_{w,i}}{\alpha_{w,i}}{\Goal_{w}}$ should be valid. Moreover, it is desired that $A_{w,i}$ is weak.
\end{minipage}
\end{equation}
Note that this is not a mathematical condition---while weak $A_{w,i}$ is desired, nothing prevents to have $\false$ as $A_{w,i}$, in which case any program qualifies as a proper response $\alpha_{w,i}$. 
However, finding ``better''  $\alpha_{w,i}$ leads to weaker (and   more widely applicable)  RSS conditions. See \cref{subsec:collectAndComputeGlobal}.

We allow the proper responses  $\alpha_{w,1},\dotsc, \alpha_{w,K_{w}}$ to have syntactic parameters  $F, G, \dotsc$; they are instantiated by concrete expressions later on \cref{line:backProp}. The use of this flexibility is demonstrated below in \cref{ex:properRespIdPullOver,ex:backPropPullOver}. 

\begin{example}\label{ex:properRespIdPullOver}
Continuing \cref{ex:subscenarioRefinePullOver}, for each subscenario $\mathcal{T}_{w}$ (\cref{fig:subscenarioDefPullOver}), we aim to find  proper responses $\alpha_{w,1},\dotsc, \alpha_{w,K_{w}}$, whose preconditions $A_{w,i}$ are weak  (cf.~\cref{eq:desiredPropResp}). The outcome is illustrated in \cref{fig:maneuverSeqs}.

\underline{\bfseries
The subscenario $\mathcal{T}_{1}$
}
We have to stop at a desired position $\ytgt$ while driving in a single lane. A sensible program $\alpha_{1, 1}$ that achieves it is to 1) first cruise with the initial velocity until braking is needed, and 2) then engage the maximum comfortable braking (i.e.\ at the rate $\bmin$) until the vehicle comes to a halt. Formally,
\begin{equation}\label{eq:alpha1}
   \alpha_{1, 1} =
    \left(\footnotesize
      \begin{array}{l}
        a := 0;\\
        \dwhileClause{\frac{v^{2}}{2 \bmin} < \ytgt - y}{\odeClause{y}{v}, \odeClause{v}{a}};\\
        a := -\bmin;\\
        \dwhileClause{v>0}{\odeClause{y}{v}, \odeClause{v}{a}}
      \end{array}
    \right).
\end{equation}
The switching point is where \SV{}'s position $y$ is $\ytgt - \frac{v^{2}}{2 \bmin}$.  We came up with this condition by high-school maths;  its correctness is confirmed later on \cref{line:backProp}. 

We can also include other programs as  proper responses $\alpha_{1, i}$---such as ones that brake more gently. We do not do so in this paper, since $\alpha_{1, 1}$ in the  above is the most powerful when it comes to goal achievement (namely, to stop at $\ytgt$). 

\underline{\bfseries
 The subscenario $\mathcal{T}_{11}$
} The goal here is to change lanes, and it can be achieved by different longitudinal manoeuvre sequences: cruise; cruise and brake; accelerate; accelerate and cruise; etc. A general approach would be to include all these manoeuvre sequences as proper responses $\alpha_{11,1},\dotsc, \alpha_{11,K_{11}}$. 

Among these possible proper responses, the ``cruise-brake'' one is the most relevant, given that our goal later is to stop at a given position. 
For simplicity, we only consider this proper response:
\begin{equation}\label{eq:alpha11}
   \alpha_{11,1} =
    \left(\footnotesize
      \begin{array}{l}
        l := 2.5; \; t:=0;\; a:=0;\\
        \dwhileClause{F_{11,1} > 0 \land t < \tlc}{\odeClause{t}{1}, \odeClause{y}{v}, \odeClause{v}{a}};\\
       a:=-\bmin;\\
        \dwhileClause{t < \tlc}{\odeClause{t}{1}, \odeClause{y}{v}, \odeClause{v}{a}};\\
        l := 3
      \end{array}
    \right).
\end{equation}
  \todooptil{Clovis: This program cannot be used ``along the way'' (i.e.,
  when the lane change has already started), since $\tlc$ is a
  constant and $t$ gets reset.
 
  Ichiro: Explaining your point will necessitate the explanation of our handling of $t$, which I suppose is too much. I hope that the sentence ``For simplicity, ...'' allows us to suppress your point.
}
Here, the change of lanes is indicated by the assignments $l := 2.5$ and  $l := 3$. The constant $\tlc$ stands for the maximum time needed for changing lanes; we use $\tlc = 3$ seconds as an estimate (see e.g.~\cite{AtaelmananPH21}). Note that assuming a larger $\tlc$ means 1) \SV{} occupies two lanes longer and 2) it takes longer to reach the destination lane, and thus makes analysis more conservative.  

The switching point is harder to find here than for $\mathcal{T}_{1}$ in the above---we therefore leave it as a syntactic parameter $F_{11,1}$. It is instantiated later on Line~\ref{line:backProp}.

\underline{\bfseries 
The subscenario $\mathcal{T}_{12}$
} 
By the same reasoning, we define
\begin{equation*}
   \alpha_{12,1} =
    \left(\footnotesize
      \begin{array}{l}
       l := 2.5; \; t:=0; \; a := 0;\\
        \dwhileClause{F_{12,1} > 0 \land t< \tlc}{\odeClause{t}{1}, \odeClause{y}{v}, \odeClause{v}{a}};\\
        a :=-\bmin; \\
        \dwhileClause{t < \tlc}{\odeClause{t}{1}, \odeClause{y}{v}, \odeClause{v}{a}};\\
        l := 3
      \end{array}
    \right).
\end{equation*}
Note that $F_{12,1}$ will be instantiated with a different expression from $F_{11,1}$, since they are constrained by different \POV{}s (namely, \POV{1} as the immediate preceding vehicle for the former, and \POV{2} for the latter). 

\underline{\bfseries 
The subscenarios $\mathcal{T}_{111}, \mathcal{T}_{121}$
} By the same reasoning as above, we define proper responses $\alpha_{111,1}, \alpha_{121,1}$ to be the same as~\cref{eq:alpha11}, using different syntactic parameters such as $F_{111,1}$. 

\underline{\bfseries The subscenario $\mathcal{T}_{1111}$} The goal here is to prepare for merging between \POV{2} and \POV{1}, by making enough distances in front (from \POV{2}) and behind (from \POV{1}) and matching the velocity with the preceding \POV{2}, while driving in Lane~1. See \cref{fig:subscenarioDefPullOver}.  This may be achieved by various longitudinal manoeuvre sequences. We choose the following four, which we believe constitutes a  quite comprehensive list.
\begin{itemize}
 \item ($\alpha_{1111,1}$: accel-brake) Accelerate, at the rate $\amax$, to make enough distance behind (from \POV{1}). Then brake in order to match the velocity with the preceding \POV{2}.
 \item ($\alpha_{1111,2}$: accel-cruise-brake) Similar to accel-brake, but in case \SV{}'s velocity reaches the legal maximum during the acceleration manoeuvre, \SV{} cruises until it has to brake.
 \item ($\alpha_{1111,3}$: accel) Accelerate only (at the  rate $\amax$). This is used when  \SV{} is initially slower than \POV{2}.
 \item ($\alpha_{1111,4}$: brake) Brake only (at the maximum comfortable rate $\bmin$). This is used when  \SV{} is initially faster than \POV{2}.
\end{itemize}


\underline{\bfseries The subscenario $\mathcal{T}_{1211}$} The goal $\Goal_{1211}$ here is to prepare for merging behind \POV{1}. For ease of logical reasoning later, we require that \SV{}'s velocity should be the legal minimum at the end ($\vmin = v$)---we did so already in~\cref{eq:subgoalsPullOver}. This requirement may delay the goal achievement (stopping at $\ytgt$ in Lane~3) by travelling slowly, but it does not reduce the possibility of the goal achievement. 
\todooptil{Clovis: but it does not make it slower in the simplex
architecture.

Ichiro: I tried to add some explanation but it took too much space and were disruptive. Sorry
}

The goal $\Goal_{1211}$ may be achieved by various longitudinal manoeuvre sequences, but  those which involve acceleration are obviously redundant. This leaves us with the following two proper responses. 
\begin{itemize}
 \item ($\alpha_{1211,1}$: brake-cruise) Brake until $v$ is as small as $\vmin$, and then cruise at $\vmin$ for the time needed to make enough distance in front (from \POV{1}). 
 \item ($\alpha_{1211,2}$: brake) Brake only. This manoeuvre is used when \SV{} is initially sufficiently behind \POV{1}, in which case braking until $\vmin=v$ already makes enough distance from \POV{1}.
\end{itemize}
Note again that there are other possible proper responses.  The above list is nevertheless  comprehensive enough and thus provide a useful RSS rule with a weak RSS condition. 
\end{example}

\begin{remark}[basic maneuvers]\label{rem:basicManeuvers}
The proper responses  in~\cref{ex:properRespIdPullOver} are composed of several \emph{basic manoeuvres}, namely
\begin{itemize}
 \item to \emph{cruise} ($a:=0;\,\dwhileHeader{A}\;\{\,\odeClause{y}{v}, \odeClause{v}{a}\,\}$),
 \item to \emph{brake} ($a:=-\bmin;\,\dwhileHeader{A}\;\{\,\odeClause{y}{v}, \odeClause{v}{a}\,\}$), 
 \item to \emph{accelerate} ($a:=\amax;\,\dwhileHeader{A}\;\{\,\odeClause{y}{v}, \odeClause{v}{a}\,\}$),
 \item to \emph{initiate lane change} (such as $l:= 2.5$), and
 \item to \emph{complete lane change} (such as $l:=3$). 
\end{itemize}
Restriction to this limited vocabulary is not mandated by our framework. Still we find it useful because 1) the logical reasoning later on \cref{line:backProp} can be modularised along basic manoeuvres (see \cref{subsec:backProp}), and 2) basic manoeuvres are easy to implement in a baseline controller (see \cref{subsec:impl}). 
\end{remark}

\subsection{Identifying Subscenario Preconditions (\cref{line:backProp})}\label{subsec:backProp}
In this step, we identify \emph{subscenario preconditions}---preconditions for subscenario proper responses $\alpha_{w,i}$ that  we identified on \cref{line:subscenarioPropRespId}. A subscenario precondition must guarantee, 
after the execution of the proper response $\alpha_{w,i}$ in question,
\begin{itemize}
 \item not only
 the achievement of the subscenario goal $\Goal_{w}$, 
 \item but also  the precondition of the next proper response $\alpha_{w',i'}$ (where $w=w'j$ with some $j$).
\end{itemize}
 The latter requirement is inductive: a subscenario precondition for $\alpha_{j_{1}j_{2}j_{3}}$ is constrained by one for $\alpha_{j_{1}j_{2}}$, which is further constrained by one for $\alpha_{j_{1}}$, etc. This forces us to identify subscenario preconditions \emph{backwards}. Such backward reasoning is common in program verification; see e.g.~\cite{Winskel93}. 

Because of this backward reasoning, too, we identify subscenario preconditions for each \emph{sequence} of subscenario proper responses, instead of for each subscenario proper response. This is made precise in the following definition.

\begin{definition}[backward condition propagation]\label{def:backProp}
 Let $\mathcal{S}$ be a scenario,  $\mathcal{T}$ be  a subscenario tree for $\mathcal{S}$, and $\alpha_{w,1},\dotsc, \alpha_{w,K_{w}}$ be proper responses for each subscenario $\mathcal{T}_{w}$ in $\mathcal{T}$.

On~\cref{line:backProp} of Procedure~\ref{alg:workflow}, we identify an assignment $(A_{w,u})_{w,u}$. Specifically,
 \begin{itemize}
  \item to each node $w=j_{1} j_{2}\dotsc j_{k}$ of $\mathcal{T}$ and each sequence $u=i_{1} i_{2}\dotsc i_{k}$ of proper response indices (where 
$
i_{1}\in [1,K_{j_{1}}], 
i_{2}\in [1,K_{j_{1}j_{2}}],
\dotsc,
i_{k}\in[1,K_{j_{1}\dotsc j_{k}}]
$, cf.\ \cref{notation:integerInterval}),
  \item we assign a $\dHL$ assertion $A_{w,u}=A_{j_{1} j_{2}\dotsc j_{k}, i_{1} i_{2}\dotsc i_{k}}$,

 \end{itemize}
 so that the assignment satisfies the following condition~\cref{eq:backPropAssignmentCond}. 
\begin{equation}\label{eq:backPropAssignmentCond}
\boxed{
\begin{minipage}{.4\textwidth}
For each $k\in[0,N-1]$, $j_{k+1}$, $i_{k+1}$,  $w = j_{1} \dotsc j_{k}$, $u= i_{1} \dotsc i_{k}$, 
the $\dHL$ quadruple
\begin{displaymath}
\begin{aligned}
   \{
 A_{w j_{k+1}, u i_{k+1}} 
 \}\,
 \alpha_{w j_{k+1},i_{k+1}}
 \,
 \{
 \Goal_{w j_{k+1}}\land  A_{w , u} 
 \}
 \\ 
 \colon{}
  \Safe_{w j_{k+1}} \land \Env_{w j_{k+1}}
\end{aligned}
\end{displaymath}
is valid. 
 \end{minipage}
}
\end{equation}
Here we set, as a convention, $A_{\varepsilon,\varepsilon}=\true$ for $k=0$.
Note that the condition~\cref{eq:backPropAssignmentCond} is defined inductively on $k$; it therefore forces us to choose $A_{w,u}$ for shorter $w,u$ first.

\end{definition}

Note that the definition does not uniquely determine the assignment $(A_{w,u})_{w,u}$: the assignment is only constrained by the condition~\cref{eq:backPropAssignmentCond}; there are generally many ways to satisfy~\cref{eq:backPropAssignmentCond}. We aim at weaker preconditions---as is usual in program verification---so that the resulting RSS rule is applicable to wider situations.

\begin{example}\label{ex:backPropPullOver}
We continue \cref{ex:properRespIdPullOver} and identify subscenario preconditions $A_{1,1}, A_{11,11}, A_{12, 11}, \dotsc$ in a backward manner. These preconditions and their relationships---such as the one required in~\cref{eq:backPropAssignmentCond}---are illustrated in \cref{fig:propagation}.

\underline{\bfseries The subscenario $\mathcal{T}_{1}$}
We aim at $A_{1,1}$ that satisfies
\begin{equation}\label{eq:A1_1Req}
 \hquad{l=3\land \Env}{A_{1,1}}{\alpha_{1,1}}{y=\ytgt\land v=0}
\end{equation}
where $\Env$ is from \cref{ex:scenarioModelingPullover} and $\alpha_{1,1}$ is from~\cref{eq:alpha1} (see also \cref{fig:subscenarioDefPullOver}). It turns out that 
\begin{equation}\label{eq:A1_1}
 A_{1,1}
 =
    \left(
      \begin{array}{l}
        \Env\land
        l=3 \land v>0 
        \land\frac{v^{2}}{2 \bmin} \le \ytgt -y
      \end{array}
    \right)
\end{equation}
satisfies~\cref{eq:A1_1Req}. A $\dHL$ proof is in \cref{fig:sub4}.

\begin{figure*}[tbp]
\scriptsize
\begin{align}
 &
 \begin{aligned}
 &
    \bigquadnnl{
      \begin{array}{l}
        \Env\land l=3\land  0 < v \le \vmax\\{}\land
           -\bmin\le a \le \amax\\{}\land
        \ytgt - y - \frac{v^{2}}{2 \bmin} \geq 0
      \end{array}
    }{
      \begin{array}{l}
        \Env\land l=3\land 0 < v \le \vmax\\{}\land
        \ytgt - y - \frac{v^{2}}{2 \bmin} \geq 0
      \end{array}
    }{
      \begin{array}{l}
        \dwhileHeader{\frac{v^{2}}{2 \bmin} < \ytgt - y}\\
        \quad\odeClause{y}{v},
        \,\odeClause{v}{0}
      \end{array}
    }{
      \begin{array}{l}
        \Env\land l=3\land 0 < v \le \vmax\\{}\land
        \ytgt - y - \frac{v^{2}}{2 \bmin} = 0
      \end{array}
    }
 \\
 & \hspace{5em} 
   \text{by $(\dwhilerule)$ with $(\invariant \sim 0)=(v>0)$, $\variant= \ytgt - y - \frac{v^{2}}{2 \bmin}$, $\terminator= -v$}
 \end{aligned}
 \label{eq:sub4pf1}
 \\[-.6em]
 &\begin{minipage}{\textwidth}
  \dotfill
 \end{minipage}
 \nonumber
 \\[-.6em]
&
 \begin{aligned}
 &
    \bigquadnnl{
      \begin{array}{l}
        \Env\land l=3\land 
	 0\le v\le\vmax\\{}\land
           -\bmin\le a \le \amax\\{}\land
        y\le \ytgt
      \end{array}
    }{
      \begin{array}{l}
        \Env\land l=3\land 0 < v \le \vmax\\{}\land
         \frac{v^{2}}{2 \bmin} \le \ytgt - y 
      \end{array}
    }{
      \begin{array}{l}
        \dwhileHeader{\frac{v^{2}}{2 \bmin} < \ytgt - y}\\
        \quad\odeClause{y}{v},
        \,\odeClause{v}{0}
      \end{array}
    }{
      \begin{array}{l}
        \Env\land l=3\land 0 < v \le \vmax\\{}\land
        \frac{v^{2}}{2 \bmin} = \ytgt - y 
      \end{array}
    }
 \\
 & \hspace{5em} 
   \text{by $(\limprule)$ and~\cref{eq:sub4pf1}}
 \end{aligned}
 \label{eq:sub4pf2}
 \\[-.6em]
 &\begin{minipage}{\textwidth}
  \dotfill
 \end{minipage}
 \nonumber
 \\[-.6em]
&
 \begin{aligned}
 &
    \bigquadnnl{
      \begin{array}{l}
        \Env\land l=3\land 0 < v \le \vmax\\{}\land
           -\bmin\le a \le \amax\\{}\land
         \frac{v^{2}}{2 \bmin} =\ytgt - y 
      \end{array}
    }{
      \begin{array}{l}
        \Env\land l=3\land 0 < v \le \vmax\\{}\land
        \frac{v^{2}}{2 \bmin} =\ytgt - y 
      \end{array}
    }{
      \begin{array}{l}
        \dwhileHeader{v>0}\\
        \quad\odeClause{y}{v},
        \,\odeClause{v}{-\bmin}
      \end{array}
    }{
      \begin{array}{l}
        \Env\land l=3\land v = 0\\{}\land
        \frac{v^{2}}{2 \bmin} =\ytgt - y 
      \end{array}
    }
 \\
 & \hspace{5em} 
   \text{by $(\dwhilerule)$ with $(\invariant \sim 0)=(\ytgt - y - \frac{v^{2}}{2\bmin}=0)$, $\variant= v$, $\terminator= -\bmin$}
 \end{aligned}
 \label{eq:sub4pf3}
 \\[-.6em]
 &\begin{minipage}{\textwidth}
  \dotfill
 \end{minipage}
 \nonumber
 \\[-.6em]&
 \begin{aligned}
 &
    \bigquadnnl{
      \begin{array}{l}
        \Env\land l=3\land
	 0\le v\le\vmax\\{}\land
           -\bmin\le a \le \amax\\{}\land
	   y\le \ytgt
      \end{array}
    }{
      \begin{array}{l}
        \Env\land l=3\land 0 < v \le \vmax\\{}\land
        \frac{v^{2}}{2 \bmin} =\ytgt - y 
      \end{array}
    }{
      \begin{array}{l}
        \dwhileHeader{v>0}\\
        \quad\odeClause{y}{v},
        \,\odeClause{v}{-\bmin}
      \end{array}
    }{
      \begin{array}{l}
        \Env\land l=3\land v = 0\\{}\land
        y =\ytgt 
      \end{array}
    }
 \\
 & \hspace{5em} 
   \text{by $(\limprule)$ and~\cref{eq:sub4pf3}}
 \end{aligned}
 \label{eq:sub4pf4}
 \\[-.6em]
 &\begin{minipage}{\textwidth}
  \dotfill
 \end{minipage}
 \nonumber
 \\[-.6em]&
 \begin{aligned}
 &
    \bigquadnnl{
      \begin{array}{l}
        \Env\land l=3\land
	 0\le v \le \vmax\\{}\land
           -\bmin\le a \le \amax\\{}\land
	   y\le \ytgt
      \end{array}
    }{
      \begin{array}{l}
        \Env\land l=3\land 0 < v \le \vmax\\{}\land
        \frac{v^{2}}{2 \bmin} \le\ytgt - y 
      \end{array}
    }{
      \begin{array}{l}
        \dwhileHeader{\frac{v^{2}}{2 \bmin} < \ytgt - y}\;\bigl\{\, \odeClause{y}{v}, \odeClause{v}{0}\,\bigr\};\\
        \dwhileHeader{v>0}\;\bigl\{\, \odeClause{y}{v}, \odeClause{v}{-\bmin}\,\bigr\}
      \end{array}
    }{
      \begin{array}{l}
        \Env\land l=3\land v = 0\\{}\land
        y =\ytgt 
      \end{array}
    }
 \\
 & \hspace{5em} 
   \text{by $(\seqrule)$,~\cref{eq:sub4pf2} and~\cref{eq:sub4pf4}}
 \end{aligned}
 \label{eq:sub4pf5}
 \end{align}
\caption{A $\dHL$ proof for $\hquad{\Safe_{1}\land\Env_{1}}{A_{1,1}}{\alpha_{1,1}}{\Goal_{1}}$, \cref{ex:backPropPullOver}. Here we use the obvious constant substitution, replacing  $\bigl(\,a:=0;\,\dwhileHeader{A}\;\{\,\odeClause{y}{v}, \odeClause{v}{a}\,\}\,\bigr)$ with $\bigl(\,\dwhileHeader{A}\;\{\,\odeClause{y}{v}, \odeClause{v}{0}\,\}\,\bigr)$, for example. The dynamics of \POV{}s is not explicit here; see \cref{rem:POVDyn}. The derivation of~\eqref{eq:sub4pf1}
and~\eqref{eq:sub4pf3}  in fact requires a more general form of $(\dwhilerule)$ than we presented in \cref{fig:dFHL-rules} (namely the ``multiple-invariant multiple-variant'' one in \cref{fig:dwhile}, \cref{app:dRSS}); see \cref{rem:extensionOfWandDW}.}
\label{fig:sub4}
\end{figure*}


\underline{\bfseries The subscenario $\mathcal{T}_{11}$}
We aim at $A_{11,11}$ that satisfies
\begin{equation}\label{eq:A11_1Req}
 \hquad{\Safe_{11}\land \Env_{11}}{A_{11,11}}{\alpha_{11,1}}{l=3 \land A_{1,1}}
\end{equation}
where $\Safe_{11}$ and $\Env_{11}$ are from \cref{fig:subscenarioDefPullOver} and $\alpha_{11,1}$ is from~\cref{eq:alpha11}.
Note that $A_{1,1}$ that we found in~\cref{eq:A1_1} is now part of the postcondition.

It turns out that the definition of $\Safe_{11}$, $\Env$ and $\alpha_{11,1}$ simplifies the reasoning a lot: for example, the safety condition $y_{2}-y\ge \dRSS(v_{2},v)$ is obviously preserved in the course of the dynamics since we require $v\le v_{2}$  in $\Safe_{11}$; moreover, $v\le v_{2}$ is preserved since $\alpha_{11,1}$ can brake or cruise but never accelerates.
Not imposing legal minimum speed on \SV{} (\cref{ex:scenarioModelingPullover}) simplifies the reasoning too. 

In the end, we arrive at the following precondition, for which we can prove~\cref{eq:A11_1Req}. 
\begin{equation*}
 A_{11,11}
 =
    \left(
      \begin{array}{l}
        \Env\land
        l=2 \land 0 < v \le v_{2}
        \land{}
        \\
        y_{2}-y\ge \dRSS(v_{2},v) \land
        \frac{v^{2}}{2 \bmin} \le \ytgt -y
      \end{array}
    \right),
\end{equation*}
Here we instantiate $F_{11,1}$ in~\cref{eq:alpha11} with $ \ytgt -y -\frac{v^{2}}{2 \bmin}$. 
\todooptil{Clovis: we need to say something about $l=2$ in the
precondition.  It's not needed by the reasoning, but imposed by
$\mathcal{T}_{111}$ (?) in the subscenario tree and maybe some meta
reasoning.\\
Ichiro: this touches upon the modeling of lateral movement, which is currently left vague. So I'd say we keep the origin of $l=2$ vague too. It is also a natural condtion to write.}

The key inequality here is $\frac{v^{2}}{2 \bmin} \le \ytgt -y$, much like in~\cref{eq:A1_1}. This is not surprising: ignoring lateral movements and the leading vehicle \POV{2} (we can do so by the simplification discussed above),  goal achievement depends solely on whether the braking is in time.

\underline{\bfseries The subscenarios $\mathcal{T}_{12}, \mathcal{T}_{111}, \mathcal{T}_{121}$}
Similarly to $\mathcal{T}_{11}$, we choose preconditions $A_{12,11}, A_{111, 111}, A_{121, 111}$, while instantiating syntactic parameters $F_{12,11}, F_{111, 111}, F_{121, 111}$ in the proper responses.

\underline{\bfseries The subscenario $\mathcal{T}_{1111}$} 
We identified four proper responses $\alpha_{1111,1},\dotsc,\alpha_{1111,4}$ in \cref{ex:properRespIdPullOver}. Here we focus on $\alpha_{1111,2}$ (accel-cruise-brake)---it is the most complicated---and identify the corresponding precondition $A_{1111,1112}$. The reasoning below subsumes those for the other three proper responses. 

The proper response $\alpha_{1111,2}$ accelerates until the legal maximum speed $\vmax$, cruises at $\vmax$ in order to increase the distance behind (from \POV{1}), and brakes to match its velocity with \POV{2}. There are two switching points. 
\begin{itemize}
 \item The one from acceleration to cruising---its timing  is easily determined by $v<\vmax$ or not.
 \item The one from cruising to braking---its timing is decided so that, at the end of braking (when $v=v_{2}$), the distance behind (from \POV{1}) is precisely the required RSS safety distance $\dRSS(v,v_{1})$. 
\end{itemize}
These arguments can easily be translated to symbolic conditions, which are used to instantiate symbolic parameters in $\alpha_{1111,2}$. It is also easy to symbolically express the positions and velocities of \SV{} and \POV{}s at the end of the proper response. 

Now, the precondition $A_{1111,1112}$ must be such that
$\hquad{\Safe_{1111} \land \Env_{1111}}{\asserta_{1111,1112}}{\coma_{1111,2}}{\Goal_{1111} \land \asserta_{111,111}}$
is valid.
In other words, we must address the following concerns.
\begin{itemize}
 \item  The subscenario goal
 $\Goal_{1111}=\bigl(
  y_{2}-y\ge \dRSS(v_{2},v)\land y-y_{1}\ge\dRSS(v,v_{1})\land v_{2}=v
\bigr)$  (see \cref{fig:subscenarioDefPullOver}) as part of the postcondition. The latter two conjuncts are trivially satisfied by the above design of the proper response; %
    therefore $y_{2}-y\ge \dRSS(v_{2},v)$ is a core part of the postcondition. Using the analytic solution of the proper response, the last postcondition is easily translated to a precondition on the initial positions, velocities, etc. 
 \item The precondition $A_{111,111}$ of the next subscenario $\mathcal{T}_{111}$, as part of the postcondition. Much like for $A_{11,11}$ (discussed above), the key inequality in $A_{111,111}$ is again $\frac{v^{2}}{2 \bmin} \le \ytgt -y$---this is imposed ultimately to ensure that \SV{} does not overshoot the stopping position $\ytgt$.
   The requirement of this inequality as a postcondition can easily be translated to a precondition, too.
 \item The condition $y_{3}-y\ge \dRSS(v_{3},v)$ as part of the safety condition. Again, using the analytic solution of the proper response, it is easy to calculate a precondition that guarantees this safety condition. The reasoning here is much like for the original RSS proof~\cite{ShalevShwartzSS17RSS} that the RSS safety distance is enough for collision avoidance (\cref{ex:onewayTraffic}). 
\end{itemize}
We define $A_{1111,1112}$ as the conjunction of the three preconditions that come from the above concerns. It requires enough distances from $\ytgt$ and \POV{2}--3, all formulated symbolically in terms of the vehicles' initial positions and velocities. 

The above calculation of a precondition $A_{1111,1112}$ is theoretically straightforward---an analysis of a quadratic dynamic system with some case distinctions. It is nevertheless laborious, with logical assertions easily blowing up to dozens of lines. We use Mathematica to manage the necessary symbolic manipulations, such as solving quadratic equations, substitution, and tracking case distinctions. See \cref{sec:toolSupport} for further discussion.

\underline{\bfseries The subscenario $\mathcal{T}_{1211}$}
We identified two proper responses  $\alpha_{1211,1},\alpha_{1211,2}$ in \cref{ex:properRespIdPullOver}. Preconditions $A_{1211,1111}, A_{1211,1112}$ for those can be found much like in the above (for $\mathcal{T}_{1211}$)---it is much easier since $\mathcal{T}_{1211}$ does not have  requirements at odds, such as $  y_{2}-y\ge \dRSS(v_{2},v)\land y-y_{1}\ge\dRSS(v,v_{1})$. 

In fact,  the subscenario $\mathcal{T}_{1211}$ and the subsequent ones in the subscenario tree (namely $\mathcal{T}_{121}, \mathcal{T}_{12}, \mathcal{T}_{1}$) are so simple that we can automate the whole task of identification of subscenario proper responses (\cref{line:subscenarioPropRespId}) and preconditions (\cref{line:backProp}). This partial automation will be presented in another venue.
\end{example}

\subsection{Global Proper Response and Precondition (\cref{line:collectAndComputeGlobal})}
\label{subsec:collectAndComputeGlobal}
The  goal of Procedure~\ref{alg:workflow} is to find $A$ (an RSS condition) and $\alpha$ (a proper response) such that $ \hquad{\Safe\land\Env}{A}{\alpha}{\Goal}$ is valid (\cref{subsec:problemRulesAsTriples}). In this last step of Procedure~\ref{alg:workflow}, 
\begin{itemize}
 \item to obtain $\alpha$,  we combine the proper responses $\alpha_{w,1},\dotsc, \alpha_{w,K_{w}}$ we have identified for different subscenarios $\mathcal{T}_{w}$,  and
 \item compute a collective precondition $A$.
\end{itemize}
We do so using the $\dHL$ rules---especially the $(\seqrule)$ and $(\caserule)$ rules, see \cref{fig:dFHL-rules} and~\cref{eq:caserule}. 

\begin{definition}[global proper response and precondition]\label{def:collectAndComputeGlobal}
Using the subscenario proper responses $(\alpha_{w,i})_{w,i}$ and subscenario preconditions $(A_{w,u})_{w,u}$ obtained on \cref{line:subscenarioPropRespId,line:backProp},
we define the \emph{global proper response} $\alpha$ and the \emph{global precondition} $A$ as follows. 
\begin{equation}\label{eq:globalPrecondAndPropRes}
 \begin{aligned}
 \alpha &= \caseKeyword \,(A_{w, u})_{%
   w=j_{1}\dotsc j_{k},u=i_{1}\dotsc i_{k}
  }\, 
 \\
 &\qquad\qquad
   \alpha_{j_{1}\dotsc j_{k},i_{k}};\,
   \alpha_{j_{1}\dotsc j_{k-1},i_{k-1}};\;
   \cdots\, ;\,
   \alpha_{j_{1},i_{1}}\;,
 \\
  A &= 
  \textstyle\bigvee_{w=j_{1}\dotsc j_{k},u=i_{1}\dotsc i_{k}}A_{j_{1}\dotsc j_{k}, i_{1}\dotsc i_{k}}.
\end{aligned}
\end{equation}
Finally, $A$ and $\alpha$ are returned as the outcome of \cref{line:backProp}. 
\end{definition}
In~\cref{eq:globalPrecondAndPropRes},  the case distinction and the disjunction range over all $w$ and $u$ considered earlier. That is, 
\begin{itemize}
 \item every word $w=j_{1} j_{2}\dotsc j_{k}$ that designates a node of
 $\mathcal{T}$ (the node $\mathcal{T}_{w}$ need not be a leaf), and
 \item all index sequences $u= i_{1} \dotsc i_{k}$ compatible with $w$ (meaning $
i_{1}\in [1,K_{j_{1}}], 
\dotsc,
i_{k}\in[1,K_{j_{1}\dotsc j_{k}}]
$, as above, cf.\ \cref{notation:integerInterval}).
\end{itemize}
See \cref{ex:backPropPullOver} and \cref{fig:propagation} for an example.

The following is our main theorem; it states that the above outcome indeed achieves the specified goal while maintaining safety. Our framework---including the design of $\dHL$---has been carefully designed so that its proof is straightforward.
\begin{theorem}[correctness of Procedure~\ref{alg:workflow}]\label{thm:correctness}
 In Procedure~\ref{alg:workflow}, the outcome $(A,\alpha)$ of  \cref{line:backProp} (\cref{def:backProp}) is a goal-aware RSS rule for $\mathcal{S}$ (\cref{def:GARSSRule}), making the $\dHL$ quadruple $\hquad{\Safe\land\Env}{A}{\alpha}{\Goal}$ valid.
\end{theorem}
\begin{proof}
The proof is shown in \cref{fig:correctnessPf}. It builds upon the assumption~\cref{eq:backPropAssignmentCond} on the precondition $A_{w,u}$ that we identified for each subscenario proper response $\alpha_{w,i}$. It also relies crucially on \cref{def:subscenario,def:subscenarioTree})---we require $\Safe_{w}\land\Env_{w}\Rightarrow \Safe$ for each subscenario $\mathcal{T}_{w}$ in $\mathcal{T}$ on~\cref{line:subscenarioRefine}. 
\end{proof}
We note that,  in the last proof, the subgoals $\Goal_{w}$ with $|w|>1$ play no role.
They are useful, however, in designing subscenarios (especially choosing $\Safe_{w}$ and $\Env_{w}$ in \cref{subsubsec:subscenarioRef}) and identifying proper responses (\cref{subsec:properRespIdSubscenario}). In other words, the subgoals $\Goal_{w}$ play the role of glue in our compositional workflow.

The proof does not use the $(\assignrule)$, $(\whilerule)$, and $(\dwhilerule)$ rules. They are used for establishing the assumption~\cref{eq:backPropAssignmentCond} for each subscenario proper response $\alpha_{w,i}$. See
e.g.\ \cref{fig:sub4}.

\begin{figure*}[tbp]
 \begin{align}
 &
 \begin{aligned}
 & \bigl\{
  A_{j_{1}\dotsc j_{k+1}, i_{1}\dotsc i_{k+1}}
 \bigr\}\,
   \alpha_{j_{1}\dotsc j_{k+1},i_{k+1}}
 \,\{\Goal_{j_{1}\dotsc j_{k+1}} \land   A_{j_{1}\dotsc j_{k}, i_{1}\dotsc i_{k}}\}
 \;\colon{}\;
 \Safe_{j_{1}\dotsc j_{k+1}}\land\Env_{j_{1}\dotsc j_{k+1}}
 \quad
 \\
 &\hspace{5em}
 \text{for each
 $j_{1}\dotsc j_{k+1}, i_{1}\dotsc i_{k+1}$
}
 \qquad
 \text{(By condition~\cref{eq:backPropAssignmentCond})}
 \end{aligned}
 \label{eq:corr30}
 \\[-.6em]
 &\begin{minipage}{\textwidth}
  \dotfill
 \end{minipage}
 \nonumber
 \\[-.6em]&
 \begin{aligned}
 &
  \bigl\{
  A_{j_{1}\dotsc j_{k+1}, i_{1}\dotsc i_{k+1}}
 \bigr\}\,
   \alpha_{j_{1}\dotsc j_{k+1},i_{k+1}}
 \,\{
     A_{j_{1}\dotsc j_{k}, i_{1}\dotsc i_{k}}\}
 \;\colon{}\;
 \Safe\land\Env
 \qquad
 \text{for each 
 $j_{1}\dotsc j_{k+1}, i_{1}\dotsc i_{k+1}$
}
 \\
 &\hspace{5em}
   \text{(By $(\limprule)$,~\cref{eq:corr30}, 
 $\Goal_{j_{1}\dotsc j_{k+1}} \land
     A_{j_{1}\dotsc j_{k}, i_{1}\dotsc i_{k}} \Rightarrow A_{j_{1}\dotsc j_{k}, i_{1}\dotsc i_{k}}$,        
  }
 \\
  &\hspace{5em}
 \text{
    $\Safe_{j_{1}\dotsc j_{k+1}}\land\Env_{j_{1}\dotsc j_{k+1}}\Rightarrow\Safe$,  
   and
      $\Safe_{j_{1}\dotsc j_{k+1}}\land\Env_{j_{1}\dotsc j_{k+1}}\Rightarrow\Env$,
  see \cref{def:subscenario,def:subscenarioTree}
)}
 \end{aligned}
 \label{eq:corr305}
 \\[-.6em]
 &\begin{minipage}{\textwidth}
  \dotfill
 \end{minipage}
 \nonumber
 \\[-.6em]&
 \begin{aligned}
  \bigl\{
  A_{j_{1}, i_{1}}
 \bigr\}\,
   \alpha_{j_{1},i_{1}}
 \,\{\Goal_{j_{1}}
\}
 \;\colon{}\;
 \Safe_{j_{1}}\land\Env_{j_{1}}
 \quad
 \text{for each $j_{1}, i_{1}$ }
 \qquad
 \text{(\cref{eq:corr30} with $k=0$. Recall that $A_{\varepsilon, \varepsilon}=\true$)}
 \end{aligned}
 \label{eq:corr31}
 \\[-.6em]
 &\begin{minipage}{\textwidth}
  \dotfill
 \end{minipage}
 \nonumber
 \\[-.6em]&
 \begin{aligned}
 &
  \bigl\{
  A_{j_{1}, i_{1}}
 \bigr\}\,
   \alpha_{j_{1},i_{1}}
 \,\{\Goal\}
 \;\colon{}\;
 \Safe_{j_{1}}\land\Env_{j_{1}}
 \quad
 \text{for each  $j_{1}, i_{1}$}
 \\
 &\hspace{5em}
   \text{(By  $(\limprule)$,~\cref{eq:corr31}, and
      $\Safe_{j_{1}}\land\Env_{j_{1}}\land\Goal_{j_{1}}\Rightarrow\Goal$, 
  see \cref{def:subscenarioTree}
       )}
 \end{aligned}
 \label{eq:corr315}
 \\[-.6em]
 &\begin{minipage}{\textwidth}
  \dotfill
 \end{minipage}
 \nonumber
 \\[-.6em]&
 \begin{aligned}
 &
  \bigl\{
  A_{j_{1}, i_{1}}
 \bigr\}\,
   \alpha_{j_{1},i_{1}}
 \,\{\Goal\}
 \;\colon{}\;
 \Safe\land\Env
 \quad
 \text{for each $j_{1}, i_{1}$ }
 \\ &\hspace{5em}
   \text{(By  $(\limprule)$,~\cref{eq:corr315}, 
      $\Safe_{j_{1}}\land\Env_{j_{1}}\Rightarrow\Safe$,  and
      $\Safe_{j_{1}}\land\Env_{j_{1}}\Rightarrow\Env$,
  see \cref{def:subscenario,def:subscenarioTree}
       )}
 \end{aligned}
 \label{eq:corr317}
 \\[-.6em]
 &\begin{minipage}{\textwidth}
  \dotfill
 \end{minipage}
 \nonumber
 \\[-.6em]&
 \begin{aligned}
 &
  \bigl\{
  A_{j_{1}\dotsc j_{k+1}, i_{1}\dotsc i_{k+1}}
 \bigr\}\,
   \alpha_{j_{1}\dotsc j_{k},i_{k}};\,
   \cdots ;\,
   \alpha_{j_{1},i_{1}}
 \,\bigl\{\Goal\bigr\}
 \;\colon{}\;
 \Safe\land\Env
 \quad
 \text{for each $j_{1}\dotsc j_{k+1}, i_{1}\dotsc i_{k+1}$ }
 \\ &\hspace{5em}
   \text{(Repeated application of $(\seqrule)$ to~\cref{eq:corr305} and~\cref{eq:corr317})}
 \end{aligned}
 \label{eq:corr35}
 \\[-.6em]
 &\begin{minipage}{\textwidth}
  \dotfill
 \end{minipage}
 \nonumber
 \\[-.6em]&
 \begin{aligned}
  \bigl\{
  \textstyle\bigvee_{w,u}A_{w, u}
 \bigr\}\,
 \begin{array}[c]{l}
  \caseKeyword \,(A_{w, u})_{w=j_{1}\dotsc j_{k},u=i_{1}\dotsc i_{k}}\, 
  \;
   \alpha_{j_{1}\dotsc j_{k},i_{k}};\,
   \cdots ;\,
   \alpha_{j_{1},i_{1}}
 \end{array} 
 \,\bigl\{\Goal\bigr\}
 \,\colon{}\,
 \Safe\land\Env
 \qquad
   \text{(By $(\caserule)$ and~\cref{eq:corr35})}
 \end{aligned}
\end{align}
\todooptil{Clovis: \cref{eq:corr315} and \cref{eq:corr317} in a single
step?\\

Ichiro: Perhaps, too lazy to fix...
}
\caption{Correctness proof for Procedure~\ref{alg:workflow} (\cref{thm:correctness})
}
\label{fig:correctnessPf}
\end{figure*}

\begin{example}\label{ex:globalPullOver}
Continuing \cref{ex:backPropPullOver}, for the pull over scenario in \cref{ex:scenarioModelingPullover}, we obtain a global proper response 
\begin{align*}
 \alpha =  
\left(
\vcenter{$\caseKeyword \footnotesize
\begin{array}[t]{ll}
 (A_{1,1})
&
 \alpha_{1,1}
\\
 (A_{11,11})
&
 \alpha_{11,1};\;\alpha_{1,1}
\\
 (A_{12,11})
&
 \alpha_{12,1};\;\alpha_{1,1}
\\
 \cdots
\\
 (A_{1111,1111})
&
\alpha_{1111,1};\;\alpha_{111,1};\;\alpha_{11,1};\;\alpha_{1,1}
\\
 (A_{1111,1112})
&
\alpha_{1111,2};\;\alpha_{111,1};\;\alpha_{11,1};\;\alpha_{1,1}
\\
 \cdots
\\
 (A_{1211,1112})
&
\alpha_{1211,2};\;\alpha_{121,1};\;\alpha_{12,1};\;\alpha_{1,1}
\end{array}
$} \hspace{-5em}\right),
\end{align*}
and a global precondition 
\begin{math}
A=  A_{1,1} 
 \lor \cdots\lor A_{1211,1112}
\end{math}. By \cref{thm:correctness}, it is guaranteed that $\hquad{\Safe\land\Env}{A}{\alpha}{\Goal}$ is valid. The resulting $(A,\alpha)$ are rather long---logically combining dozens of symbolic inequalities. This makes them hard to read for humans, but computers have little problem checking and executing them (\cref{sec:exp}).

\end{example}

\begin{remark}
 In~\cref{eq:globalPrecondAndPropRes},  
we do not require $w$ to designate a leaf node---this is strange if we think of $\alpha$ to lead from the \emph{beginning} of the driving scenario in question to its goal. We include non-leaf nodes because of the use of the resulting RSS rule in the simplex architecture (cf.~\cref{subsec:introRSSSafetyArchitecture,subsec:impl}). In the simplex architecture, a global proper response (as BC's control) can be switched on and off, depending on whether the RSS conditions are true or not at each moment. It is then beneficial if we can start a global proper response from somewhere in its middle; this is what the program
$\alpha_{j_{1}\dotsc j_{k},i_{k}};\,
   \cdots\, ;\,
   \alpha_{j_{1},i_{1}}$ stands for, when  $j_{1}\dotsc j_{k}$ designates a non-leaf node. 
It should be noted too that the formal notion of scenario (\cref{def:drivingScenario}) does not state where to start---it only specifies where to reach, and what conditions to maintain.
\end{remark}

\subsection{Another Example Scenario: Emergency Stop with Limited Visibility}\label{subsec:limitedVisibility}

\begin{figure}[tbp]\centering
  \centering
 \includegraphics[bb=0 0 440 505,clip,scale=.45]{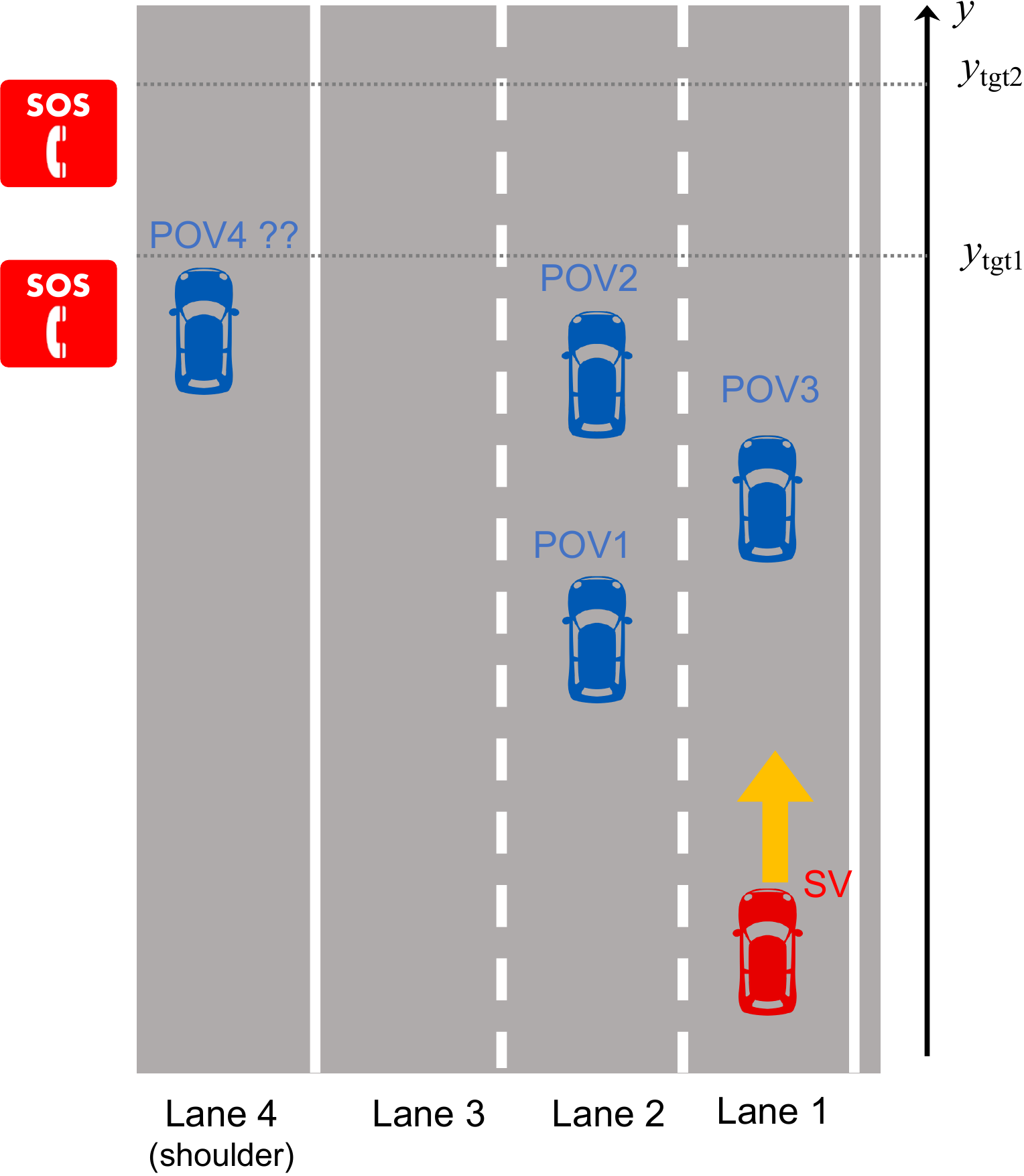}
 \caption{The scenario in \cref{subsec:limitedVisibility}: emergency stop with limited visibility}
 \label{fig:limitedVisibility}
\end{figure}

  We have given a detailed account of how to apply the workflow to the
  pull over scenario (\cref{ex:pullover}).
  Here, we present how the workflow applies to another scenario, in order to
  validate the applicability of the workflow.

The scenario is illustrated in \cref{fig:limitedVisibility}. 
  In this scenario there are 4 lanes: 3 driving lanes (Lane 1-3), and
  the hard shoulder (Lane 4). \SV{} starts in Lane 1, there is a single \POV{} in
  Lane 1 (ahead of \SV{}) and two \POV{}s in Lane 2, as in \cref{ex:pullover}.
  \SV{} is required to stop on
  the hard shoulder at one of two possible locations ($\ytgtn{1}$ or
  $\ytgtn{2}$).
  Stopping at the first target location ($\ytgtn{1}$) is preferred,
  but there may be a \POV{} (\POV{4}) parked there, in which case
  \SV{} should stop at the second location ($\ytgtn{2}$).
  Furthermore, \SV{} only becomes aware of \POV{4} when
  it comes within sensing distance $D$ of it, which we assume to be at
  least a fixed constant, say $\SI{50}{\metre}$.

The scenario differs from the previous one (\cref{ex:pullover}) in, among others, 1) the number of lanes and 2) the dynamic character (the first location $\ytgtn{1}$ may be occupied). As we sketch below, our logical workflow is  applicable to this scenario.

  To model the existence of \POV{4}, we use a
  variable $p$ whose value is $1$ when \POV{4} exists, and $0$
  otherwise.
  In the case that $p = 1$, variable $y_{4}$ gives the position of
  \POV{4}.
  Another variable $d$ takes on the values $0$ or $1$ depending
  on whether \SV{} has \emph{detected} \POV{4}.
  The goal, safety and environment assertions are defined as follows:
  \begin{align*}
    \Goal 
    &= 
      \left(
      \begin{aligned}
           & l = 4 \land v = 0 \land (y = \ytgtn{1} \lor y = \ytgtn{2})\\
           &\land (p = 0 \Rightarrow y = \ytgtn{1})
      \end{aligned}
      \right) \rlap{,}\\
    \Safe 
    &= 
      \left(
      \begin{aligned}
           & \textstyle\bigwedge_{i=1,2,3} (\aheadSL_{i} \Rightarrow y_{i} - y > \dRSS(v_{i}, v))\\
           & \land (p = 1 \land \aheadSL_4) \Rightarrow y_{4} - y > \dRSS(v_{4}, v) \\
           &\land (0 \leq v \leq \vmax \land -\bmin \leq a \leq \amax)
      \end{aligned}
      \right) \rlap{,}\\
    \Env 
    &= 
      \left(
      \begin{aligned}
        & \textstyle\bigwedge_{i=1,2,3} \bigl(\,\vmin\le v_{i}\le \vmax \land a_{i} = 0 \land l_{i} = i \,\bigr)\\
        & \land y_{2} > y_{1}\\
        & \land v_{4} = 0 \land y_{4} = \ytgtn{1} \land D > 50
      \end{aligned}
      \right) \rlap{.}
  \end{align*}

  Decomposition into subscenarios proceeds as in
  \cref{subsec:subscenarioId} for \cref{ex:pullover}.
  In total there are six subscenarios:
  1)~prepare to merge into Lane~2 (by adjusting speed and position);
  2)~merge into Lane~2;
  3)~merge into Lane~3;
  4)~prepare to merge into Lane~4, while waiting for $\ytgtn{1} - y < D$;
  5)~merge into Lane~4;
  6)~stop at target location.
  The subscenario subgoals are similar to those in
  \cref{subsec:subscenarioId}, except for $\Goal^{(4)} = (l = 3 \land
  \ytgtn{1} - y = D)$.
  The subscenario tree branches at the level of subscenario 6
  depending on which target location \SV{} aims to stop at, and at the
  level of subscenario 1 depending on whether \SV{} merges in front or
  behind \POV{1}.

  The subscenario proper responses were derived in much the same way
  as in \cref{subsec:properRespIdSubscenario} for \cref{ex:pullover}.
  Of note were those derived for subscenarios 4 and 5 which took the
  form:
  \begin{align*}
    \alpha_{4} \ &= \ \left( \dwhileClauseNb{\ytgtn{1} - y > D}{\boxed{ \ \beta_4 \ }} \, \right) \rlap{,} \\
    \alpha_{5} \ &= \ \left( \assignClause{d}{p}; \ifThenElse{d=1}{\boxed{ \ \beta_{5,1} \ }}{\boxed{ \ \beta_{5,0} \ }} \, \right) \rlap{.}
  \end{align*}
  \SV{} waits to find out if there is a parked vehicle or not
  ($\beta_4$ is a hybrid program that stays in Lane~3),
  and then responds in one of two ways: if a \POV{} is detected, then
  \SV{} merges before $\ytgtn{2}$ ($\beta_{5,1}$), otherwise it merges
  before $\ytgtn{1}$ ($\beta_{5,0}$). 
  From this, we derive preconditions and proper responses as in the
  pull over example.
  The preconditions and proper responses were successfully derived by
  following the workflow.

\subsection{Discussions}\label{subsec:workflowDiscussions}
We conclude  with some discussions of our workflow.

\subsubsection{Compositionality}
We argue that our workflow (Procedure~\ref{alg:workflow}) is compositional. 

Firstly, the design of proper responses is split up from the whole scenario to individual subscenarios, and it can be done independently for each subscenario (\cref{line:subscenarioPropRespId}). We showed through our leading example that it can be done systematically, combining possible longitudinal and lateral movements, now that  a goal  and a safety condition  are much simplified. It is also worth noting that many subscenarios are similar to each other, allowing one to reuse  previous analysis. 

In this paper, for simplicity, we focused on a limited number of subscenario proper responses that we see as more important than others (see \cref{ex:properRespIdPullOver}). A viable alternative is to systematically list possible proper responses, even if some of them have limited applicability (i.e.\ strong preconditions). A well-developed software support and/or ample human resources would allow this brute-force approach. 
See also \cref{sec:toolSupport}. 

Secondly, identification of subscenario preconditions (\cref{line:backProp}) is compositional, too---in the same sense as program verification in Floyd--Hoare logic is compositional. Unlike \cref{line:subscenarioPropRespId}, identification of $A_{w,u}$ is not independent for different $w,u$---there is a backward interdependence in the form of~\cref{eq:backPropAssignmentCond}. However, splitting up the task of precondition identification  to  simpler subscenarios certainly makes it easier, as we see in \cref{ex:backPropPullOver}. There are a lot of duplicates, too, so that one can reuse previous reasoning.

\subsubsection{Completeness} 
There are many ``best-effort'' elements in our workflow: 
\begin{itemize}
 \item 
 On~\cref{line:subscenarioPropRespId}, the list of subscenario proper responses should better be more comprehensive, but there is no formal criterion on what is enough or what is the best. 
 \item On~\cref{line:backProp},  subscenario preconditions are only subject to~\cref{eq:backPropAssignmentCond} that can be satisfied even by $\false$. It is only desired that they are weak.
 \item Moreover,  on~\cref{line:goalDecomposition,line:subscenarioRefine}, there is no formal criterion what is a good subscenario decomposition. The conditions in \cref{def:subscenario,def:subscenarioTree} are only minimal sanity checks. 
\end{itemize}
Consequently, the question ``how useful is the obtained RSS rule $(A,\alpha)$?'', that is, ``is $A$ weak enough?'', will always stand. 

We argue, however, that this completeness issue should not  block the use of  our workflow.
\begin{itemize}
 \item Firstly, it is not hard to come up with proper responses whose preconditions are fairly weak. This can be done by mimicking what human drivers would do, in which case the RSS-supervised ADS is at least as goal-achieving as human drivers. 
 \item Secondly, we can always incrementally improve $(A,\alpha)$ by identifying more $\alpha_{w,i}$ and weaker $A_{w,u}$. Note that this process \emph{monotonically weakens} the precondition $A$ since it adds new disjuncts to $A$ (see~\cref{eq:globalPrecondAndPropRes}). The process makes an RSS rule increasingly complete, without fallbacks. 
 \item Thirdly,  that $(A,\alpha)$ comes with a correctness guarantee (\cref{thm:correctness}) means that they can be used for many years to come, as a solid basis of safe driving. 
   The efforts for better $(A,\alpha)$ therefore pay off in the long run.
\end{itemize}

\subsubsection{On Environmental Assumptions}\label{subsubsec:discussionOnEnv}
In our leading example (the pull over scenario), we assumed constant speeds of the other vehicles ($\Env$ in \cref{ex:scenarioModelingPullover}). We note that 
\begin{itemize}
 \item while violation of this assumption may threaten goal achievement (namely stopping at $\ytgt$ in Lane~3), 
 \item it does \emph{not} threaten collision avoidance,
\end{itemize}
because the scenario's safety condition requires the RSS safety distance ($\dRSS$ from \cref{ex:onewayTraffic}) from every other vehicle (see~\cref{eq:pullOverSafetyCond}). 

The above point has the following practical implication, in the expected use of (CA- and GA-)RSS rules in the simplex architecture (\cref{subsec:introRSSSafetyArchitecture}; see also \cref{sec:exp}). 
In actual ADS, we expect another layer of the simplex architecture on top of the one based on our goal-aware RSS rules. The ``collision avoiding'' BC of this other simplex architecture monitors the RSS safety distance and brakes if necessary, thus ensuring collision avoidance at the possible sacrifice of goal achievement. 

\begin{auxproof}
 \begin{definition}[subgoal decomposition, subscenario]\label{def:subscenario}
 Let $\mathcal{S}=(\Var, \Safe, \Env, \Goal)$ be a scenario, and  $N\in \mathbb{Z}_{>0}$. We say that a list  $(\Goal_{1},\dotsc,\Goal_{N})$ of $\dHL$ assertions is a \emph{subgoal decomposition} of $\mathcal{S}$ if
 \begin{itemize}
 \item there exist a $\dHL$ assertion $A$ (called a \emph{global precondition}) and $\dHL$ programs $\alpha_{1},\dotsc, \alpha_{N}$ (called \emph{proper responses}), 
 \item such that the following Hoare quadruples and logical implication are all valid.
       \begin{align*}
	\begin{aligned}
 	&\hquad{\Safe\land\Env}{A}{\alpha_{1}}{\Goal{1}}        
	\\
	&\hquad{\Safe\land\Env}{\Goal_{i-1}}{\alpha_{i}}{\Goal_{i}} \quad(i=2,3,\dotsc, N)
	\\
	&\Goal_{N}\Longrightarrow \Goal
	\end{aligned}       
       \end{align*}
 \end{itemize}
 The relationship among $A, G_{i}, G, \alpha_{i}$ is informally  as follows. 
 \begin{displaymath}
 A \xrightarrow{\alpha_{1}} G_{1} \xrightarrow{\alpha_{2}} \cdots
 \xrightarrow{\alpha_{N}} G_{N} \stackrel{\text{implies}}{\Longrightarrow} G
 \end{displaymath}
 The \emph{subscenarios} of $\mathcal{S}$ induced by the subgoal decomposition $(\Goal_{1},\dotsc,\Goal_{N})$ are the scenarios (in the sense of \cref{def:drivingScenario})
 \begin{displaymath}
 (\Var, \Safe, \Env, \Goal_{1}), \dotsc,
 (\Var, \Safe, \Env, \Goal_{N}).
 \end{displaymath}
 \end{definition}
 Note that subscenarios differ from the original scenario $\mathcal{S}$ only in their goals $\Goal_{i}$.

 In the actual execution of \cref{line:goalDecomposition} of
 Procedure~\ref{alg:workflow}, it is usually not guaranteed if the identified subgoals $\Goal_{1},\dotsc, \Goal_{N}$ in fact give a subgoal decomposition in the sense of \cref{def:subscenario}. One needs to specify For such a guarantee, 
\end{auxproof}

\section{Software Support for Rule Derivation}\label{sec:toolSupport}
We discuss software support for our workflow (Procedure~\ref{alg:workflow}). Note that, in this section, we focus on software for \emph{deriving} goal-aware RSS rules. In contrast, software for \emph{using} goal-aware RSS rules in the simplex architecture is heavily dependent on the choice of AC (\cref{subsec:introRSSSafetyArchitecture})---it is discussed separately in \cref{subsec:impl}. 

 Our workflow (Procedure~\ref{alg:workflow}) involves two types of tasks:
\begin{itemize}
 \item \emph{human discovery} tasks, namely  
 of subscenarios (\cref{line:subscenarioRefine}), proper responses (\cref{line:subscenarioPropRespId}), and preconditions (\cref{line:backProp}), and 
 \item 
 \emph{logical reasoning} tasks in $\dHL$, such as ensuring the condition~\cref{eq:backPropAssignmentCond} and computing $A, \alpha$ (see~\cref{eq:globalPrecondAndPropRes}).
\end{itemize}
Much like other formal verification problems, we can imagine different ways to execute them.
\begin{itemize}
 \item A \emph{pen-and-paper} execution. This requires less preparation/infrastructure work, but is more error-prone.
 \item A \emph{fully formalised} execution, much like in formal verification by theorem proving (see e.g.~\cite{NipkowPW02}). 
\end{itemize}

\begin{auxproof}
 The rest of the section is organised as follows. We discuss our current execution of the workflow---it lies somewhere between the above two extremes---in \cref{subsec:currentSoftwSupport}. We discuss the software support by Mathematica  used there, too.  Prospects of fully formalised logical reasoning are discussed in~\cref{subsec:towardsFullFormalization}. Automation of the human discovery tasks is a direction of software support that goes orthogonal to formalising logical reasoning. Our initial results in this direction are in \cref{subsec:partialAutomation}.
\end{auxproof}

\subsection{Current Software Support with Mathematica}\label{subsec:currentSoftwSupport}
Our current execution scheme of the workflow (Procedure~\ref{alg:workflow}) is only partially formalised. Its software support principally uses \emph{Mathematica notebooks}~\cite{Mathematica_12_3_1}, an interactive environment in which users can mix 
\begin{itemize}
 \item symbol manipulations such as solving quadratic equations, substitution, and tracking case distinctions, exploiting advanced algorithms of Mathematica as a  computer algebra system, and
 \item rich annotations for human readers, such as structured natural language descriptions, figures, and tables.
\end{itemize}
Our logical reasoning in the workflow is currently formalised as much as Mathematica can accommodate. Specifically,
\begin{itemize}
 \item (\emph{static reasoning is formalised}) all $\dHL$ assertions are expressed formally in Mathematica (which is possible since each $\dHL$ assertion is a predicate logic formula over reals), and their implications are formally checked using Mathematica functions such as $\mathtt{Simplify}$; but
 \item (\emph{dynamic reasoning is not formalised}) neither $\dHL$ quadruples nor their derivation using the rules in \cref{fig:dFHL-rules} is formalised, since the language of Mathematica does not accommodate them. 
\end{itemize}
Our current execution scheme therefore has room for improvement---formalisation of dynamic reasoning is certainly desirable. See \cref{subsec:towardsFullFormalization} for its prospects.

Nevertheless,  our current Mathematica-based and partially formalised execution scheme has the following distinctive advantages.
\begin{itemize}
 \item (Well-documented informal reasoning) Dynamic reasoning  for deriving $\dHL$ quadruples
   is recorded in Mathematica notebooks in an informal yet trackable manner, with  natural language explanations that explicate the $\dHL$ rules used therein. Therefore these proofs can be efficiently checked by human reviewers, if not machine checkable. 
 \item (Interaction for discovery) The interactive nature of Mathematica notebooks allows us to make trials and errors quickly for the human discovery part of our workflow (Procedure~\ref{alg:workflow}).
 \item (No static reasoning errors) A large part of mistakes in executing our workflow is in the treatment of vehicle dynamics expressed in the double integrator model. Formalisation of static reasoning in Mathematica purges these mistakes. Note that our vehicle dynamics  (\cref{rem:basicManeuvers})  have closed-form solutions, which easily reduce dynamic reasoning to static one.
\end{itemize}
Given these advantages, 
we claim that our current execution scheme gives high confidence in the correctness of derived goal-aware RSS rules, and that the execution scheme is a viable option for practical use.
\begin{auxproof}
 Recall that the correctness here means the validity of $\hquad{\Safe\land\Env}{A}{\alpha}{\Goal}$ (\cref{def:GARSSRule}), which essentially follows from the correctness~\cref{eq:backPropAssignmentCond} for each subscenario (as we showed in   the proof of~\cref{thm:correctness}).
\end{auxproof}

\subsection{Estimated Workload for Rule Derivation}
We estimate the workload as follows: an expert in our workflow and its software support would need several days to derive a GA-RSS rule,  for a scenario of the complexity of \cref{ex:pullover}. This was our experience when one of the current authors conducted the task. Getting acquainted with our workflow and its software support is not hard, either, especially for people with backgrounds in formal logic. We expect that the required training would take a couple of weeks. 

Moreover, the compositional workflow and its  implementation in Mathematica (featuring informal yet well-documented reasoning) allow efficient collaboration of multiple people. For example, we could parallelise the identification of subscenario proper responses $\alpha_{1111,1}$ and $\alpha_{1111,2}$---together with the identification of the preconditions $A_{1111,1111}$ and $A_{1111,1112}$, see \cref{fig:propagation}---and distribute the task to different people. The same happened between the subscenarios $\mathcal{T}_{1111}$ and  $\mathcal{T}_{1211}$ in \cref{fig:propagation}. The communication cost between different workers was kept minimal, since they could communicate semi-formally via Mathematica notebooks.

Overall, while the workload of deriving GA-RSS rules is not very light (it is hardly a matter of minutes, for example), we claim that it is light enough to be useful, especially given that the derived GA-RSS rules can be used as a rigorous basis of safe ADS in many years to come.

\subsection{Towards Full Formalisation}\label{subsec:towardsFullFormalization}
For fully formalised execution of our workflow (Procedure~\ref{alg:workflow}), natural tools to use are theorem provers for differential dynamics such as \KeYmaeraX~\cite{MitschP17}. Our logic $\dHL$ is designed with its translation to differential dynamic logic $\dL$~\cite{Platzer18} in mind, so the use of \KeYmaeraX\ (which is a theorem prover for $\dL$) should not be hard. 

We are currently working on systematic translation of $\dHL$ to $\dL$, and on some dedicated proof tactics in \KeYmaeraX. Our preliminary experience of manually formalising part of the reasoning for the pull over scenario (\cref{ex:backPropPullOver}) is encouraging. Building a fully formalised  infrastructure for the workflow will take considerable time and effort, though.

\begin{auxproof}

\subsection{Automating Human Discovery Tasks}\label{subsec:partialAutomation}
\todoil{What do we do with this subsection? See the discussion notes}
We have identified some automation opportunities in a human discovery part of our workflow---for proper responses (\cref{line:subscenarioPropRespId}) and preconditions (\cref{line:backProp}), specifically---in case the required conditions are simple and \emph{monotone}. The last monotonicity means that there are no conditions in $\Safe_{w}, \Env_{w}, \Goal_{w}$ that are at odds.  Monotonicity holds for $\mathcal{T}_{1}, \mathcal{T}_{12}, \mathcal{T}_{121}, \mathcal{T}_{1211}$ in \cref{fig:subscenarioDefPullOver}---driving more slowly makes all the conditions more true. Monotonicity does \emph{not} hold for $\mathcal{T}_{1111}$ in \cref{fig:subscenarioDefPullOver}---the two RSS safety distance requirements in $\Goal_{1111}$ are at odds with each other. 

As an example, we sketch the automation that we have implemented for  $\mathcal{T}_{1}, \mathcal{T}_{12}, \mathcal{T}_{121}, \mathcal{T}_{1211}$ in \cref{fig:subscenarioDefPullOver}. A general framework for such automation is future work.

\begin{example}

(Can be extended to any number of lanes, and any number of \POV{}s driving there in front of \SV{}) 
\end{example}

In the case where \SV{} merges behind \POV{1}, our methodology has been
fully automated, and applies more generally to any number of \POV{}s, in any number
of lanes, in any configuration.

The proper response is conceptually described as the concurrent operation of the
following \emph{longitudinal} and \emph{lateral} manoeuvre sequences:
\begin{itemize}
  \item Longitudinal:
  \begin{itemize}
    \item Brake at $\bmin$ until \SV{}'s speed reaches the
          minimum allowed speed for the driving lanes (lanes 1 and 2),
    \item Cruise at the minimum driving speed until reaching the \emph{stopping
          distance} to $\xtgt$.
    \item Brake at $\bmin$ so as to stop at $\xtgt$.
  \end{itemize}
  \item Lateral:
  \begin{itemize}
    \item Keep in lane 1 until the RSS safety distance is achieved
          behind \POV{1}.
    \item Change lanes from lane 1 to lane 2.
    \item Change lanes from lane 2 to lane 3.
    \item Keep in lane 3 until stopping at the target.
  \end{itemize}
\end{itemize}
The essence of the strategy is for \SV{} to drive to the target while
decreasing speed as quickly as possible while also merging behind \POV{1} as
early as possible. By doing so, the preconditions generated will be as weak as
possible\footnote{For we can show that this behaviour will always be
  positionally behind, and at lower speed, than any other behaviour at all
  times.}.

While the strategy seems to sacrifice progress at the cost of generality, this
trade-off becomes irrelevant once the resulting controller is used as \BC{}
in the simplex architecture: progress is handled by \AC{}, while \BC{} only
interrupts when it is strictly necessary to maintain RSS safety or to achieve the
goal, e.g.~breaking for a short amount of time until the precondition is
restored and \AC{} takes back control.

The latitudinal and longitudinal events above may interleave in a number of
different ways, for example \SV{} may reach the minimum speed before or
after reaching the RSS safety distance behind \POV{1}. In the end we obtain 4
rules for subscenario 1, with the following set of Hoare triples:
\[
  \arraycolsep=1pt\def\arraystretch{1.2}
  \begin{array}{ccccccccc}
    \{A_{1,1}\} & \bb     & \{A_{2,1}\} & \bb     & \{A_{3,1}\} & \bb;\cc & \{A_{4,1}\}\\
         &         &      &         &      &         & \{A_{4,1}\} & \cc;\bb & \{B\}\\
    \{A_{1,2}\} & \bb     & \{A_{2,2}\} & \bb;\cc & \{A_{3,2}\} & \cc     & \{A_{4,1}\}\\
    \{A_{1,3}\} & \bb;\cc & \{A_{2,3}\} & \cc     & \{A_{3,2}\}\\
    \{A_{1,4}\} & \bb     & \{A_{2,4}\} & \bb     & \{A_{3,3}\} & \bb     & \{A_{4,2}\} & \bb;\cc;\bb & \{B\}
  \end{array}
\]
Here we are using the abbreviation $\bb$ for a braking manoeuvre, and $\cc$ for a
cruising manoeuvre\footnote{Note that each instance of $\bb$ or $\cc$ above uses
  a different duration for the manoeuvre, so they are not the same program.}. For
example the rule
\[
  \{A_{1,3}\}~\bb;\cc~\{A_{2,3}\}~\cc~\{A_{3,2}\}~\cc~\{A_{4,1}\}~\cc;\bb~\{B\}
\]
describes the proper response which brakes until \SV{} reaches the minimum
speed, which occurs before reaching the RSS safety distance behind
\POV{1}, cruises until it does reach this distance, cruises throughout both
lanes changes, and stops at the target in lane 3.

The process works its way backwards, starting from the final goal. At each step,
a full lateral subscenario is processed. Equations are solved to obtain formulas
for the durations of the manoeuvres. With this, positions and velocities of all
vehicles at all times can be re-expressed in terms of the conditions at the
start of the subscenario. This can then be used to write down conditions for the
RSS safety distances being respected during the whole subscenario. These, along
with the (transformed) postcondition and conditions for non-negativity of the
durations of the manoeuvres then combine into a single precondition for the
subscenario.

Our implementation uses (embedded) logic programming to find the full set of
valid interleavings of events and all the RSS conditions that must be checked,
makes API calls to Wolfram Engine for solving solving equations and simplifying
boolean expressions, generates C++ code for the proper responses, and outputs
full proof-trees certifying all the Hoare triples which can be checked with an
automated theorem prover.

\todoil{James:

  Sketchy fully automated example starts here.
}

Suppose that \SV{} is driving on a highway with $m$ lanes, and that
\SV{} must stop in a specific lane, at some target location. We assume an
arbitrary amount of \POV{}s, and that they will keep a constant velocity. There are
many strategies that \SV{} may pursue in order to achieve its objective; we
concentrate here on the strategy which slows down as much as is permitted by the
scenario. That is, we assume that each lane has some minimal speed limit, and
the overall proper response \SV{} will pursue is to decelerate as much as
possible in each lane, while also changing lanes as early as it is able to to so
in order to reach its target lane. This strategy is particularly suited to the
case when the target is already quite close.

Without loss of generality, we may assume that the target lane is to the
right of the starting lane, so that we can number the intermediate lanes as:
\[
  l_{0}, \ l_{1}, \ \ldots, \ l_{n},
\]
with $l_{0}$ being \SV{}'s initial lane, and $l_{n}$ being the target lane.
\SV{} will thus perform $n$ lane changes in order to reach the target.

When on lane $l_{i}$, the proper response will be to decelerate at the
maximum allowed rate until one of the following two conditions is reached:
\begin{itemize}
  \item The minimum speed for the lane $l_{i}$ has been reached.
  \item \SV{} is able to merge into lane $l_{i+1}$, that is, there is a pair
        of consecutive \POV{}s $p_{1}$ and $p_{2}$ on lane $l_{i+1}$ such that
        \SV{} has achieved the RSS safety distance behind $p_{1}$ and in front
        of $p_{2}$, and may thus initiate a lane-change manoeuvre\footnote{We
        allows $p_{1}$ or $p_{2}$ to be $\mathtt{null}$; in other words we also
        consider the case of merging in front of the first \POV{} of the lane, or
        merging behind the last \POV{} of the lane.}.
\end{itemize}
In the first case \SV{} continues cruising at the minimum speed until the
second condition is met. If lane $l_{i+1}$ has $m$ \POV{}s, there is thus $2(m+1)$
cases to consider for the lane change from $l_{i}$ to lane $l_{i+1}$. We obtain
in this way a scenario tree with branching factor at most $2(m+1)$ for
performing the $n$ lane changes and stopping at the target, where $m$ is now the
maximum number of \POV{}s per lane. This scenario tree can be generated fully
automatically, and the proper-responses and RSS conditions derived
automatically without human intervention.



\end{auxproof}

\section{Experiments and Evaluation}\label{sec:exp}

In \cref{sec:workflow} we presented a general workflow to
generate GA-RSS rules. The ``implementation'' of the workflow, specifically software support for its execution, was discussed in \cref{sec:toolSupport}.

In this section, we seek to quantitatively evaluate our workflow by conducting experiments on a
specific \emph{output} of the workflow.
Concretely, we evaluate the  GA-RSS rule set for the pull over scenario (\cref{ex:pullover})---obtained
using the workflow in \cref{sec:workflow} and finalised in
\cref{ex:globalPullOver}---using it as the baseline controller (\BC{}) in the
simplex architecture (\cref{subsec:introRSSSafetyArchitecture}).

The resulting RSS-supervised controller is denoted by  $\ACGA$; we will  compare its performance 
to other similar controllers (namely $\ACCA$ and $\AC$, introduced later).

The rest of the section is organised as follows. We pose several research questions in \cref{subsec:exp:rqs}, based on which we designed our experiments. The implementation of $\ACGA$ is introduced in \cref{subsec:impl}, together with those of $\ACCA$ and $\AC$. Experiment settings and results are described in \cref{subsec:exp:setting}. Based on these results, in \cref{subsec:exp:discussion},  we address the research questions that we posed earlier. Finally, in \cref{subsec:exp:notable}, we take a closer look at two notable scenario instances that further demonstrate the value of GA-RSS.

\subsection{Research Questions}
\label{subsec:exp:rqs}

To fully evaluate the GA-RSS rule set for the pull over scenario (\cref{ex:globalPullOver}), we need to
answer several research questions.
Our claim is that the rule set we derived with our workflow is able
to achieve a given goal (namely, pulling over) safely, which leads to
our first research questions.
\begin{researchquestion}
  \label{rq:safety}
  \safetystring
\end{researchquestion}
\begin{researchquestion}
  \label{rq:goal}
  \goalstring
\end{researchquestion}
\noindent
Another crucial point of the method, if we want it to be used in
automated driving, is whether this controller can be used
in practice.
For example, are the RSS conditions not too complex to be computed
repeatedly in a control loop at run-time? Are they not too restrictive to apply to many common driving situations?
\begin{researchquestion}
  \label{rq:practicality}
  \practicalitystringlong
\end{researchquestion}
\noindent
While reaching the goal and maintaining safety are the two principal requirements of
our controller, it is also desirable to test it for other metrics.
Reaching the goal in good time and comfortably are desirable, even if
these concerns are secondary to safety and goal achievement.
\begin{researchquestion}
  \label{rq:metrics}
  \metricsstring
\end{researchquestion}
\noindent
Finally, the controller we build is based on the simplex architecture,
and contains an advanced controller (\AC{}), which may be unsafe, but
is usually optimised for speed and comfort.
Our simplex architecture should thus interrupt \AC{} as rarely as
possible.
\begin{researchquestion}
  \label{rq:intrusion}
  \intrusionstring
\end{researchquestion}

%
%
%
%
%
%
%

\subsection{Implementation of the Controllers}\label{subsec:impl}
Our controller $\ACGA$  that uses the GA-RSS rule set (\cref{ex:globalPullOver}) is based on the simplex architecture. As AC of the architecture (\cref{subsec:introRSSSafetyArchitecture}),
we used a prototype
planner\footnote{This is a research prototype that is provided by Mazda Motor Corporation. It is however  unrelated to any of its products. } based on the algorithm in~\cite{mcnaughton2011motion}. 
\AC{} is a sampling-based controller that, at each time step, generates a large
number of candidate short-term paths and chooses the best in terms of a cost
function. The cost function is a weighted sum of costs for multiple concerns;
they are namely safety, progress, vehicle dynamics feasibility, traffic law
compliance, and comfort.


In our GA-RSS-supervised controller $\ACGA$,
the current implementation of BC is a ``surrogate'' one: instead of directly implementing the proper responses we identified in~\cref{sec:workflow}, we implemented them as an alternative cost
function for the sampling-based controller used as AC. Specifically, this cost function favours the short-term path that
is the closest to the desired proper response. The decision module (\DM{})
checks all RSS
conditions $A_{w,u}$ (computed on \cref{line:backProp} of
Procedure~\ref{alg:workflow}), and allows  \AC{} to remain in control as long
as one of them remains valid. When the last one fails,  \DM{} switches control
to BC which engages in the corresponding proper response $\alpha_{w,i}$
(identified on \cref{line:subscenarioPropRespId} of
Procedure~\ref{alg:workflow}).


\begin{remark}[prioritisation of GA-RSS rules]\label{rem:prioritization}
 The above description of \DM{} and \BC{} is simplified: for enhanced progress, we additionally employ the  \emph{prioritisation} mechanism, explained below. 

Specifically,
 some of our proper responses are designed with progress in mind
 (while still ensuring safety), while others reach the goal but do not make
 significant progress. In order to make our controller efficient, we separate
 rules into a high-priority and a low-priority group, and only use rules from the
 high-priority group whenever possible. If at any point during the execution,
 some high-priority rule can be engaged, then the low-priority set is discarded
 to only allow the use of high-priority rules.
\end{remark}


We compare our controller \ACGA{} to two other controllers:

\underline{\bfseries \AC{}}: 
AC alone---the sampling-based controller discussed
  above---without any BC.

\underline{\bfseries \ACCA{}}: A collision-avoiding RSS-supervised controller. In this instance of the
    simplex architecture, BC implements the following CA-RSS rules. They are a straightforward adaptation of the classic CA-RSS rule in~\cref{ex:onewayTraffic}. 
        \begin{itemize}
          \item  \SV{} must maintain the RSS safety distance from any
                \POV{}  1) that \SV{} shares a lane with and 2) that is in front of \SV{}. The
                proper response is to brake until the RSS safety
                distance is restored.
          \item Additionally, when \SV{} changes lanes, it must allow RSS safety distances both in front of it and behind it.\footnote{This is how we formalise the RSS responsibility principle 2) ``Don't cut in recklessly''---see \cref{ex:goalDecompPullover}.}
The proper response is to abort
                changing lanes if these RSS safety distances are not secured.
    \end{itemize}



\subsection{Experiment Settings and Results}
\label{subsec:exp:setting}


We ran simulations with the three controllers above under different instances of the
pull over scenario (\cref{ex:pullover}). The scenario instances were generated by the following parameter
values---we found them generate relevant scenario instances. 
 Here the positions are in meters (\si{\metre}) and
velocities are in \si{\metre\per\second}:

\begin{displaymath}\small
\begin{array}{l}
  v(0) \in \{10, 14\}, \quad y(0)=0,\quad v_{i}(0) \in \{10, 14\} \quad(i=1,2,3);\\
  y_{1}(0) \in \{-10, -5, 0, 5, 10\};\quad
  y_{2}(0) \in \{75, 80, 85, 90, 95\};\\
  y_{3}(0) \in \{85, 90, 95, 100, 105\};\quad
  \xtgt \in \{140, 160, 180\},
\end{array}
\end{displaymath}
We imposed the constraint $v_{1}(0)\le v_{2}(0)$ to avoid collisions between
\POV{}s.
For constants, we used the following values taken from our AC: $\vmin = \SI{10}{\metre\per\second}$, $\vmax = \SI{28}{\metre\per\second}$, $\rho = \SI{0.3}{\second}$,
$\amax = \SI{0.98}{\metre\per\second^2}$, $\bmax = \SI{8}{\metre\per\second^2}$, and
$\bmin = \SI{2.94}{\metre\per\second^2}$.

We nevertheless found that some scenario instances are clearly irrelevant  (e.g.\ $\ytgt$ is too close to stop at); we ruled out those instances as follows.
While simulating, if none of the RSS conditions for our GA-RSS
rule set are satisfied for the first \SI{0.5}{\second}, then we discarded the scenario instance.
In total, we kept $2350$ instances, that is $52\%$ of the total number
of instances.


\begin{table*}
  \caption{Experimental results.
  In \emph{goal}, we count the number of instances that reach the
  goal.
  In \emph{RSS violation}, we count the number of instances where some
  RSS safety distance is violated, the average and maximal violation
  times, and by how much the RSS safety distance was violated.
  In \emph{\BC{} time}, we quantify how long \BC{} has been in
  control on average.}
  \label{tab:results}
  \centering
  \input{subs/stats}
\end{table*}
The statistics of  the simulation results are given in \cref{tab:results}.
In the \emph{goal} column, we count the number of instances that reach
the goal.
In the \emph{collision} column, we count how many instances resulted
in a collision.
In the \emph{RSS violation}, we count the number of instances where
some RSS safety distance is violated; the average and maximal
violation times.
We also compute by how much the RSS safety distance was violated: it
is computed as the maximal value along any execution, for any time
$t$, and relevant \POV{} $i$, of $1- (y_i(t) - y(t)) / \dRSS(v_i(t),
v(t))$.
In the \emph{time} column, we record the average and maximal travel
times.
In the \emph{jerk} column, we record the average and maximal amount of
uncomfortable jerk accumulated along the trajectory.
We only accumulate jerk over \SI{0.5}{\metre\per\second^{3}}, as jerk below
that threshold is considered
comfortable~\cite{Czarnecki18ADSReqAnalysisDBC}.
In the \emph{\BC{} time} column, we quantify how long \BC{} has
been in control on average.

In \cref{fig:progress} and \cref{fig:comfort}, we give more details on
the distributions of travel times and accumulated jerk.
In \cref{fig:progress}, we show \ACGA{}'s total travel time
compared to both \ACCA{} and \AC{}.
The size of a disc is proportional to the number of scenario instances
with that travel time.
A red disc indicates that \ACCA{} failed to achieve the goal in this
case.

\begin{figure*}[tbp]
  \centering
  \begin{subfigure}{0.48\textwidth}
    \centering
    \includegraphics[width=\textwidth]{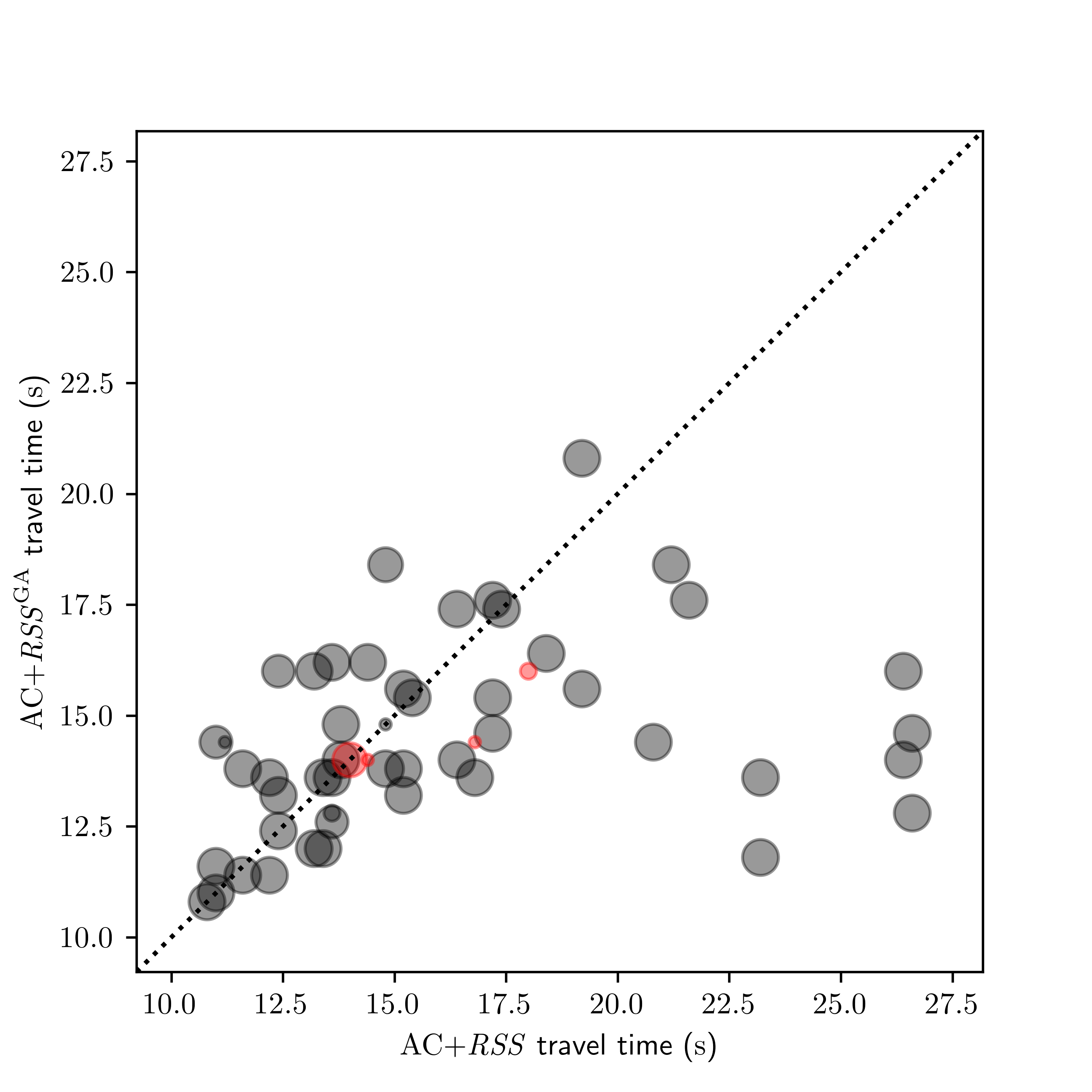}
    \caption{\ACCA{} vs \ACGA{}.}
    \label{fig:AC+CA-AC+GA-total_time}
  \end{subfigure}
  \hfill
  \begin{subfigure}{0.48\textwidth}
    \centering
    \includegraphics[width=\textwidth]{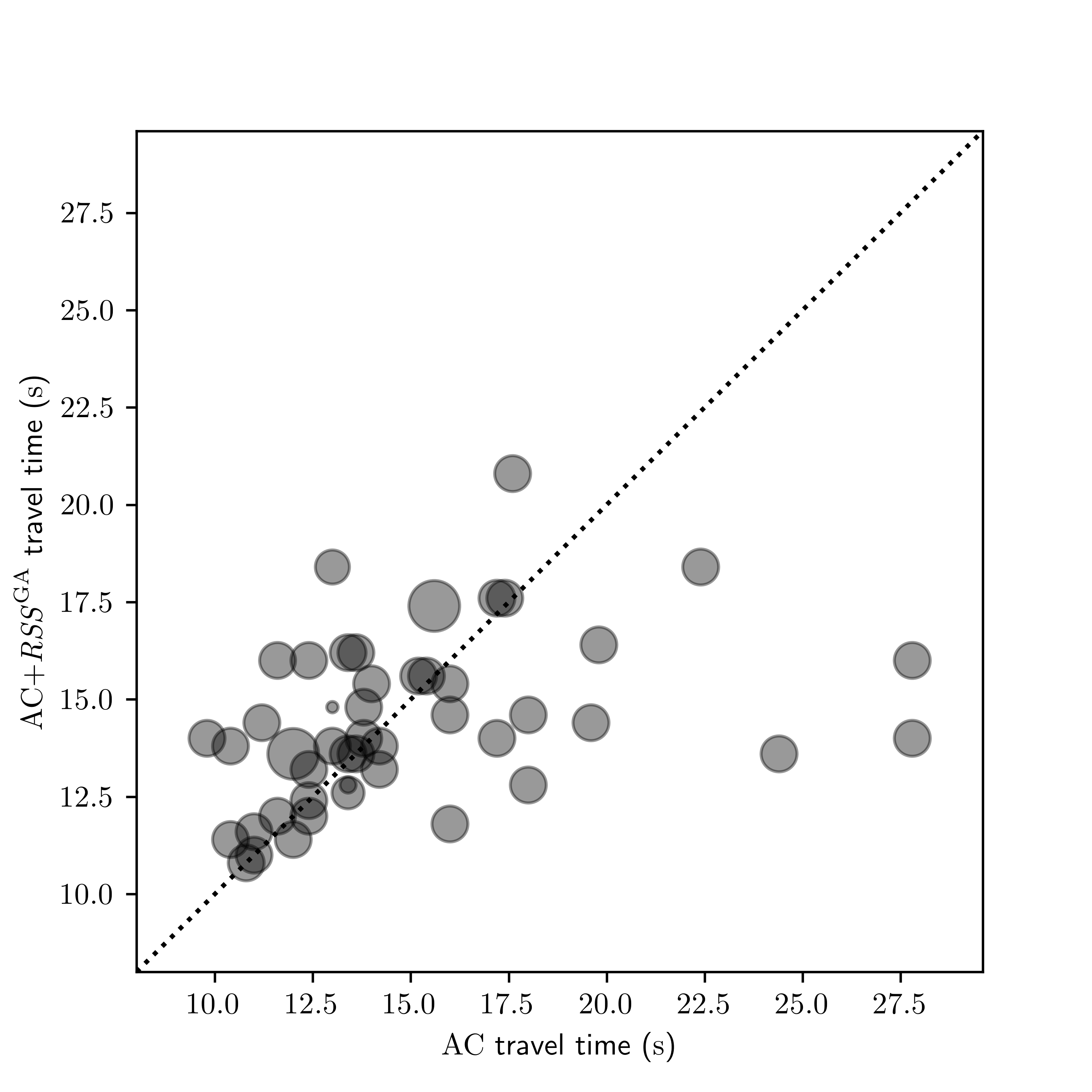}
    \caption{\AC{} vs \ACGA{}.}
    \label{fig:AC-AC+GA-total_time}
  \end{subfigure}
  \caption{Comparison for progress. A red disc indicates that \ACCA{} failed to achieve the goal in this scenario instance.}
  \label{fig:progress}
\end{figure*}

\begin{figure*}[tbp]
  \centering
  \begin{subfigure}{0.48\textwidth}
    \centering
    \includegraphics[width=\textwidth]{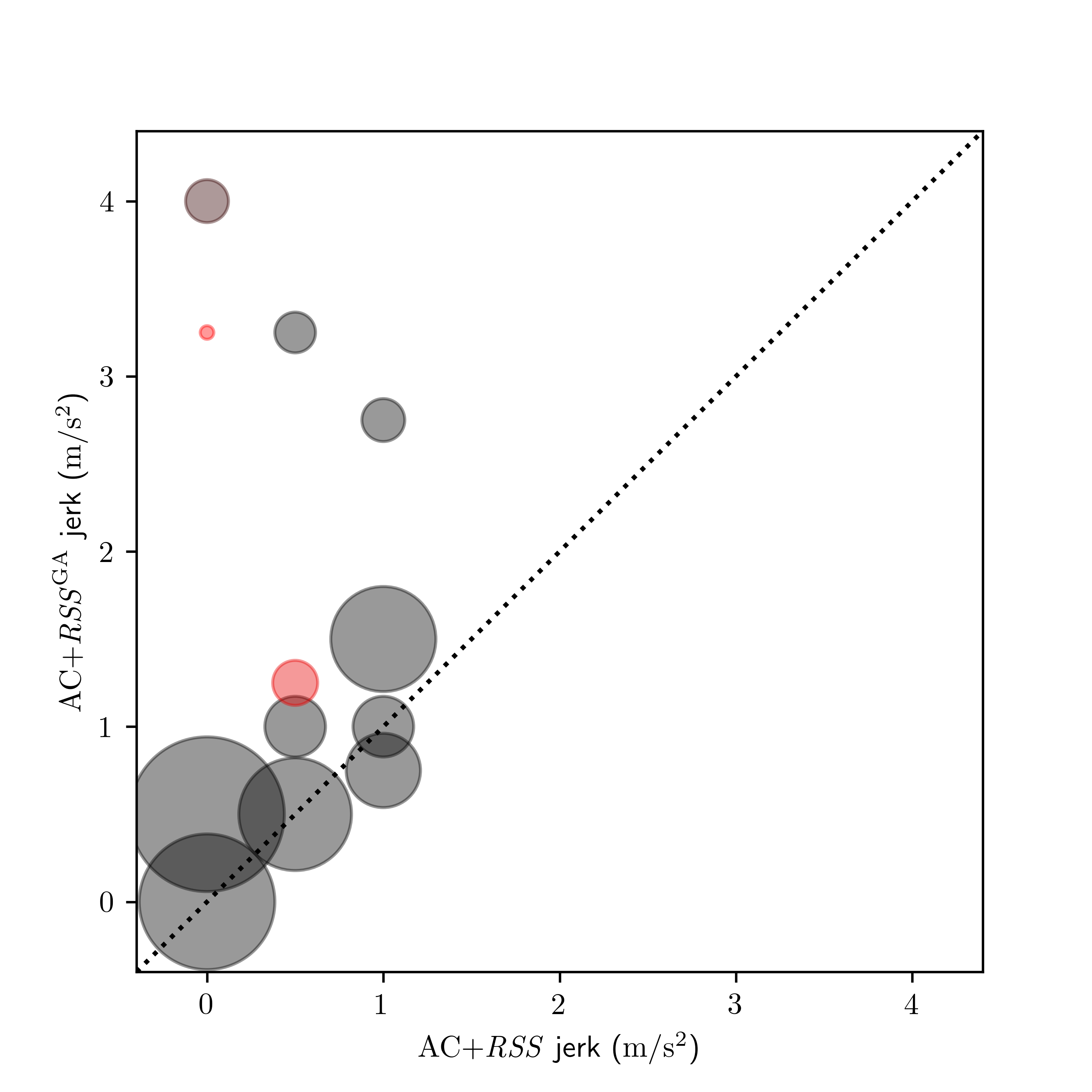}
    \caption{\ACCA{} vs \ACGA{}.}
    \label{fig:AC+CA-AC+GA-jerk}
  \end{subfigure}
  \hfill
  \begin{subfigure}{0.48\textwidth}
    \centering
    \includegraphics[width=\textwidth]{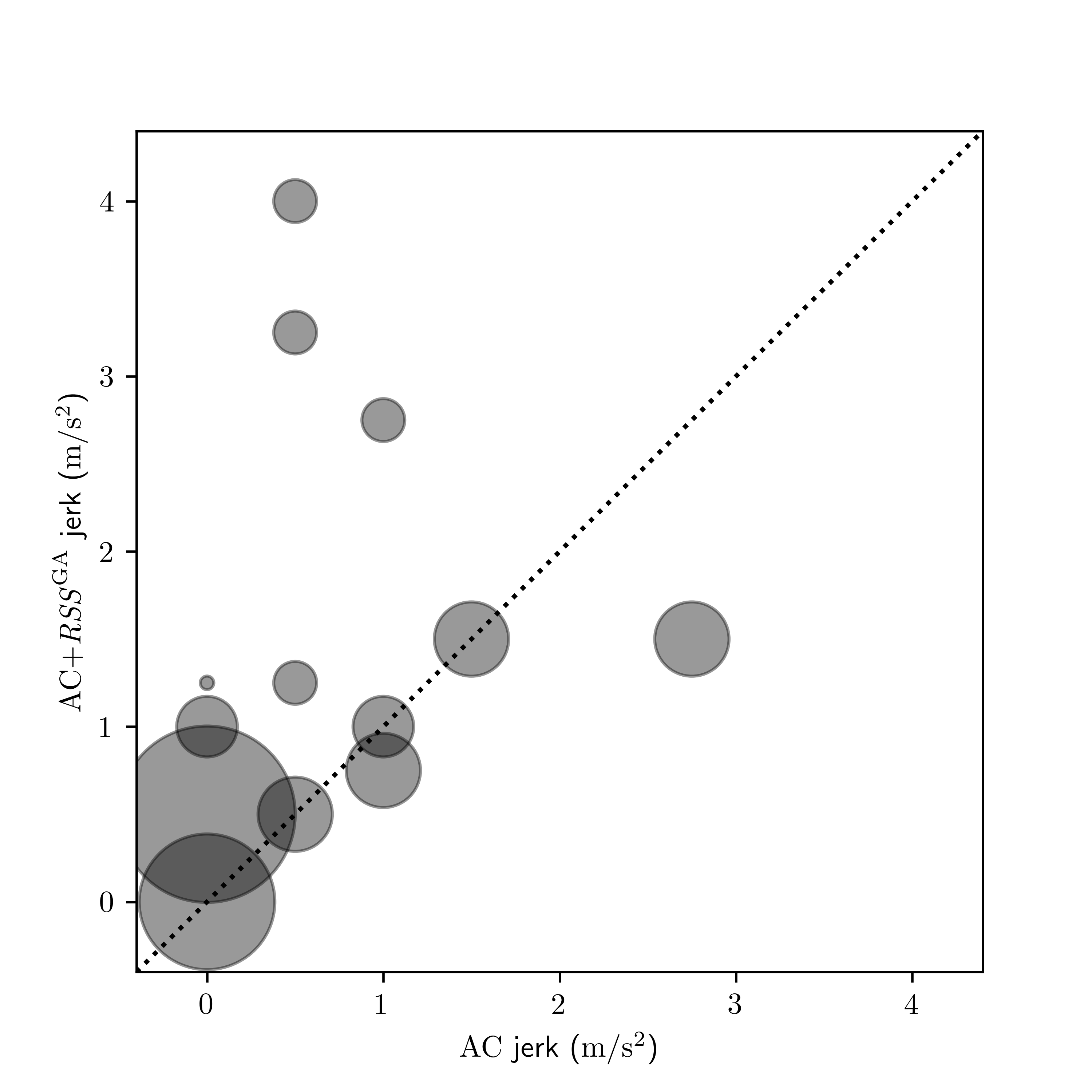}
    \caption{\AC{} vs \ACGA{}.}
    \label{fig:AC-AC+GA-jerk}
  \end{subfigure}
  \caption{Comparison for comfort. A red disc indicates that \ACCA{} failed to achieve the goal in this scenario instance.}
  \label{fig:comfort}
\end{figure*}


All these experiment results indicate comparative advantages and values of GA-RSS. We discuss them in detail below, along the research questions we posed in \cref{subsec:exp:rqs}. 

\subsection{Discussion}
\label{subsec:exp:discussion}

Let us address the different research questions in light of the
experimental results.

\subsubsection{\cref{rq:safety}: \safetystring}

%
%
%

All three controllers successfully avoided collisions. However, when it comes to maintaining RSS safety distances in order to prepare for sudden changes of behaviours of other cars, their performance varied a lot. 

The worst performer in terms of RSS violation was \AC{}, with a number of violations (12.8\% of the scenarios) which tend to be longer (max.\ \SI{0.8}{\second}) and bigger (max.\ 82.16\%). The last number is particularly alarming---it means the controller leaves only a fraction of the necessary safety distance. At the same time, the bad performance of \AC{} was predicted, too---it has no safety mechanism that tries to ensure RSS safety distances. 

Both \ACCA{} and \ACGA{} come equipped with BC and \DM{} that implement RSS rules that guarantee RSS safety distances. We see, indeed, that the RSS violation was zero or nearly zero for these controllers. We attribute the rare RSS violations by \ACGA{}  to the implementation details---especially to the current ``surrogate implementation'' of proper responses (cf.\ \cref{subsec:impl}). 
In any case, the degree of RSS violation by  \ACGA{}  was small (max.\ 5.16\% of the required distance), the level of error that can be easily caused by other uncertainties such as sensor inaccuracies. We do not expect  serious safety concerns from these small RSS violations.

To conclude, we observed that \ACGA{}  successfully ensured safety, by not only avoiding collisions but also  respecting RSS safety distances (modulo minor exceptions that we attribute to our surrogate implementation of BC). This is much like \ACCA{} and unlike \AC{}.


\subsubsection{\cref{rq:goal}: \goalstring}

Both \AC{} and \ACGA{} managed to achieve the goal 100\% of the time,
as expected.
\ACCA{} did not perform as well, and only achieved the goal in 97.3\%
of scenario instances.
See~\cref{subsubsec:exp:notable:overshoot} for an example
scenario instance where \ACCA{} does not achieve the goal.
In situations where achieving the objective is of high
importance (e.g.~exiting a highway) or cannot be delayed (e.g.~an
automated emergency stop), this 2.7\% difference is significant.

To conclude,
\ACGA{} managed to achieve the goal (while maintaining safety).
\AC{} also managed to achieve the goal (but at the expense of safety), while for \ACCA{}, safety comes at the expense of goal achievement.


\subsubsection{\cref{rq:practicality}: \practicalitystring}

This question can be split into two concerns.
\begin{itemize}
 \item (The strength of RSS conditions) Are the RSS conditions  weak enough, so that they are true in many driving situations? If yes, it means the GA-RSS rule set (\cref{ex:globalPullOver}) is widely applicable.

 \item (The computation cost) Is the computational time manageable? Is it small enough to be computable at each control loop?
\end{itemize}

The first point is hard to analyze quantitatively. Our rule set was applicable to only $52\%$ of the total number
of scenario instances---as we discussed in \cref{subsec:exp:setting}---but this does not mean that our rule set has overly restrictive RSS conditions. We expect that the remaining $48\%$ are such that \emph{no possible control} can safely achieve the goal there (e.g.\ $\ytgt$ is too close to stop at). To show that this expectation of ours is indeed the case, we need to show the behavior of an ideal controller for each scenario instance, which is hard and is left as future work. 

As a more practical consequence of the above consideration, we will pursue an automated search-based method for identifying proper responses---using e.g.\ evolutionary computation techniques similar to~\cite{LuoZAJZIWX21ASEtoAppear}---so that it  either 1) shows probable adequacy of an existing current rule set (in case it does not find a new proper response) or 2) adds a new proper response to the rule set (in case it does). 

In any case, after the compositional derivation described in \cref{sec:workflow} and our manual inspection of experiment results (which allowed us to identify the notable instances shown in \cref{subsec:exp:notable}), we are pretty confident that the rule set in \cref{ex:globalPullOver} is as extensive and widely applicable as it can be. Let us nevertheless emphasize that adding a newly discovered proper response is easy and done modularly (cf.\ \cref{fig:propagation}).


For the second point, recall that the GA-RSS rule set is derived in advance; therefore, the runtime task is merely to check the truth of RSS conditions (the truth of the preconditions $A_{w,u}$ to be precise, see \cref{ex:globalPullOver}). We can expect that the computational cost for doing so is light. 

Indeed, the typical execution time for one of the more complicated
preconditions on commodity hardware (2.9 GHz quad-core Intel Core i7) was
\SI{19.83}{\micro\second}, averaged over $10^{6}$ computations. This means that the computational cost is manageable. It will remain so, even in the future where we have hundreds or thousands of RSS conditions.

To conclude, the GA-RSS rule set in \cref{ex:globalPullOver} can indeed be used in practice, both
from the point of view of applicability and computation cost.

\subsubsection{\cref{rq:metrics}: \metricsstring}

We measured progress and comfort.
Here, we expect \AC{} to be the best in both these metrics, \ACCA{} to perform slightly worse
 (because it is more constrained by its RSS
conditions), and \ACGA{} to perform poorer still (because its RSS conditions
are stronger than those of \ACCA{}).

In terms of progress, \ACCA{} does not reach the goal as fast as \AC{}
on average (\SI{16.23}{\second} and \SI{14.97}{\second} respectively).
However, contrary to expectations, \ACGA{} performs comparably to
\AC{} on average (\SI{14.47}{\second}), and better than \ACCA{}.
To compare \ACGA{} to other controllers, let us look more closely at
\cref{fig:progress}.
We can see that, in most instances, the controllers behave similarly, with the distribution of scenario instances
concentrated around the diagonal.
There are however some outliers on the bottom right of the figures,
where \ACGA{} performs better than the other controllers.
See~\cref{subsubsec:exp:notable:bold} for an example of such an
outlier.

Moreover, when comparing the maximal travel times of the controllers,
\ACGA{} performed much better (\SI{20.7}{\second}) than both \AC{} and
\ACCA{} (\SI{27.8}{\second} and \SI{26.6}{\second} respectively).
This can be explained by the existence of scenario instances in which
\ACGA{} accelerates to overtake \POV{1} (because it knows it is safe
to do so), while the other controllers do not.
In emergency situations (e.g.~health emergency), this can be a
significant time gain.


In terms of comfort, as expected, \ACGA{} does not perform as well as
\AC{} or \ACCA{}.
On average, it accumulates \SI{0.80}{\metre\per\second^2} of jerk,
against \SI{0.45}{\metre\per\second^2} for \AC{} and
\SI{0.36}{\metre\per\second^2} for \ACCA{}.
Maximal accumulated jerk follows the same pattern.
In \cref{fig:comfort}, we give more details about the comparison of
accumulated jerk.
We can see that \ACGA{} consistently performs worse than both \AC{}
and \ACCA{} (with nearly all scenario instances above the diagonal),
and sometimes much worse (outliers at the top-left
of~\cref{fig:AC+CA-AC+GA-jerk}).

This can be explained by the fact that the GA-RSS rule set takes
control more often, and the proper response can be harsh,
e.g.~accelerating quickly to overtake \POV{1}.

To conclude, as expected, \ACGA{} performs poorly in terms of comfort.
Surprisingly however, it performs comparably to or better than the
other controllers in terms of progress.

\subsubsection{\cref{rq:intrusion}: \intrusionstring}

We expect \ACGA{} to take control more often than \ACCA{}, since its
RSS conditions are more strict.
And indeed, the GA-RSS-supervised controller was more intrusive, taking control for 34.6\%
of the journey, on average, compared to only 9.8\% for \ACCA{}.


We would say that the level of intrusiveness (34.6\%) by \ACGA{} is acceptable, especially because the goal it achieves is an imminent one (such as pull over or taking a highway ramp). We can also say that the level of intrusiveness is rather low, which is enabled by the use of GA-RSS rules in the simplex architecture (\cref{subsec:introRSSSafetyArchitecture})---the control is given back to AC whenever it can.




\subsection{Notable Scenario Instances}
\label{subsec:exp:notable}

To better understand some of the results, we analyse in detail two notable
scenario instances in which the GA-RSS-supervised controller \ACGA{} improves upon the
behaviour of \AC{} and \ACCA{}.

\begin{figure}[tbp]\centering
  \includegraphics[scale=0.5]{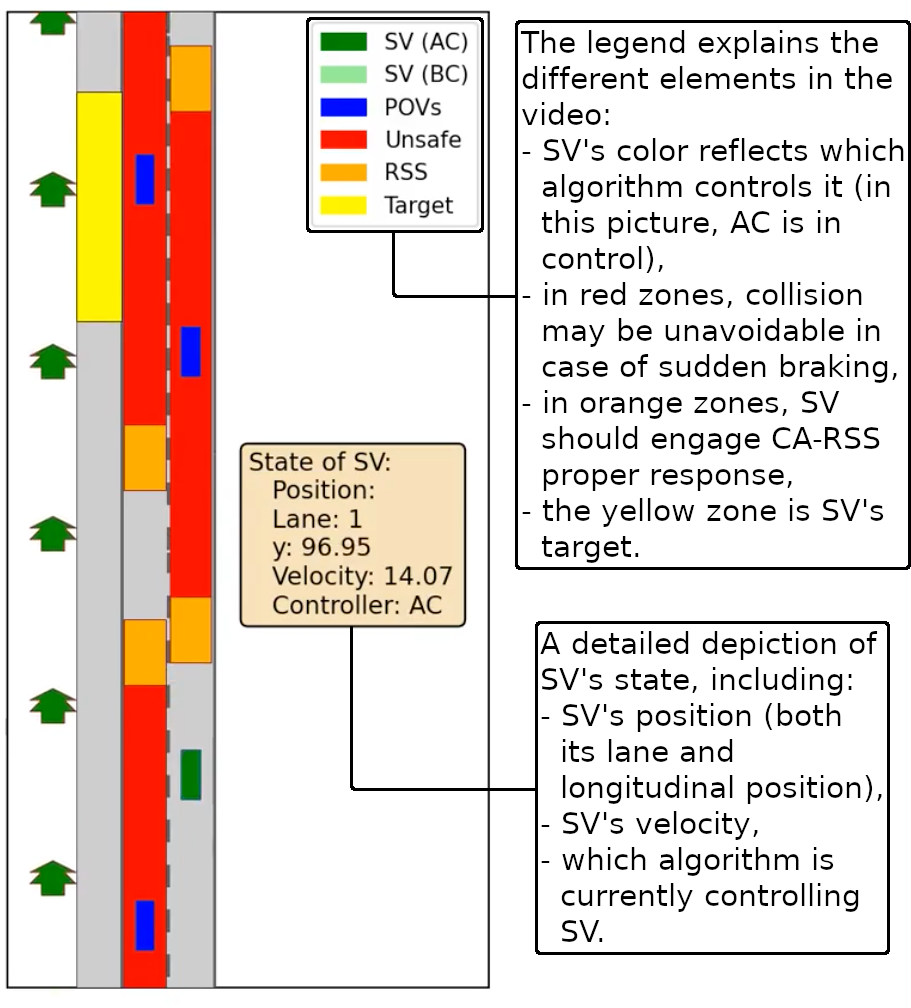}
  \caption{A sample screenshot from a notable scenario instance}
\label{fig:notableScenarioScreenshot}
\end{figure}

Video animations of these scenario instances are provided on the web; see \cref{fig:notableScenarioScreenshot} for a sample screenshot.
  We present this screenshot to explain the videos, and the reader
  should look at the videos rather than the screenshot for more
  information.

\todooptil{make the hosting more precise. (and host! (google drive and bit.ly is fine for the moment)) \\

Ichiro: I suggest we have several screenshots in the paper. Perhaps 4 or 6, arranged horizontally, occupying half a page. 
} 
In these videos the zones
for which $\dRSS(v_{f}, v_{r}, 0)$ would be violated are coloured in red. 
Similarly, the orange zones indicate a violation of $\dRSS(v_{f}, v_{r}, \rho)$. (Recall that the three-argument version of $\dRSS$ is from \cref{ex:oneway-proof}.)
Entering the red zone is unsafe, as collisions may no longer be avoidable. When
the \SV{} is in an orange zone, it must engage in the CA-RSS proper response
within time $\rho$, or risk entering the red zone. \SV{}'s colour reflects which
controller is presently active: dark green for \AC{} and light green for \BC{}.

\subsubsection{Preventing Overshoot}
\label{subsubsec:exp:notable:overshoot}


In this scenario, \POV{1} starts slightly behind \SV{}. All other cars and the
target are rather close, and all cars travel rather fast. The concrete scenario
instance parameters are as follows:
  \[
    \begin{array}{l}
      y_{1}(0) = -5, \ y_{2}(0) = 75, \ y_{3}(0) = 85, \ \xtgt = 140,\\
      v(0) = 14, \ v_{i}(0) = 14.
    \end{array}
  \]
The observed behaviours are as follows:
\begin{itemize}
  \item \AC{}: \SV{} merges in front of \POV{1} when it is unsafe to do so. It
        manages to accomplish the goal in \SI{9.8}{\second}, but violates the
        RSS safety distance by 82\% with respect to \POV{1}, for
        \SI{0.8}{\second}. In the video\videoAC, we can
        see that, when changing lanes, \SV{} crosses over \POV{1}'s red zone.
  \item \ACCA{}: The CA-RSS-supervised controller repeatedly interrupts \AC{} as
        it is attempting to merge in front of \POV{1}, because the distance in
        front of \POV{1} is less than the RSS safety distance. Eventually \AC{}
        is forced to abandon merging into lane 2, and this results in \SV{}
        failing to accomplish the goal. The RSS minimum safety distance is never
        violated. In the video\videoACRSSCA we see that
        \AC{} tries to overtake \POV{1}, but is repeatedly blocked by \BC{},
        which prevents \SV{} from entering the red zone. \BC{} then immediately
        returns control to \AC{} to make the same action again.
  \item \ACGA{}: For the GA-RSS-supervised controller, none of the RSS
        conditions for any of the rules which merge in front of \POV{1} were
        satisfied. This resulted in a proper response for merging behind \POV{1}
        to be engaged. \SV{} brakes so as to merge behind \POV{1} and
        successfully stops at the target in \SI{14}{\second} without ever
        violating the RSS safety distance. In the
        video\videoACRSSGA, at first our \ACGA{}
        exhibits the same behaviour as \ACCA{}. However, when \SV{} has no
        choice but to brake in order to safely reach the goal, then \BC{} takes
        control to slow down and merge behind \POV{1}.

\end{itemize}
In this scenario, \AC{} accomplishes the goal at the expense of safety: an RSS
safety distance violation of 82\% would surely lead to \POV{1} engaging
in dangerous evasive actions, which may in turn lead to a loss of control and
possibly a collision.

Both \ACCA{} and \ACGA{} prevent \SV{} from violating the RSS safety
distances, but in different ways. \ACCA{} corrects \AC{}'s behaviour but only
takes safety into account, which leads to failing the goal. \ACGA{} also
corrects \AC{}'s behaviour by taking the goal into account as well, and
therefore manages to accomplish the goal while maintaining safety.

The accumulated uncomfortable jerk was \SI{1.17}{\metre\per\second^2},
more than double that of \ACCA{} (\SI{0.47}{\metre\per\second^2}) or \AC{} (\SI{0.50}{\metre\per\second^2}).

In conclusion, in this scenario instance, \ACGA{} was the only controller to safely achieve the goal: \AC{}
achieved the goal, but violated the RSS safety distance when cutting in front of
\POV{1}, and \ACCA{} overshot the goal.

\subsubsection{Bold but Safe}
\label{subsubsec:exp:notable:bold}
In this scenario, \POV{1} is in front of \SV{}, but \SV{} is faster. At first
glance, this does not seem like a situation where merging in front of \POV{1} can
be done safely. The concrete scenario instance parameters are as follows:
  \[
    \begin{array}{l}
      y_{1}(0) = 10, \ y_{2}(0) = 75, \ y_{3}(0) = 90, \ \xtgt = 160,\\
      v(0) = 14, \ v_{i}(0) = 10.
    \end{array}
  \]
The observed behaviours are as follows:
\begin{itemize}
  \item \AC{}: As can be seen in the video\videoACtwo,
        \SV{} merged behind \POV{1}, and successfully stopped at the target
        area, taking a total time of \SI{15.9}{\second}. The RSS safety
        distance is violated for \SI{0.6}{\second}, by a maximum of 25.6\%.
  \item \ACCA{}: The behaviour of \ACCA{} similar to that of \AC{}, taking a
        total time of \SI{23.3}{\second} (see the
        video\videoACRSSCAtwo), however the RSS safety distance is not violated.
  \item \ACGA{}: \SV{} accelerated so as to overtake \POV{1} (knowing it is safe
        to do so), merging in front of it, and then stopped in the target area.
        The total time taken was \SI{11.8}{\second}. The RSS safety
        distance was respected. As can be seen in the
        video\videoACRSSGAtwo, at first \SV{}'s path to
        merging in front of \POV{1} seems totally blocked by overlapping red
        zones. However, by accelerating (to put itself in a favourable position
        between \POV{1} and \POV{2}) then braking (thus reducing the sizes of
        the red zones), it manages to open a window through which it can merge
        in front of \POV{1}.
\end{itemize}
In this case we observe that \ACGA{} is able to engage in bold behaviour,
overtaking \POV{1}, in a situation where \AC{} and \ACCA{} simply merge behind
\POV{1}. \ACGA{} is able to engage in such bold behaviour to improve progress
because a mathematical proof exists that it will be able to respect
safety distances while also achieving the goal.

The discomfort level was roughly the same for all controllers:
\SI{0.88}{\metre\per\second^2} for \AC{}, \SI{0.82}{\metre\per\second^2} for
\ACCA{}, and \SI{0.88}{\metre\per\second^2} for \ACGA{}.

To conclude, \ACGA{} performed better than the other controllers in this scenario instance. Indeed, all
three controllers managed to reach the goal, but \ACGA{} performed better
than both \AC{} and \ACCA{} in terms of progress.

\section{Conclusions}\label{sec:concl}

In this paper, we proposed a \emph{goal-aware} extension of responsibility-sensitive safety (RSS), so that RSS rules ensure not only  collision-avoidance but also achievement of goals such as pulling over at a desired position. 

Derivation of goal-aware RSS rules involves
 complex planning that ranges over multiple manoeuvres. Our approach is to deal with such complex reasoning with \emph{program logic}, specifically a program logic $\dHL$ that we introduce as an extension of classic Floyd--Hoare logic. 

We presented a $\dHL$-based compositional workflow for deriving goal-aware RSS rules, in which one can systematically 1) split a driving scenario into smaller subscenarios, 2) design proper responses for those subscenarios, and 3) compute the preconditions for those proper responses. Our current software support by Mathematica is only partially formal, yet provides enough automation and traceability to be practical. We are also working on a fully formalized implementation. 

We conducted experiments in which RSS rules were used in the simplex architecture. Our comprehensive experiments showed the value of goal-aware RSS rules in 1) statistics (they can realize both goal achievement and collision-avoidance) and 2) notable scenarios (they can realize unexpected bold behaviours whose safety is nevertheless guaranteed).



\bibliographystyle{IEEEtran}
\bibliography{myrefs.bib}

%

\begin{IEEEbiography}[{\includegraphics[width=1in,height=1.25in,clip,keepaspectratio]{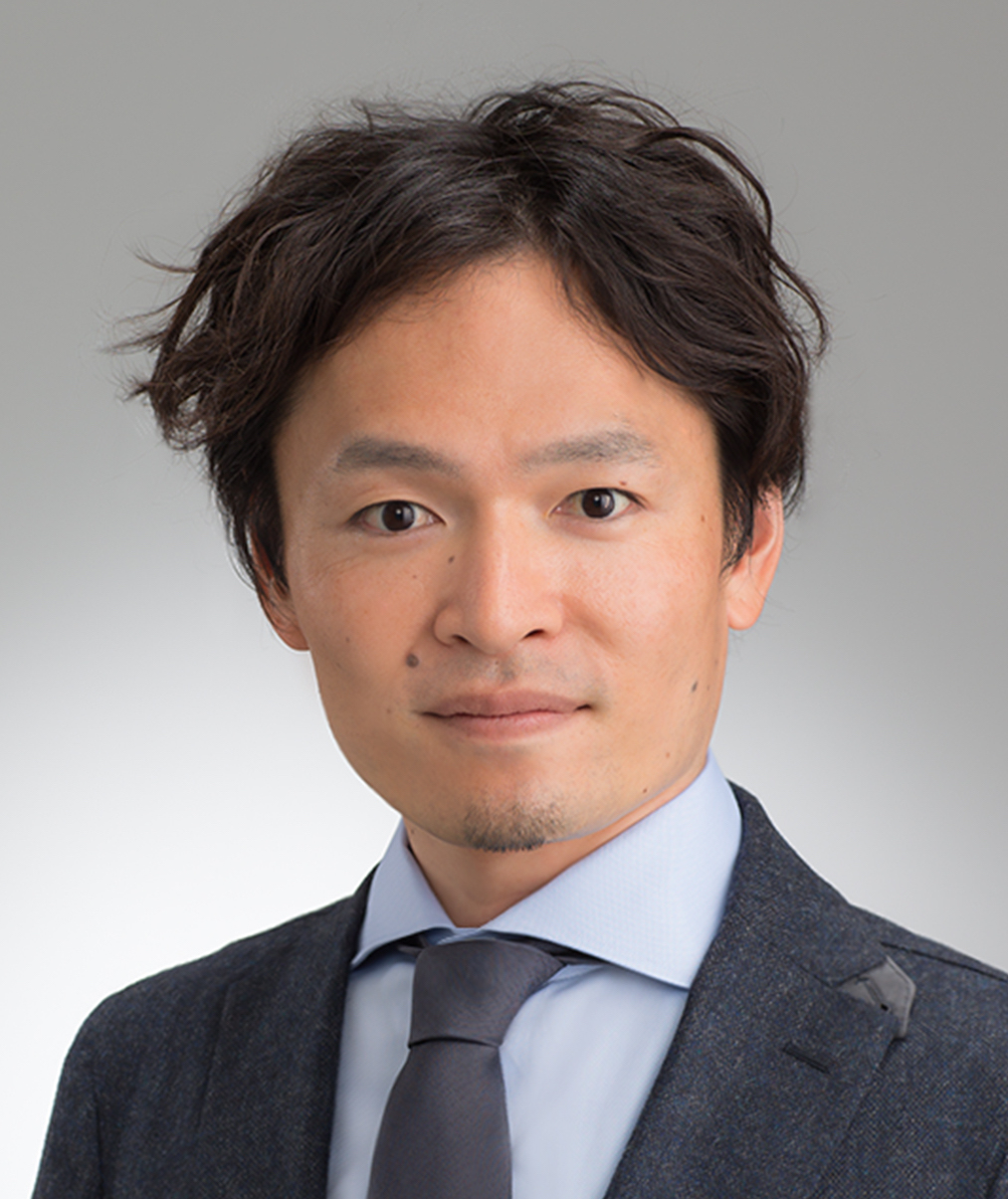}}]{Ichiro Hasuo}
is a Professor at National Institute of Informatics (NII), Tokyo, Japan. He is at the same time the Research Director of the JST ERATO \emph{Metamathematics for Systems Design} Project, and the Director of Research Center for Mathematical Trust in Software and Systems at NII. He received PhD (cum laude) in Computer Science from Radboud University Nijmegen, the Netherlands, in 2008. His research interests include mathematical (logical, algebraic and categorical) structures in software science, abstraction and generalization of deductive and automata-theoretic  techniques in formal verification; integration of formal methods and testing; and their application to cyber-physical systems and systems with statistical machine learning components.
\end{IEEEbiography}

\begin{IEEEbiography}[{\includegraphics[width=1in,height=1.25in,clip,keepaspectratio]{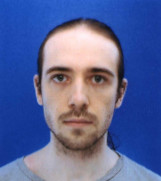}}]{Clovis Eberhart}
  is a project researcher in the JST ERATO \emph{Metamathematics for
  Systems Design} Project at National Institute of Informatics, Tokyo,
  Japan.
  He is also a member of the Japanese-French Laboratory for
  Informatics.
  He received his PhD in Mathematics and Computer Science from
  Universit\'{e} Savoie Mont Blanc, France, in 2018.
  His research interests include semantics of programming languages,
  logic and category theory in computer science, as well as their
  applications to verification.
\end{IEEEbiography}

\begin{IEEEbiography}[{\includegraphics[width=1in,height=1.25in,clip,keepaspectratio]{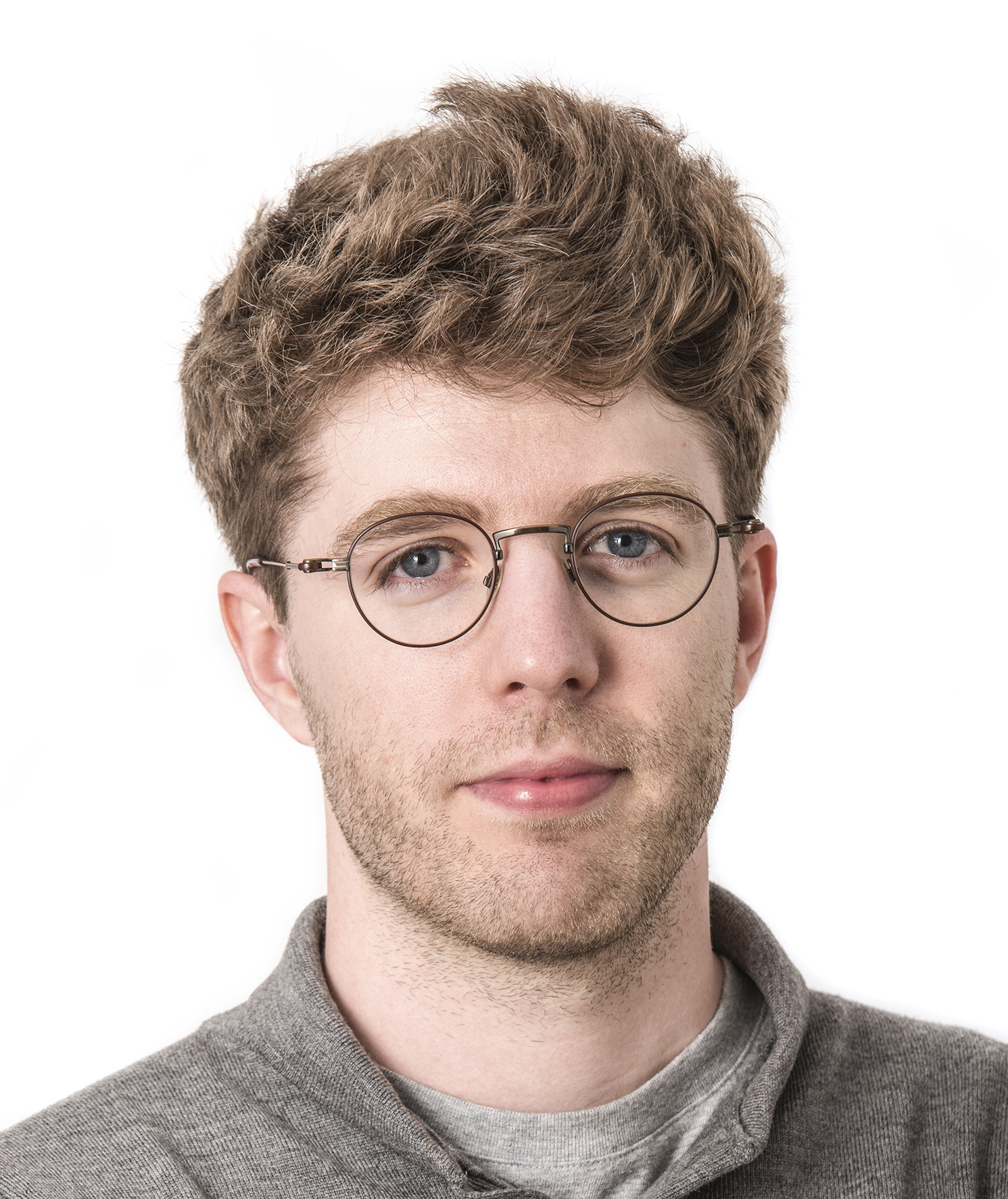}}]{James Haydon}
  is a project technical specialist in the JST ERATO \emph{Metamathematics for
  Systems Design} Project.
  He received his PhD in Mathematics from University of Oxford, United Kingdom, in 2014.
  His research interests include semantics of programming languages, functional programming,
  logic and categories in computer science.
\end{IEEEbiography}

\begin{IEEEbiography}[{\includegraphics[width=1in,height=1.25in,clip,keepaspectratio]{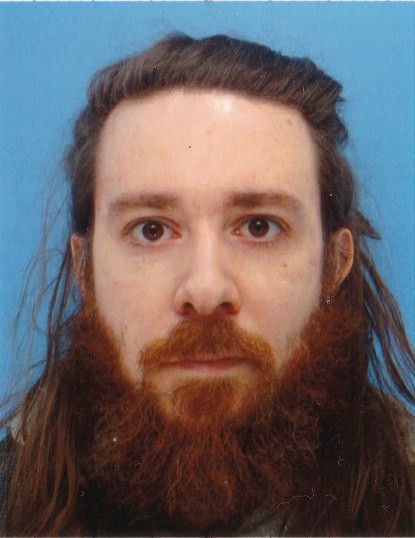}}]{J\'{e}r\'{e}my Dubut}
  is a project assistant professor in the JST ERATO
  \emph{Metamathematics for Systems Design Project}.
  He is also a member of the Japanese-French Laboratory for
  Informatics.
  He received his PhD in Mathematics and Computer Science from
  Université Paris-Saclay, France, in 2017.
  His research interests include category theory, algebraic topology,
  and formalised mathematics.
\end{IEEEbiography}

\begin{IEEEbiography}[{\includegraphics[width=1in,height=1.25in,clip,keepaspectratio]{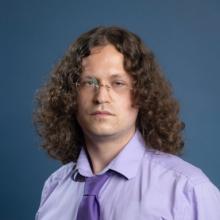}}]{Rose Bohrer}
  is an assistant professor in the Computer Science
  Department at Worcester Polytechnic Institute, USA.
  Their research focuses on formal methods and programming language
  foundations for cyber-physical systems, including interactive
  theorem proving for hybrid systems with applications to driving.
\end{IEEEbiography}

\begin{IEEEbiography}[{\includegraphics[width=1in,height=1.25in,clip,keepaspectratio]{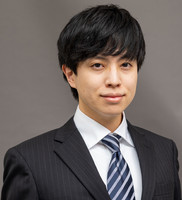}}]{Tsutomu Kobayashi}
  is a researcher at the JST ERATO
  \emph{Metamathematics for Systems Design} Project at National
  Institute of Informatics, Tokyo, Japan.
  He received his PhD from the University of Tokyo in 2017.
  His research interests include formal modeling and verification of
  software systems, theorem proving methods, and software testing.
\end{IEEEbiography}

\begin{IEEEbiography}[{\includegraphics[width=1in,height=1.25in,clip,keepaspectratio]{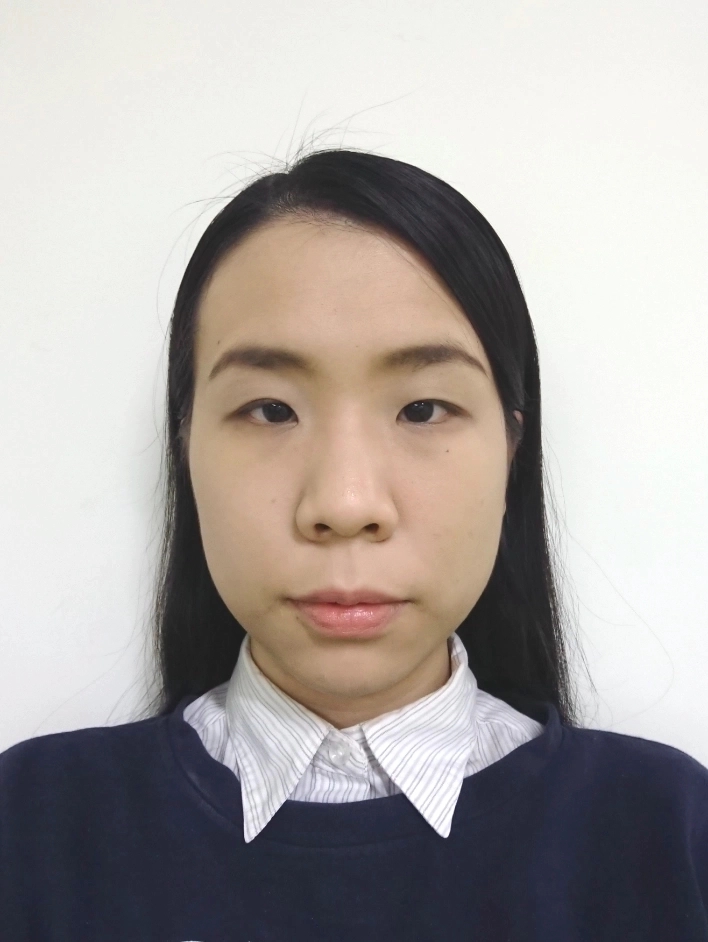}}]{Sasinee Pruekprasert}
  is a project assistant professor in the JST
  ERATO \emph{Metamathematics for Systems Design} Project.
  She received her PhD in Engineering from Osaka University in
  2016.
  Her research interests include supervisory control of discrete event
  systems, abstraction-based controller design, and decision-making of
  autonomous vehicles.
\end{IEEEbiography}

\begin{IEEEbiography}[{\includegraphics[width=1in,height=1.25in,clip,keepaspectratio]{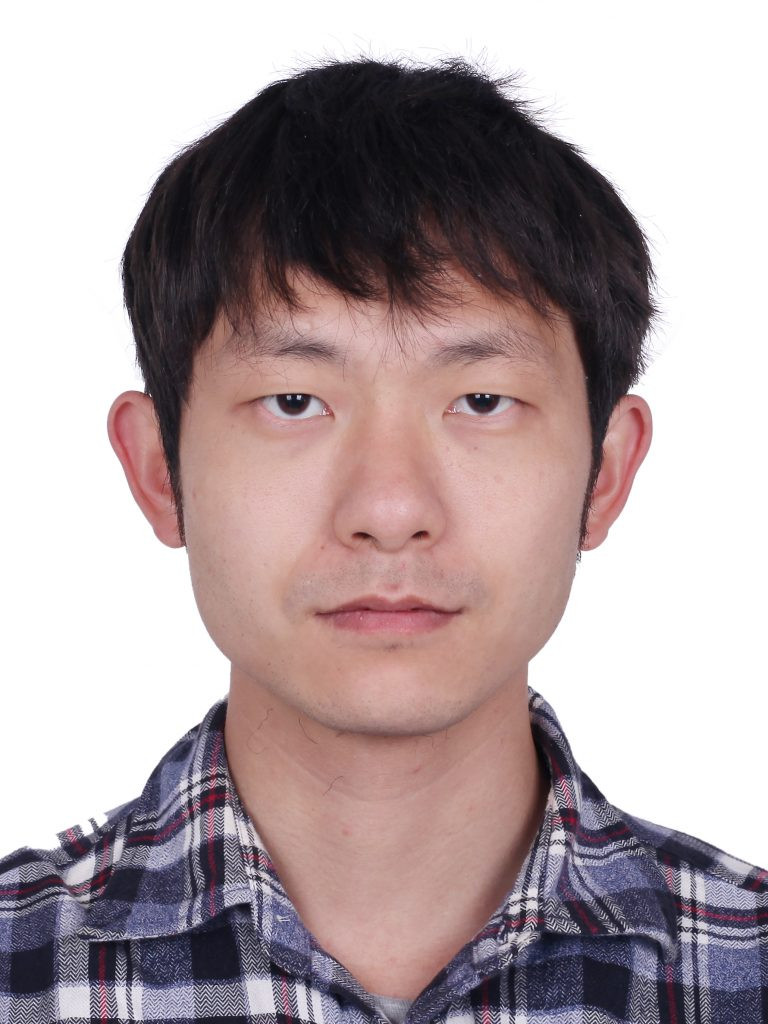}}]{Xiao-Yi Zhang}
  is a project assistant professor at the National Institute of
  Informatics (NII), Japan.
  His main research interests are related to software testing,
  software fault localisation, and hazard analysis for cyber-physical
  systems.
\end{IEEEbiography}

\begin{IEEEbiography}[{\includegraphics[width=1in,height=1.25in,clip,keepaspectratio]{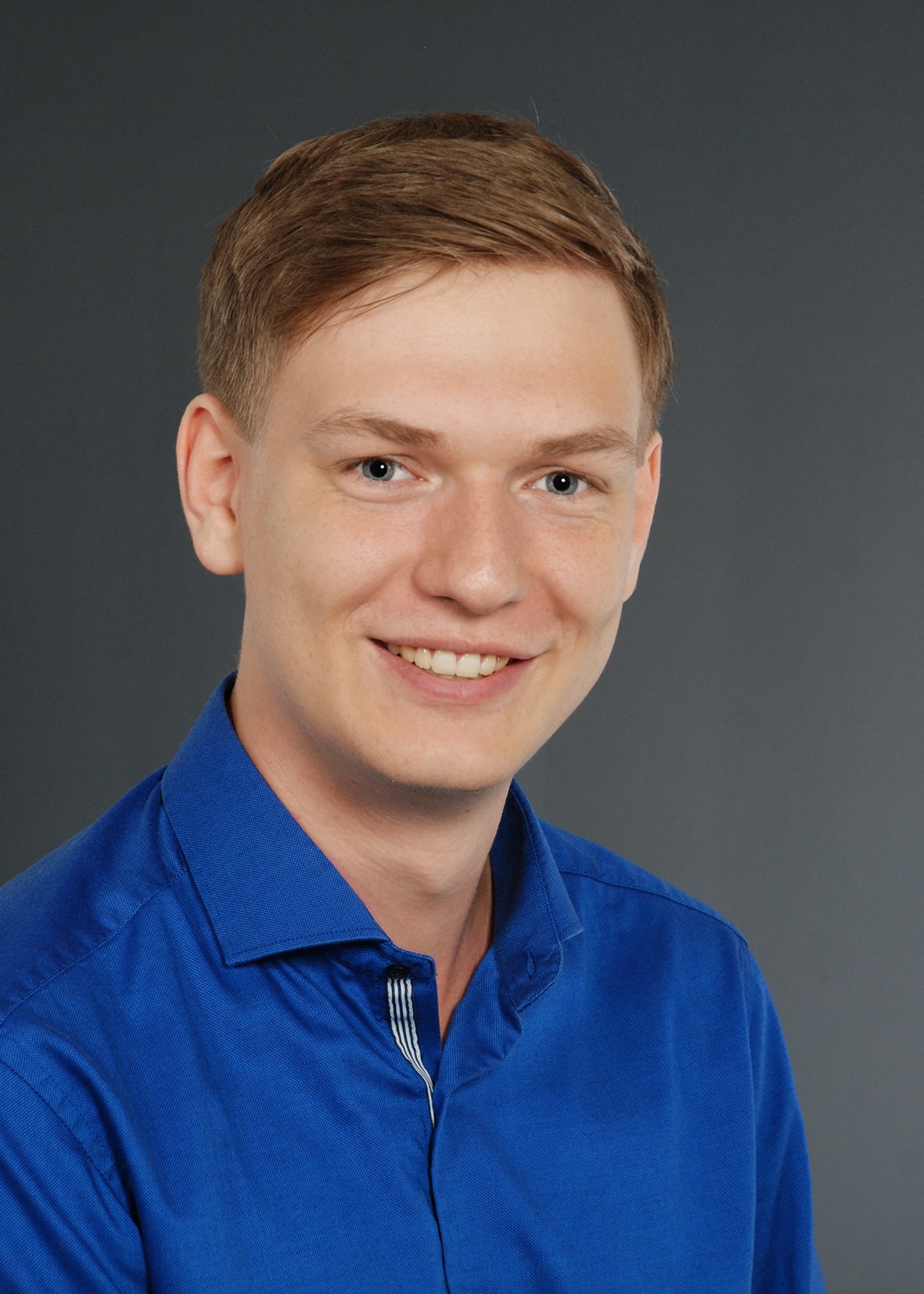}}]{Erik Andr\'{e} Pallas}
  is a student in the Elite Graduate Program Software
  Engineering at University of Augsburg, Technical University of
  Munich and LMU Munich, Germany.
  He is currently writing his Master's thesis on deductive
  verification of safety rules for traffic scenarios in autonomous
  driving.
  His research interests are formal methods for modeling and
  verification of software systems.
\end{IEEEbiography}

\begin{IEEEbiography}[{\includegraphics[width=1in,height=1.25in,clip,keepaspectratio]{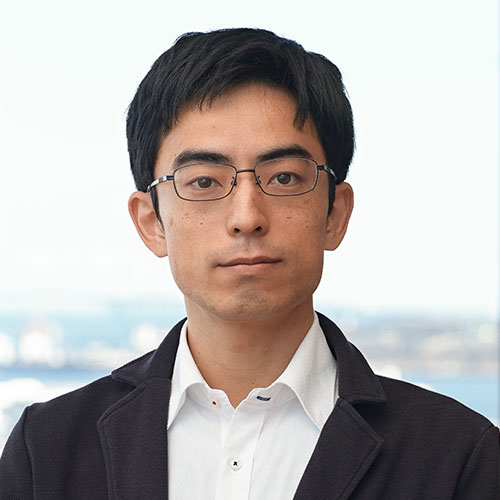}}]{Akihisa Yamada}
  is a senior researcher at National Institute of
  Advanced Industrial Science and Technology, Japan.
  He received his PhD in Information Science from Nagoya University in
  2014.
  His research interest includes term rewriting, termination and
  complexity analysis, and interactive theorem proving.
\end{IEEEbiography}

\begin{IEEEbiography}[{\includegraphics[width=1in,height=1.25in,clip,keepaspectratio]{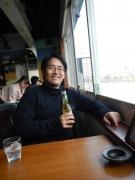}}]{Kohei Suenaga}
  is an associate professor at the Graduate School of
  Informatics, Kyoto University.
  His research focuses on formal verification of software and hybrid
  systems and verification and testing of/for machine learning
  systems.
\end{IEEEbiography}

\begin{IEEEbiography}[{\includegraphics[width=1in,height=1.25in,clip,keepaspectratio]{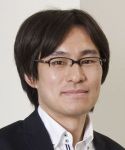}}]{Fuyuki Ishikawa}
  is an associate professor at Information Systems
  Architecture Science Research Division and the deputy director at
  GRACE Center, in National Institute of Informatics, Japan.
  His research focuses on dependability of advanced software systems,
  including testing and verification techniques for autonomous driving
  systems and machine learning-based systems.
\end{IEEEbiography}

\begin{IEEEbiography}[{\includegraphics[width=1in,height=1.25in,clip,keepaspectratio]{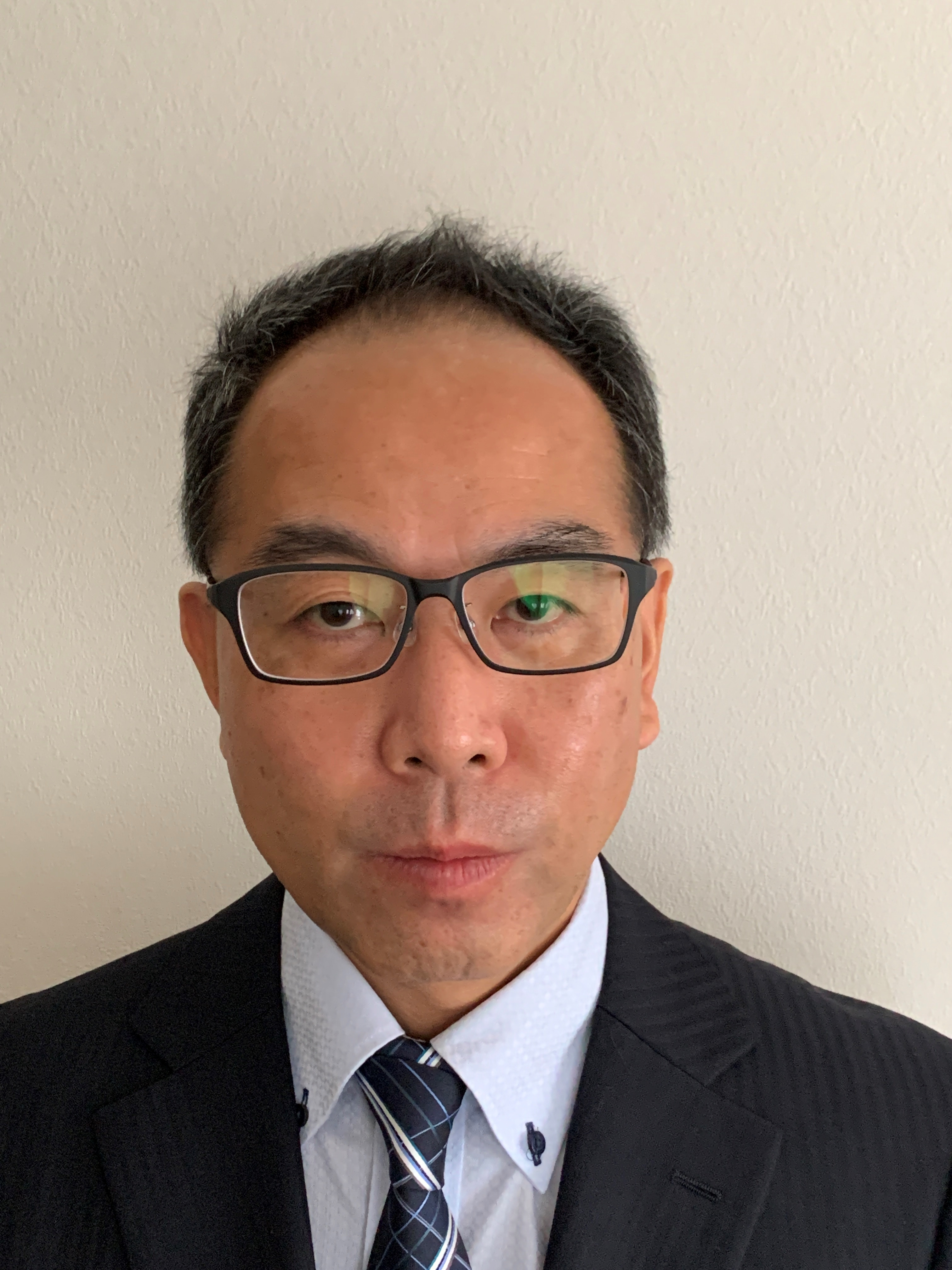}}]{Kenji Kamijo}
  is an Assistant Manager of the Integrated Control
  System Development Division at Mazda Motor Corporation, Hiroshima,
  Japan.
  He received his BS in Electrical and electronic engineering in
  1991 from the Faculty of Engineering, Tokyo Institute of Technology,
  Tokyo, Japan.
\end{IEEEbiography}

\begin{IEEEbiography}[{\includegraphics[width=1in,height=1.25in,clip,keepaspectratio]{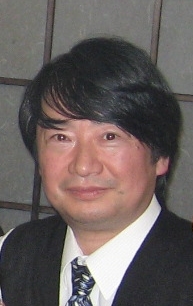}}]{Yoshiyuki Shinya}
  is a Senior Principal Engineer of Integrated
  Control System Development Division at Mazda Motor Corporation,
  Hiroshima, Japan.
  He received M.E.\ degree in Electronics Engineering from the
  University of Osaka in 1984, and MBA degree from University of Kobe
  in 2008.
  After joining Mazda in 1984, he has been engaged in research on
  engine control systems and computer aided control system design.
\end{IEEEbiography}

\begin{IEEEbiography}[{\includegraphics[width=1in,height=1.25in,clip,keepaspectratio]{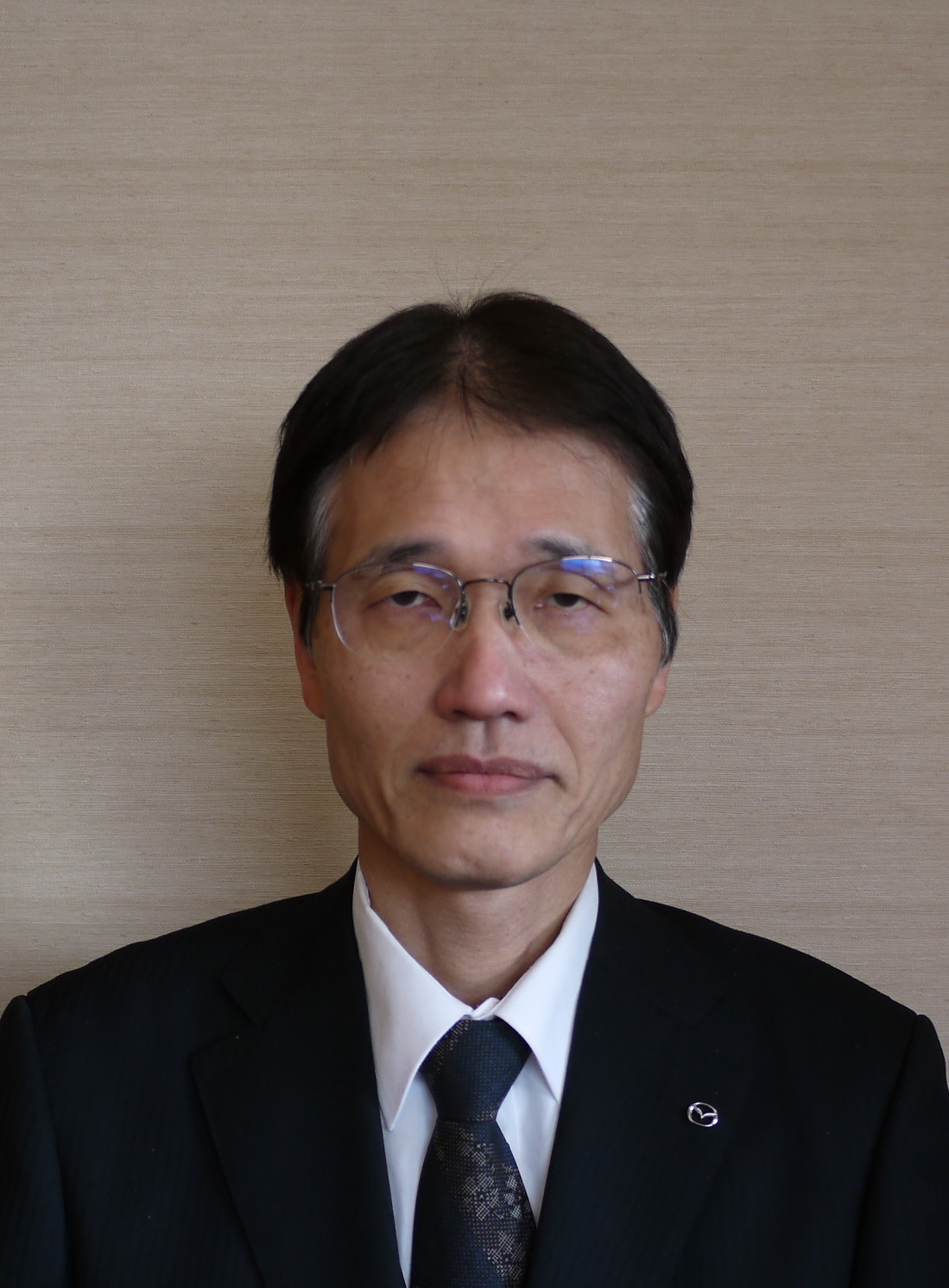}}]{Takamasa Suetomi}
  is a Senior Principal Engineer of Integrated
  Control System Development Division at Mazda Motor Corporation,
  Hiroshima, Japan.
  He received M.E.\ degree in Mechanical Engineering from the
  University of Tokyo in 1987.
  After joining Mazda in 1987, he has been engaged in research on
  man-machine systems, driving simulators, advanced driver assistant
  systems and battery electric-drive systems.
  He is now responsible for development technology for vehicle control
  models.
\end{IEEEbiography}

%
%




\appendix

\subsection{A Formal Proof of the One-Way Traffic Scenario}
\label{app:dRSS}

We want to prove the Hoare triple
\begin{equation*}
  \hquad{\safetya}{\asserta}{\coma}{\assertb}
\end{equation*}
as defined in~\cref{eq:oneway-quad} is valid.
Remember that the different $\dHL$ assertions are defined as
\begin{align*}
  \asserta \; &= \;
  \left(
    v_r \geq 0 \land v_f \geq 0 \land y_f - y_r > \dRSS(v_f,v_r, \rho)
  \right) \rlap{,} \\
  \assertb \; &= \; \left( v_r = 0 \land v_f = 0 \right) \rlap{,} \\
  \safetya \; &= \; \left( y_r < y_f \right) \rlap{,}
\end{align*}
and $\coma$ is defined in \cref{fig:system-one-way-new}.

\begin{figure*}
  \begin{mathpar}
    \bottomAlignProof
    \AxiomC{$\begin{array}{rlll}
      \mathsf{inv}_1\colon & \asserta \Rightarrow \invarianti{1} \sim_1 0 & \bigland_{i=1}^n \varianti{i} \geq 0 \land \bigland_{i=1}^m \invarianti{i} \sim_i 0 \Rightarrow \lieder{\vars}{\funs}{\invarianti{1}} \simeq_1 0\\
        \ldots\\
        \mathsf{inv}_m\colon & \asserta \Rightarrow \invarianti{m} \sim_m 0 & \bigland_{i=1}^n \varianti{i} \geq 0 \land \bigland_{i=1}^m \invarianti{i} \sim_i 0 \Rightarrow \lieder{\vars}{\funs}{\invarianti{m}} \simeq_m 0\\
        \mathsf{var}_1\colon & \asserta \Rightarrow \varianti{1} \geq 0 & \bigland_{i=1}^n \varianti{i} \geq 0 \land \bigland_{i=1}^m \invarianti{i} \sim_i 0 \Rightarrow \lieder{\vars}{\funs}{\varianti{1}} \leq \terminatori{1}\\
        \mathsf{ter}_1\colon & \asserta \Rightarrow \terminatori{1} < 0 & \bigland_{i=1}^n \varianti{i} \geq 0 \land \bigland_{i=1}^m \invarianti{i} \sim_i 0 \Rightarrow \lieder{\vars}{\funs}{\terminatori{1}} \leq 0\\
        \ldots\\
        \mathsf{var}_n\colon & \asserta \Rightarrow \varianti{n} \geq 0 & \bigland_{i=1}^n \varianti{i} \geq 0 \land \bigland_{i=1}^m \invarianti{i} \sim_i 0 \Rightarrow \lieder{\vars}{\funs}{\varianti{n}} \leq \terminatori{n}\\
        \mathsf{ter}_n\colon & \asserta \Rightarrow \terminatori{n} < 0 & \bigland_{i=1}^n \varianti{i} \geq 0 \land \bigland_{i=1}^m \invarianti{i} \sim_i 0 \Rightarrow \lieder{\vars}{\funs}{\terminatori{n}} \leq 0\\
      \end{array}$}
    \RightLabel{(\dwhilerule)}
    \UnaryInfC{$\hquad{\bigland_{i=1}^m \invarianti{j} \sim_j 0 \land \bigland_{i=1}^n \varianti{i} \geq 0}{\asserta}{\dwhileClauseNb{\bigland_{i=1}^n \varianti{i} > 0}{\odeClause{\vars}{\funs}}}{\biglor_{i=1}^n (\varianti{i} = 0 \land \bigland_{j \neq i} \varianti{j} \geq 0)}$}
    \DisplayProof
  \end{mathpar}
  \caption{A more general $\dwhileKeyword$ rule.
  It accommodates any number of invariants $\invarianti{1}$, \ldots,
  $\invarianti{m}$ and variants $\varianti{1}$, \ldots,
  $\varianti{n}$.}
  \label{fig:dwhile}
\end{figure*}

In order to do this, we need to be able to define $\dRSS(v_f, v_r,
\rho)$ as a term of our syntax, so we make the following addition to
our setting (see \cref{rem:term_extension}): if $e,e'$ are terms, then
$\max(e,e')$ is a term, and its partial derivatives are
\[
  \frac{\partial}{\partial x}(\max(e,e')) = \left\{\begin{array}{ll}
    \displaystyle \frac{\partial e}{\partial x} & \text{if $e \geq e'$} \\
    \displaystyle \frac{\partial e'}{\partial x} & \text{otherwise,}
  \end{array}\right.
\]
which is also a term.
Note that, if $\funsa$ and $\funsb$ are locally Lipschitz continuous,
then so is $\dot{\vars} = \max(\funsa,\funsb)$ (where $\max$ on a list
is defined as the pointwise $\max$).

\begin{remark}
  Since $\dRSS(v_f, v_r, \rho)$ is equal to $\max(0,\dRSS_\pm(v_f,
  v_r, \rho))$ (see \cref{ex:oneway-proof}), one may think of
  replacing $y_f - y_r > \dRSS(v_f, v_r, \rho)$ with the equivalent
  formula
  \[
    y_f - y_r > 0 \land y_f - y_r > \dRSS_\pm(v_f, v_r, \rho) \rlap{,}
  \]
  as this would spare us the need to add $\max$ to our setting.
  However, these assertions are not preserved by the dynamics, so we
  do need to introduce $\max$ into our setting for the proof to go
  through.
\end{remark}

Because, in this scenario, we do not know in which order different
events happen (e.g.~which car stops first), we will need a
$(\dwhilerule)$ rule that uses several variants.
We thus need a more general $(\dwhilerule)$ rule than the one in
\cref{fig:dFHL-rules}, and use the one in Figure~\ref{fig:dwhile}
(see~\cref{rem:extensionOfWandDW}).

\begin{figure*}
  \centering
  \AxiomC{(\cref{subsec:drss-proof:1})}
  \UnaryInfC{$\hquad{\safetya}{\asserta}{\comaone}{\assertc}$}
  \AxiomC{(\cref{subsec:drss-proof:2})}
  \UnaryInfC{$\hquad{\safetyinv}{\assertc}{\comatwo}{\assertd}$}
  \RightLabel{$\rlap{(\limprule)}$}
  \UnaryInfC{$\hquad{\safetya}{\assertc}{\comatwo}{\assertd}$}
  \AxiomC{(\cref{subsec:drss-proof:3})}
  \UnaryInfC{$\hquad{\safetyinv}{\assertd_\top}{\comathree}{\assertb}$}
  \AxiomC{(\cref{subsec:drss-proof:4-7})}
  \UnaryInfC{$\hquad{\safetyinv}{\assertd_\bot}{\comafourseven}{\assertb}$}
  \RightLabel{$(\limprule)$}
  \UnaryInfC{$\hquad{\safetyinv}{\assertd \land v_f \neq 0}{\comafourseven}{\assertb}$}
  \RightLabel{$(\ifrule)$}
  \BinaryInfC{$\hquad{\safetyinv}{\assertd}{\comathreeseven}{\assertb}$}
  \RightLabel{$(\limprule)$}
  \UnaryInfC{$\hquad{\safetya}{\assertd}{\comathreeseven}{\assertb}$}
  \RightLabel{$(\seqrule_2)$}
  \TrinaryInfC{$\hquad{\safetya}{\asserta}{\coma}{\assertb}$ as
    in~\cref{eq:oneway-quad}}
  \RightLabel{$(\limprule)$}
  \DisplayProof
  \caption{Overall proof structure.
  Here, $\coma$ denotes the program in \cref{fig:system-one-way-new},
  $\comai$'s are the program fragments on Line $i$ of
  \cref{fig:system-one-way-new}, and $\comaij$'s are the program
  fragments on Lines $i$--$j$ of \cref{fig:system-one-way-new}.
  $(\seqrule_2)$ denotes two applications of the $(\seqrule)$ rule.
  The rest of the proof is detailed in the corresponding sections.}
  \label{fig:drss-proof}
\end{figure*}
The overall proof structure is illustrated in \cref{fig:drss-proof},
and is simply repeated application of the $(\seqrule)$ rule (as
dictated by the structure of $\coma$), composed with $(\limprule)$
rules to strengthen $\safetya$ to
\[
  S_\inv \; = \; \left( y_f - y_r > \dRSS(v_f, v_r, \rho - t)
  \right) \rlap{.}
\]
The assertions $\assertc$, $\assertd$, $\assertd_\top$, and
$\assertd_\bot$ will be defined later in the corresponding sections.

\subsubsection{Step 1: Line 1 of \cref{fig:system-one-way-new}}
\label{subsec:drss-proof:1}

We define $\assertc$ as follows:
\[
  \assertc \; = \;
  \left(
    \begin{array}{l}
      v_r \geq 0 \land v_f \geq 0 \land t = 0 \land {} \\
      y_f - y_r > \dRSS(v_f,v_r)
    \end{array}
  \right) \rlap{.}
\]
Since $\comaone$ is $\assignClause{t}{0}$, we can show the validity of
$\hquad{\asserta \lor \assertc}{\asserta}{\assignClause{t}{0}}{\assertc}$
directly by $(\assignrule)$.
We can then show the validity desired Hoare quadruple by
$(\limprule)$, since $\asserta \lor \assertc$ (which is equivalent to
$\asserta$) implies $\safetya$.

\subsubsection{Step 2: Line 2 of \cref{fig:system-one-way-new}}
\label{subsec:drss-proof:2}

This part of the proof is detailed in \cref{ex:oneway-proof}.
We define $\assertd$ as
\[
  \assertd \; = \;
    \left(
    \begin{array}{l}
      ((v_f \geq 0 \land t = \rho) \lor (v_f = 0 \land t \leq \rho)) \land {} \\
      v_r \geq 0 \land y_f - y_r > \dRSS(v_f, v_r, \rho - t)
    \end{array}
    \right) \rlap{.}
\]
(Note that, in \cref{ex:oneway-proof}, $\comatwo$ is called $\coma'$,
$\assertc$ is called $\asserta'$, and $\assertd$ is called
$\assertb'$.)

The only point that was left implicit in \cref{ex:oneway-proof} was
the computation of the Lie derivative of $\invarianti{2}$.
We compute the Lie derivative as follows (remember that
$\dynamics{}{}$'s are snippets from~\cref{eq:defDeltas}):
\begin{align*}
  &\liederdyn{\dynamics{f}{},\dynamics{r}{1}}{\invarianti{2}} \\
  &\;= v_f \frac{\partial \invarianti{2}}{\partial y_f}
    -\bmax \frac{\partial \invarianti{2}}{\partial v_f}
    + v_r \frac{\partial \invarianti{2}}{\partial y_r}
    + \amax \frac{\partial \invarianti{2}}{\partial v_r} \\
  &\;\phantom{=}\,+ 1 \frac{\partial \invarianti{2}}{\partial t} \\
  &\;= v_f
    -\bmax
      \left\{
        \begin{array}{l}
          \frac{v_f}{\bmax} \\
          0
        \end{array}
      \right.
    - v_r \\
  &\;\phantom{=}\,+ \amax
      \left\{
        \begin{array}{l}
          - (\rho-t) - \frac{v_r + \amax (\rho-t)}{\bmin} \\
          0
        \end{array}
      \right. \\
  &\;\phantom{=}\,+ 1
      \left\{
        \begin{array}{l}
          v_r + \amax (\rho-t) + \frac{\amax (v_r + \amax (\rho-t))}{\bmin} \\
          0
        \end{array}
      \right. \\
  &\;=
    \left\{
      \begin{array}{ll}
        0 & \text{if $\dRSS_\pm(v_f,v_r,\rho-t) \geq 0$} \\
        v_f - v_r & \text{otherwise.}
      \end{array}
    \right.
\end{align*}

With \cref{ex:oneway-proof}, this concludes this part of the proof.

%
\subsubsection{Step 3: Line 3 of \cref{fig:system-one-way-new}}
\label{subsec:drss-proof:3}

We define $\assertd_\top$ as
\[
  \assertd_\top \; = \;
    \left(
    \begin{array}{l}
      v_f = 0 \land t \leq \rho \land v_r \geq 0 \land {} \\
      y_f - y_r > \dRSS(v_f, v_r, \rho - t)
    \end{array}
    \right) \rlap{.}
\]
Note that $\assertd_\top$ is equivalent to $\assertd \land v_f = 0$,
which is needed for the $(\ifrule)$ rule in \cref{fig:drss-proof} to
be applicable.
We denote by $\comathree'$ and $\comathree''$ the two sub-programs of
$\comathree$, as in
\[
  \comathree \; = \; (\seqClause{\comathree'}{\comathree''}) \rlap{.}
\]
To prove the desired Hoare quadruple, we thus use the $(\seqrule)$ and
$(\limprule)$ rules as follows:
\[
  \AxiomC{$\vdots$}
  \UnaryInfC{$\hquad{\safetyinv}{\assertd_\top}{\comathree'}{\asserte}$}
  \AxiomC{$\vdots$}
  \UnaryInfC{$\hquad{\safetya'}{\asserte}{\comathree''}{\assertb'}$}
  \RightLabel{$(\limprule)$}
  \UnaryInfC{$\hquad{\safetyinv}{\asserte}{\comathree''}{\assertb}$}
  \RightLabel{$(\seqrule)$}
  \BinaryInfC{$\hquad{\safetyinv}{\assertd_\top}{\comathree}{\assertb}$}
  \DisplayProof
\]
Here, $\assertb'$ and $\safetya'$ are assertions that will be defined
later, and $\asserte$ is defined as
\[
  \asserte \; = \;
    \left(
    \begin{array}{l}
      v_f = 0 \land t = \rho \land v_r \geq 0 \land {} \\
      y_f - y_r > \dRSS(v_f, v_r, \rho - t)
    \end{array}
    \right) \rlap{.}
\]

Both branches of the proof above can be proved by an application of
$(\dwhilerule)$ followed by $(\limprule)$.

For $\comathree'$, we apply $(\dwhilerule)$ with the following
invariants, variants, and terminators:
\begin{itemize}
  \item $\invarianti{1}  \; = \; (v_r \geq 0)$,
  \item $\invarianti{2}  \; = \; (v_f = 0)$,
  \item $\invarianti{3}  \; = \; (y_f - y_r - \dRSS(v_f,v_r,\rho-t) >
    0)$,
  \item $\varianti{1}    \; = \; \rho-t$, \quad
        $\terminatori{1} \; = \; -1$.
\end{itemize}
Again, the only non-obvious point is the $\invarianti{3}$ is preserved
by the dynamics, for which we compute
\[
  \liederdyn{\dynamics{r}{1}}{\invarianti{3}}
  =
    \left\{
      \begin{array}{ll}
        0 & \text{if $\dRSS_\pm(v_f,v_r,\rho-t) \geq 0$} \\
        - v_r & \text{otherwise.}
      \end{array}
    \right.
\]
We can show that this quantity is always non-negative:
\begin{align*}
  &\dRSS_\pm(v_f,v_r,\rho-t) < 0 \\
  &\Longleftrightarrow v_r (\rho-t) + \frac{\amax (\rho-t)^2}{2} \\
  &\phantom{\Longleftrightarrow}\, + \frac{(v_r + \amax (\rho-t))^2}{2 \bmin} - \frac{v_f^2}{2 \bmax} < 0 \\
  &\Longrightarrow v_r (\rho-t) + \frac{\amax (\rho-t)^2}{2}  \\
  &\phantom{\Longleftrightarrow}\, + \frac{(v_r + \amax (\rho-t))^2}{2 \bmin} < 0 & (i) \\
  &\Longrightarrow v_r (\rho-t) + \frac{(v_r + \amax (\rho-t))^2}{2 \bmin} < 0 & (ii) \\
  &\Longrightarrow v_r (\rho-t) + \frac{v_r^2}{2 \bmin} < 0 & (iii) \\
  &\Longrightarrow v_r < 0 & (iv)
\end{align*}

Here, $(i)$ holds because $v_f = 0$ (by $\invarianti{2}$) and $\bmax
0$; $(ii)$ because $t \leq \rho$ (by $\varianti{1}$) and $\amax > 0$;
$(iii)$ because $v_r \geq 0$, $t \leq \rho$ (by $\invarianti{1}$
and $\varianti{1}$), $\amax > 0$, and $\bmin > 0$; and $(iv)$ because
$v_r \geq 0$ (by $\invarianti{1}$) and $\bmin > 0$.

This proves the validity of
\[
  \hquad{
    \left(
      \begin{array}{l}
        v_r \geq 0 \land v_f = 0 \land t \leq \rho \land {} \\
        y_f - y_r > \dRSS(v_f, v_r, \rho - t)
      \end{array}
    \right)
  }{\assertd_\top}{\comathree'}{\asserte}
  \rlap{.}
\]
The proof of validity of
$\hquad{\safetyinv}{\assertd_\top}{\comathree'}{\asserte}$ follows by
$(\limprule)$, since the safety condition above implies $\safetyinv$.

For $\comathree''$, we apply $(\dwhilerule)$ with the following
invariants, variants, and terminators:
\begin{itemize}
  \item $\invarianti{1}  \; = \; (v_f = 0)$,
  \item $\invarianti{2}  \; = \; (\rho - t = 0)$,
  \item $\invarianti{3}  \; = \; (y_f - y_r - \dRSS(v_f,v_r,\rho-t) >
    0)$,
  \item $\varianti{1}    \; = \; v_r$, \quad
        $\terminatori{1} \; = \; -\bmin$.
\end{itemize}
We can compute the Lie derivative for $\invarianti{3}$:
\[
  \liederdyn{\dynamics{r}{2}}{\invarianti{3}} =
  \left\{
    \begin{array}{ll}
      (\amax + \bmin) \rho & \text{if $\dRSS_\pm(v_f, v_r, \rho-t)
        \geq 0$} \\
      -v_r & \text{otherwise.}
    \end{array}
  \right.
\]
The first term above is positive because $\amax$, $\bmin$, and $\rho$
all are.
The proof that the second term is non-negative follows the same
pattern as for $\comathree'$.

This proves the validity of
\[
  \hquad{\safetya'}{\asserte}{\comathree''}{\assertb'}
  \rlap{,}
\]
where
\begin{align}
  \begin{aligned}
  \assertb' \; &= \;
    \left(
      \begin{array}{l}
        v_r = 0 \land v_f = 0 \land t = \rho \land {} \\
        y_f - y_r > \dRSS(v_f, v_r, \rho - t)
      \end{array}
    \right) \rlap{,}
  \\
  \safetya' \; &= \;
    \left(
      \begin{array}{l}
        v_r \geq 0 \land v_f = 0 \land t = \rho \land {} \\
        y_f - y_r > \dRSS(v_f, v_r, \rho - t)
      \end{array}
    \right) \rlap{.}
  \end{aligned}
  \label{eq:drss-proof:assertb}
\end{align}
The proof of validity of
$\hquad{\safetyinv}{\asserte}{\comathree''}{\assertb}$ follows by
$(\limprule)$, since $\assertb'$ implies $\assertb$ and $\safetya'$
implies $\safetyinv$.
This concludes the proof of validity of
$\hquad{\safetyinv}{\assertd_\top}{\comathree}{\assertb}$.

\subsubsection{Step 4: Lines 4--7 of \cref{fig:system-one-way-new}}
\label{subsec:drss-proof:4-7}

We define $\assertd_\bot$ as
\[
  \assertd_\bot \; = \;
    \left(
    \begin{array}{l}
      v_f \geq 0 \land t = \rho \land v_r \geq 0 \land {} \\
      y_f - y_r > \dRSS(v_f, v_r, \rho - t)
    \end{array}
    \right) \rlap{.}
\]
Note that $\assertd \land v_f \neq 0$ does imply $\assertd_\bot$,
which is necessary to apply the $(\limprule)$ rule.

We can decompose $\comafourseven$ as
\[
  \comafourseven \; = \;
  \left(
    \seqClause{\comafive}{\ifThenElse{v_f = 0}{\comasix}{\comaseven}}
  \right) \rlap{,}
\]
where the $\comai$'s correspond to the program fragments on Line $i$
of \cref{fig:system-one-way-new}.

To prove the desired Hoare quadruple, we use $(\seqrule)$ as follows
(where $\comasixseven$ denotes the program fragment on Lines 6--7 of
\cref{fig:system-one-way-new}):
\[
  \AxiomC{$\vdots$}
  \UnaryInfC{$\hquad{\safetyinv}{\assertd_\bot}{\comafive}{\assertf}$}
  \AxiomC{$\vdots$}
  \UnaryInfC{$\hquad{\safetyinv}{\assertf}{\comasixseven}{\assertb}$}
  \RightLabel{$(\seqrule)$}
  \BinaryInfC{$\hquad{\safetyinv}{\assertd_\bot}{\comafourseven}{\assertb}$}
  \DisplayProof
\]
Here, $\assertf$ is defined as
\[
  \assertf \; = \;
    \left(
    \begin{array}{l}
      ((v_f = 0 \land v_r \geq 0) \lor (v_f \geq 0 \land v_r = 0)) \land {} \\
      t = \rho \land y_f - y_r > \dRSS(v_f, v_r, \rho - t)
    \end{array}
    \right) \rlap{.}
\]

For $\comafive$, we use the $(\dwhilerule)$ rule with the following
invariants, variants, and terminators:
\begin{itemize}
  \item $\invarianti{1}  \; = \; (\rho - t = 0)$,
  \item $\invarianti{2}  \; = \; (y_f - y_r - \dRSS(v_f,v_r,\rho-t) >
    0)$,
  \item $\varianti{1}    \; = \; v_f$, \quad
        $\terminatori{1} \; = \; \bmax$,
  \item $\varianti{2}    \; = \; v_r$, \quad
        $\terminatori{2} \; = \; \bmin$.
\end{itemize}
Once again, we compute the Lie derivative for $\invarianti{2}$:
\[
  \liederdyn{\dynamics{f}{},\dynamics{r}{2}}{\invarianti{3}} =
  \left\{
    \begin{array}{ll}
      (\amax + \bmin) \rho & \text{if $\dRSS_\pm(v_f, v_r, \rho-t)
        \geq 0$} \\
      v_f - v_r & \text{otherwise.}
    \end{array}
  \right.
\]
The top term above is positive because $\amax$, $\bmin$, and $\rho$
are, and the proof that the bottom one is non-negative is the same as
in \cref{ex:oneway-proof}.
This proves the validity of
\[
  \hquad{
    \left(
      \begin{array}{l}
        v_r \geq 0 \land v_f \geq 0 \land t = \rho \land {} \\
        y_f - y_r > \dRSS(v_f, v_r, \rho - t)
      \end{array}
    \right)
  }{\assertd_\bot}{\comafive}{\assertf}
  \rlap{.}
\]
The validity of
$\hquad{\safetyinv}{\assertd_\bot}{\comafive}{\assertf}$ follows
directly by $(\limprule)$, since the safety condition above implies
$\safetyinv$.

Since $\comasixseven$ is an $\ifKeyword$ construct, the proof
structure is as follows:
\[
  \def\defaultHypSeparation{\hskip.1in}
  \AxiomC{$\vdots$}
  \UnaryInfC{$\hquad{\safetyinv}{\assertf_\top}{\hspace{-3pt}\comasix\hspace{-3pt}}{\assertb}$}
  \AxiomC{$\vdots$}
  \UnaryInfC{$\hquad{\safetya'}{\assertf_\bot}{\comaseven}{\assertb'}$}
  \RightLabel{$(\limprule)$}
  \UnaryInfC{$\hquad{\safetyinv}{\assertf \land v_f \neq 0}{\hspace{-3pt}\comaseven\hspace{-3pt}}{\assertb}$}
  \RightLabel{$(\ifrule)$}
  \BinaryInfC{$\hquad{\safetyinv}{\assertf}{\comasixseven}{\assertb}$}
  \DisplayProof
  \def\defaultHypSeparation{\hskip.2in}
\]
Here, $\assertf_\top$ and $\assertf_\bot$ are defined as follows:
\begin{align*}
  \assertf_\top \;&=\;
    \left(
    \begin{array}{l}
      v_f = 0 \land v_r \geq 0 \land t = \rho \land {} \\
      y_f - y_r > \dRSS(v_f, v_r, \rho - t)
    \end{array}
    \right) \rlap{,} \\
  \assertf_\bot \;&=\;
    \left(
    \begin{array}{l}
      v_f \geq 0 \land v_r = 0 \land t = \rho \land {} \\
      y_f - y_r > \dRSS(v_f, v_r, \rho - t)
    \end{array}
    \right) \rlap{,}
\end{align*}
and $\assertb'$ and $\safetya'$ were defined
in~\cref{eq:drss-proof:assertb}.

Note that $\assertf_\top$ is equivalent to $\assertf \land v_f = 0$,
which allows application of $(\ifrule)$.
Similarly, $\assertf \land v_f \neq 0$ implies $\assertf_\bot$,
$\assertb'$ implies $\assertb$, and $\safetya'$ implies $\safetyinv$,
which allows us to apply $(\limprule)$.

For $\comasix$, we use $(\dwhilerule)$ with the following invariants,
variants, and terminators:
\begin{itemize}
  \item $\invarianti{1}  \; = \; (\rho - t = 0)$,
  \item $\invarianti{2}  \; = \; (v_f = 0)$,
  \item $\invarianti{3}  \; = \; (y_f - y_r - \dRSS(v_f,v_r,\rho-t) >
    0)$,
  \item $\varianti{1}    \; = \; v_r$, \quad
        $\terminatori{1} \; = \; -\bmin$.
\end{itemize}
If we compute the Lie derivative of $\invarianti{3}$, we get
\[
  \liederdyn{\dynamics{r}{1}}{\invarianti{3}}
  =
    \left\{
      \begin{array}{ll}
        0 & \text{if $\dRSS_\pm(v_f,v_r,\rho-t) \geq 0$} \\
        - v_r & \text{otherwise,}
      \end{array}
    \right.
\]
and the proof that this is non-negative follows the same pattern as
that in \cref{subsec:drss-proof:3}.
This proves that
\[
  \hquad{
    \left(
      \begin{array}{l}
        v_r \geq 0 \land v_f = 0 \land t = \rho \land {} \\
        y_f - y_r > \dRSS(v_f, v_r, \rho - t)
      \end{array}
    \right)
  }{\assertf_\top}{\comasix}{\assertb}
\]
is valid.
By $(\limprule)$, we get that
$\hquad{\safetyinv}{\assertf_\top}{\comasix}{\assertb}$, since the
safety condition above implies $\safetyinv$.

For $\comaseven$, we use the $(\dwhilerule)$ rule with the following
invariants, variants, and terminators:
\begin{itemize}
  \item $\invarianti{1}  \; = \; (v_r = 0)$,
  \item $\invarianti{2}  \; = \; (\rho - t = 0)$,
  \item $\invarianti{3}  \; = \; (y_f - y_r - \dRSS(v_f,v_r,\rho-t) >
    0)$,
  \item $\varianti{1}    \; = \; v_f$, \quad
        $\terminatori{1} \; = \; -\bmax$.
\end{itemize}
Again, let us compute the Lie derivative of $\invarianti{3}$:
\[
  \liederdyn{\dynamics{f}{}}{\invarianti{3}}
  =
    \left\{
      \begin{array}{ll}
        0 & \text{if $\dRSS_\pm(v_f,v_r,\rho-t) \geq 0$} \\
        v_f & \text{otherwise.}
      \end{array}
    \right.
\]
The term above is non-negative by $\varianti{1}$, which proves the
validity of
\[
  \hquad{\safetya'}{\assertf_\top}{\comasix}{\assertb'} \rlap{,}
\]
where $\assertb'$ and $\safetya'$ were defined
in~\cref{eq:drss-proof:assertb}.
This concludes the proof of validity of
$\hquad{\safetyinv}{\assertd_\bot}{\comafourseven}{\assertb}$, and
finally that of $\hquad{\safetya}{\asserta}{\coma}{\assertb}$.

\end{document}


%% file: subs/stats.tex
%
%
%
%

{
  \setlength{\tabcolsep}{5pt}
\begin{tabular}{l||r||r|rrrr||rr|rr||r}
  & & & \multicolumn{4}{c||}{RSS violation} & \multicolumn{2}{c|}{time} & \multicolumn{2}{c||}{jerk} &
  \\
  & goal (\%) & collision & num. (\%) & avg. time & max time & max dist & avg. & max & avg. & max & \BC{} time
  \\
  \hline
  \AC{} & 2350 (100\%) & 0 & 300 (12.8\%) & \SI{0.09}{\second} & \SI{0.8}{\second} & 82.16\% & \SI{14.97}{\second} & \SI{27.8}{\second} & \SI{0.45}{\metre\per\second^2} & \SI{2.65}{\metre\per\second^2} & \NA{}
  \\
  \ACCA{} & 2285 (97.2\%)  & 0 & 0 (0\%) & \NA & \NA & \NA & \SI{16.23}{\second} & \SI{26.6}{\second} & \SI{0.36}{\metre\per\second^2} & \SI{0.88}{\metre\per\second^2} & 9.8\%
  \\
  \ACGA{} & 2350 (100\%) & 0 & 15 (0.6\%) & \SI{0.00}{\second} & \SI{0.3}{\second} & 5.16\% & \SI{14.47}{\second} & \SI{20.7}{\second} & \SI{0.80}{\metre\per\second^2} & \SI{4.12}{\metre\per\second^2} & 34.6\%
  \\
\end{tabular}
  \setlength{\tabcolsep}{6pt}
}